\documentclass[review]{elsarticle}

\usepackage{lineno,hyperref}
\modulolinenumbers[5]

\journal{Journal of \LaTeX\ Templates}
\graphicspath{figures/}
\usepackage{amsmath}
\usepackage{subfig}
\usepackage{adjustbox}
\begin{document}

\begin{frontmatter}

\title{Student Mixture Model Based Visual Servoing
}

\author[mainaddress]{Mithun. P}

\author[secondaryaddress]{Shaunak A. Mehta}

\author[secondaryaddress]{Suril V. Shah\corref{correspondingauthor}}
\cortext[correspondingauthor]{Corresponding author}
\ead{surilshah@iitj.ac.in}

\author[secondaryaddress]{Gaurav Bhatnagar}
\author[mainaddress]{K. Madhava Krishna}

\address[mainaddress]{International Institute of Information Technology, Hyderabad-500032, India}
\address[secondaryaddress]{Indian Institute of Technology, Jodhpur-342037, India}
\begin{abstract}
Classical Image-Based Visual Servoing (IBVS) makes use of geometric image features like point, straight line and image moments to control a robotic system. Robust extraction and real-time tracking of these features are crucial to the performance of the IBVS. Moreover, such features can be unsuitable for real world applications where it might not be easy to distinguish a target from rest of the environment. Alternatively, an approach based on complete photometric data can avoid the requirement of feature extraction, tracking and object detection. In this work, we propose one such probabilistic model based approach which uses entire photometric data for the purpose of visual servoing. A novel image modelling method has been proposed using Student Mixture Model (SMM), which is based on Multivariate Student's $t$-Distribution. Consequently, a vision-based control law is formulated as a least squares minimisation problem. Efficacy of the proposed framework is demonstrated for 2D and 3D positioning tasks showing favourable error convergence and acceptable camera trajectories. Numerical experiments are also carried out to show robustness to distinct image scenes and partial occlusion.
\end{abstract}

\end{frontmatter}

\section{Introduction}
In robotic applications where direct control by human intervention is inaccurate and error prone, an autonomous operation can be a more reliable solution. To achieve this autonomy the control system must rely on sensory feedback. Computer vision serves as a powerful component of such a system. The information obtained from such a vision sensor can be used for robot control tasks such as object inspection, grasping, tracking a seam in arc welding etc. Visual servoing \cite{espiau1992new,corke1993visual} is one such approach where a control law regulates an error in the image space to achieve desired positioning of the robot. To achieve this, a set of visual features has to be selected from the image of the object of interest as a primary step. Following this, the current and desired pose can be defined in terms of the selected visual features. Once the feature selection is done, an error function between desired and current image features is modeled. Subsequently, a control law \cite{hutchinson1996tutorial,chaumette2006visual,chaumette2007} is designed to provide motion to the camera such that the error between current and desired image features is minimized to zero leading to the required configuration of the robot. The IBVS commonly uses points \cite{comport2005robust},\cite{lazar2007performance},\cite{xin2016robot}, lines, contours \cite{marchand2005feature},\cite{marchand2005visp} optical flow estimates, etc. as the features. For such features real-time feature tracking and correspondence are important but computationally involved steps of vision-based control. Various approaches and advancements have been proposed for feature tracking \cite{marchand2005feature,pressigout2005real,hafez2007combining}
to overcome this problem upto a limit but it is an unavoidable step. An extension of these approaches has been proposed in \cite{chaumette2004image} where shape and moments are considered as visual features. These features provide more intuitive than the geometric features but a segmentation step is still required before extraction of features. Vision-based control technique applicable to real life environments with diverse objects and robust to noises are the main challenges of IBVS. Therefore, the present work proposes a solution to overcome these challenges.

Previous approaches rely mainly on geometric features. However, additional computations for feature extraction, matching and real-time tracking are some overheads for these visual servoing methods. In this regard, researchers have concentrated towards feature less tracking to avoid the drawbacks of geometric approaches. A work in that direction has been initially presented in \cite{nayar1996subspace} and \cite{deguchi2000direct} where image intensity was used as indirect visual features to perform control action. Later image intensity was directly considered to perform 2D matching of images without any feature extraction in \cite{kallem2007kernel} and \cite{dame2011mutual}. Here, a kernel projected value of intensities in image at each pixel and information theoretic approach of mutual information were illustrated. A similar method called Photometric visual servoing developed in \cite{collewet2011photometric} used image intensities as visual features. Use of the concept of Sum of Conditional Variance \cite{delabarre2012visual} results in a direct visual servoing task which is easy and fast to compute. Even though it is robust towards non-linear illumination variations, it is less robust towards local variations. A depth map obtained from a range sensor was used as a visual feature in \cite{teuliere2012direct}. This approach is both direct (without any 3D pose estimation) and dense (without feature extraction) but it requires an expensive range sensor which may have problems like noise in data and the absence of measurements. The above limitation is overcome in \cite{bateux2017histograms} where a histogram is considered as visual feature. Here, the applicability of the framework was shown to many histograms, like, intensity histograms, Histograms of Oriented Gradients (HOG) and colour histograms. However, this method suffers from low convergence and impact from uninformative image regions of an image under consideration. Although the above mentioned approaches vary in efficiency and precision, it also suffers from several drawbacks. For example, usage of direct image intensities makes it sensitive to illumination changes. Having a relatively small convergence domain of the cost function for these methods requires large overlapping regions between the current and desired images for the algorithm to converge. Many variations of direct visual servoing methods have been studied by considering different image and cost function representations (Photometric Moments \cite{bakthavatchalam2018direct}, Principal Component Analysis \cite{marchand2019subspace}, wavelet transforms \cite{ourak2019direct},  Discrete Cosine Transform \cite{marchand2020direct}) to improve the convergence and robustness. A new approach using feature based mixture models in visual servoing has been initially studied in \cite{hafez2008visual}. Here, feature points extracted from current and desired images were represented by two separate mixture models and the visual servoing minimized the difference between the two mixture models. It is advantageous since it considers feature position uncertainties by probabilistic models. However, the feature point detection is still a mandatory step and points extracted from the current images has to be the same as the points extracted from the desired image. Furthermore, it is implemented to control only three degrees-of-freedom (DOF) of the robot and also lacks an analytical formulation of interaction matrix. In \cite{crombez2015photometric}, dense features were used to model mixture models alternatively to the above method. The main idea of this method is to model every pixel in the image as Gaussian function, and the algorithm tries to reduce the error between desired and current gaussian function of images. However, suffers from the drawback that it requires tuning of an extension parameter experimentally for providing overlapping of image regions.

\begin{figure}[t]
\centering
\includegraphics[width=3.3in]{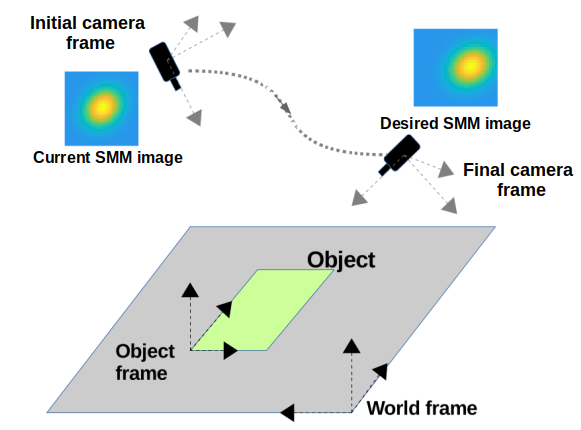}
\caption{Performing SMM based visual servoing using a 6-DOF camera towards an object. The servoing task requires the camera to move to the desired view of target such that an error between proposed SMM images  become zero.}
    \label{fig:smm_teaser}
\end{figure}

In this paper, we propose a novel approach to visual servoing (see Fig. \ref{fig:smm_teaser}) using student $t$-distribution mixture model. In literature, it can be seen that $t$-distribution mixture models are successfully used in image processing applications like image registration \cite{gerogiannis2007robust}. Geometrically a $t$-distribution function has heavily-tailed alternative to the normal distributions with same symmetric and bell-shaped density function. Such a characteristics of $t$-distribution guarantees inclusion of values which is far from its mean value. This property serves as one of the motivation for the proposed visual servoing framework. The general idea of proposed SMM based visual servoing techniques is to minimize error between current and desired SMMs. These SMMs are obtained from corresponding current and desired images, where each image is represented by mixture of $t$-distributions. This way to proceed not only has the advantages of the direct photometric method but also increases the convergence domain considerably. The proposed framework of visual servoing using Students's $t$-distribution is not reported in the literature to the best of authors’ knowledge and forms one of the fundamental contributions of this work. When the image is modeled using $t$-distributions, the control output results in better and smoother convergence than other classical servoing methods. Hence, the present work makes the following contributions.
\begin{itemize}
\item Image modeling using Student $t$-distribution Mixture Model (SMM) and an analytical formulation of interaction matrix. 
\item SMM based framework for IBVS without any requirement of feature tracking and correspondence.
\item Systematic experimental validation of proposed framework.
\end{itemize}

The paper is organized as follows. Section~\ref{sec:prelim} presents outline of basic visual servoing framework and the concept and formation of multivariate student $t$-distribution. This forms the foundation of the proposed research work. Section~\ref{sec:outline} describes outline of the proposed SMM based visual servoing method. The initial part of Section~\ref{sec:img_model}  explains modeling of an image using SMM and later part explains the development of the interaction matrix. 
Section~\ref{sec:smm_vs} explains the modeling of visual servoing framework as an optimization problem. Section~\ref{sec:result} provides extensive experiments which validate task completion, robustness and convergence. Finally, conclusions and future works are presented in Section~\ref{sec:future_work}.
\section{Preliminaries}\label{sec:prelim}
This section presents some preliminaries that surface behind the framework presented.
Initially, an introduction to basic visual servoing control law is provided. Later, discussion on Student's $t$-distribution is carried out which forms a basis for the proposed framework. 
\subsection{Visual servoing}
Visual servoing uses visual information for autonomous control of a robot. On a very basic level the vision based control proposes that a set of simple visual behaviours can be used to accomplish various application tasks. In such applications the visual information extracted from a camera is used as a feedback to close the control loop \cite{chaumette2006visual} so that the robot can reach the desired position accurately. In classical approach \cite{chaumette2007}, the visual information extracted from the images are in the form of a set of features $\boldsymbol{s}$. Therefore, the basic working principle of any IBVS method is to minimise the error between the current features $\boldsymbol{s}$
and the desired features $\boldsymbol{s^*}$, i.e.,
\begin{equation}
\boldsymbol{e} = (\boldsymbol{s}-\boldsymbol{s^*}).
\label{eq:error}
\end{equation}
Here $\boldsymbol{s} \in R^k$ is the feature vector of $k$ feature points. The minimization of such an error function (\ref{eq:error}) is undertaken by providing appropriate velocity to the camera mounted on the end-effector of the robot. The relationship between the camera velocity and feature rate is given by 
\begin{equation}
\boldsymbol{\dot{s}}= \boldsymbol{L}\boldsymbol{t}_c
\label{eq:feature_rate}
\end{equation}
where $\boldsymbol{t}_c \in R^6$ is camera velocity consisting of linear and angular velocities $(\boldsymbol{v},\boldsymbol{w})$, and $\boldsymbol{L} \in R^{k \times 6}$ is the interaction matrix. If the desired feature in \eqref{eq:error} is not changing with time, error rate can be expressed as $\boldsymbol{\dot{e}} =\boldsymbol{\dot{s}}$.
As the main objective of visual servoing is to nullify the error function defined in \eqref{eq:error}, an exponential decay of error is selected, i.e., 
\begin{equation}
\boldsymbol{\dot{e}}=-\lambda \boldsymbol{e},
\label{eq:exp_decay}
\end{equation}
where $\lambda$ is a scalar gain which determines the speed of convergence. Upon substituting \eqref{eq:exp_decay} in to \eqref{eq:feature_rate}, the visual servoing controller can be expressed as
\begin{equation}
\boldsymbol{t}_c = -\lambda \boldsymbol{L^+ e}.
\label{eq:vs_controller}
\end{equation}
Here $\boldsymbol{L^+}$ is the pseudo inverse of the image Jacobian. The specific form of
the interaction matrix depends on the features that are used to define $\boldsymbol{s}$.
The above control law minimizes the error in feature space by guiding the camera with rate, $\boldsymbol{t}_c$ and exponentially decreasing the errors to zero. Several approaches in visual servoing use this generic control law for different applications and scenarios. 

\subsection{Multivariate $t$-distribution}
\begin{figure}[!ht]
\centering
\subfloat[]{\includegraphics[width=1.7in]{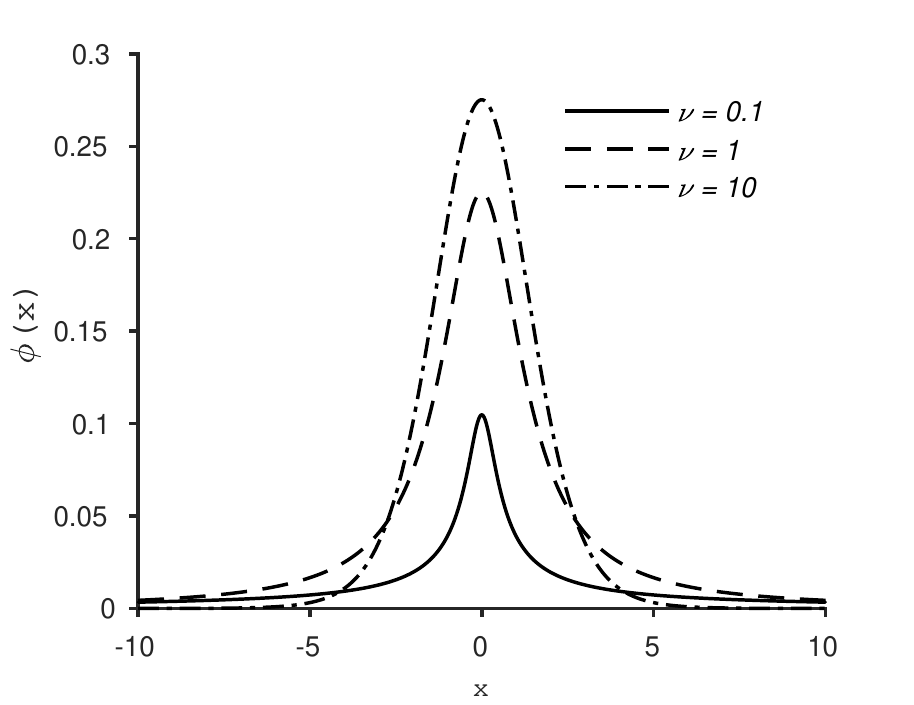}}
\hspace{0.01cm}
\subfloat[]{\includegraphics[width=1.8in]{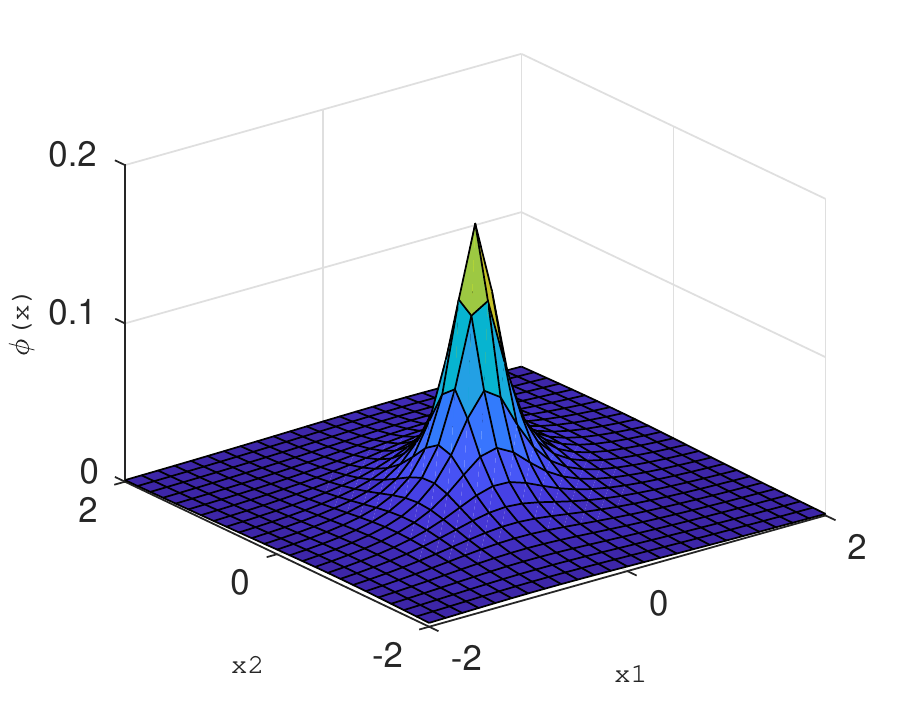}}
\hspace{0.01cm}
\subfloat[]{\includegraphics[width=1.8in]{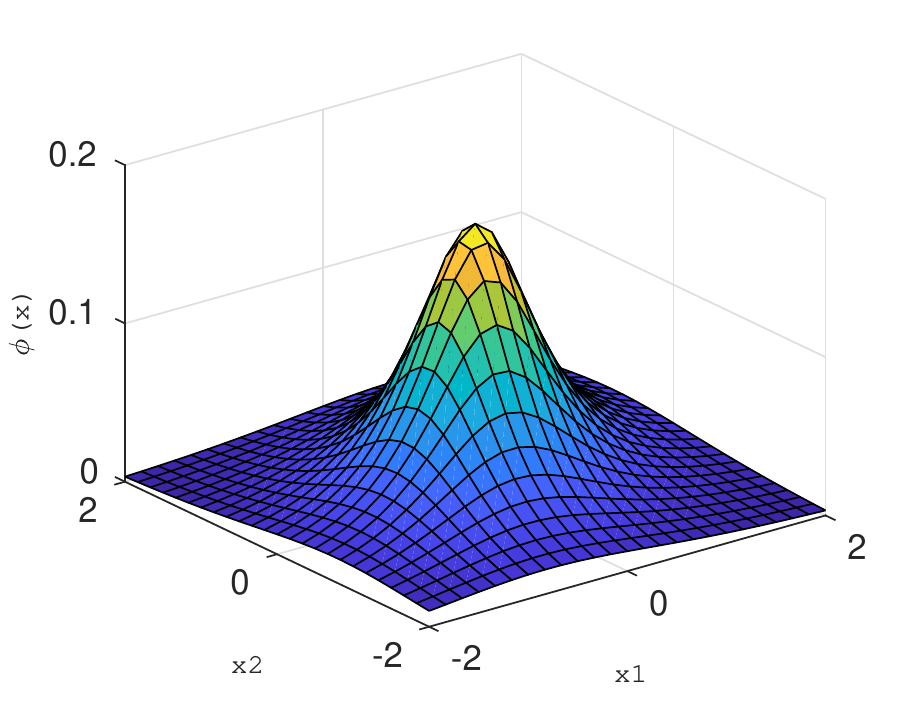}}
\hspace{0.01cm}
\subfloat[]{\includegraphics[width=1.8in]{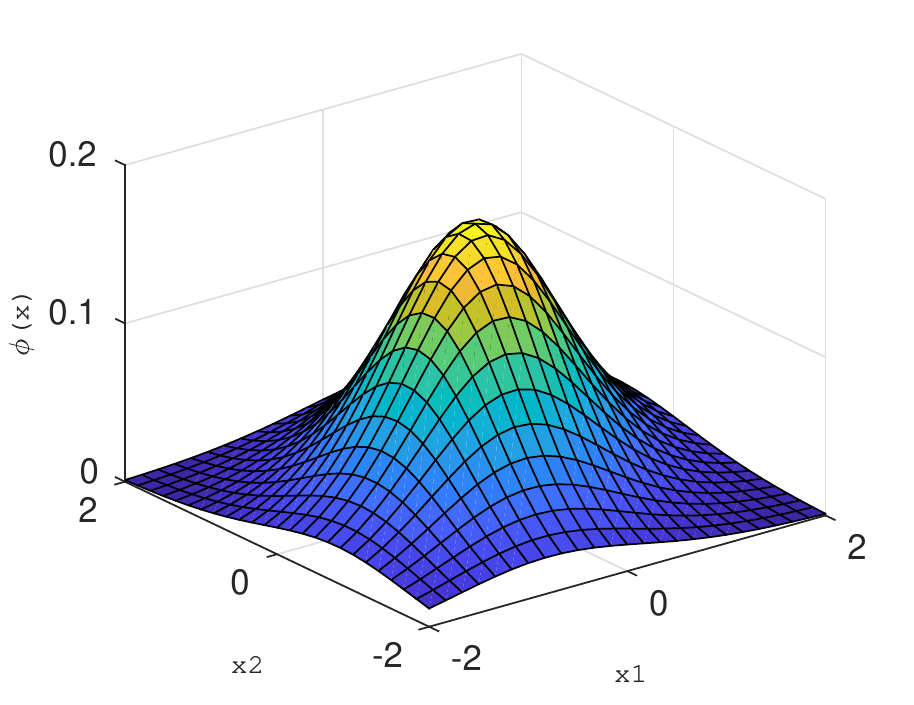}}
\caption{Student’s $t$-distribution for various degrees-of-freedom $\nu$. (a) Univariate case, (b-d) Multivariate cases with $\nu = 0.1,1,10$ respectively.}
\label{fig:t-dist}
\end{figure}

In probability theory, student $t$-distribution is a continuous distribution function describing the probability that a given value will occur. 
Suppose we have a simple random sample of size $n$ drawn from a normal population with mean $\mu$ and standard deviation $\sigma$. Then the quantity   
\begin{equation}
t = \frac{x-\mu}{\sigma^2/\sqrt{n}}
\label{eq:t-dist}
\end{equation}
where $x \in (-\infty,\infty)$
has a $t$-distribution function with $n-1$ degrees of freedom for univariate case. The $t$-distribution density curves defined by (\ref{eq:t-dist}) are symmetric and bell-shaped in nature like the normal distribution (Fig. \ref{fig:t-dist}(a)). However, the advantage lies in the spread of the distribution and is more for $t$-density function than that of standard normal distribution. In multivariate case the distribution consists of more than one random variable. For two random variables this is known as a bivariate distribution, but the concept generalizes to any number of random variables, giving a multivariate distribution. Since the input for visual servoing is a two-dimensional image feature, the representing distribution will also be multivariate in nature. With the concept of multivariate normal distribution a multivariate student $t$-distribution can be easily modeled as follows.

Consider a $d$-dimensional random variable $\boldsymbol{x}$ following a multivariate normal distribution with mean $\boldsymbol{\mu}$ and positive definite, symmetric and
real $d \times d$ covariance matrix $\boldsymbol{\Sigma}$. The corresponding density function is represented by 
\begin{equation}
    N(\boldsymbol{x};\boldsymbol{\mu},\boldsymbol{\Sigma}) = \frac{1}{(2\pi)^{\frac{d}{2}}\boldsymbol{\Sigma}^{\frac{1}{2}}}exp\left \{ -\frac{1}{2}(\boldsymbol{x}-\boldsymbol{\mu})^T \boldsymbol{\Sigma}^{-1}(\boldsymbol{x}-\boldsymbol{\mu}) \right \}.
\end{equation} 
To produce a longer tail to normal distribution, two normal distributions are added, which is represented by
\begin{equation}
(1-\epsilon)N(\boldsymbol{x};\boldsymbol{\mu},\boldsymbol{\Sigma})+\epsilon N(\boldsymbol{x};\boldsymbol{\mu},c\boldsymbol{\Sigma}),
\label{eq:two_normal}
\end{equation}
where $c$ is large and $\epsilon$ is small. Equation (\ref{eq:two_normal}) can be represented in integral form as
\begin{equation}
\int N(\boldsymbol{x};\boldsymbol{\mu},\boldsymbol{\Sigma}/u)\ dH(u), 
\end{equation}
where $H$ is a distribution function that places $(1-\epsilon)$ at point $u=1$ and $\epsilon$ at the point $u=1/c$. The probability function $H$ can be represented by a chi-squared function having $\nu$ degree-of-freedom with random variable $K$ such that
\begin{equation}
K \sim gamma(\nu/2,\nu/2).
\end{equation}
The $gamma(a,b)$ has density function
\begin{equation}
f(u,a,b)= \frac{b^a u^{a-1}}{\Gamma (a)}exp\left ( -bu \right )\boldsymbol{I}_{0,\infty}(u); \ \ (a,b>0)
\end{equation}
where $\boldsymbol{I}_{0,\infty}(u)=1$ for $u>0$ and zero elsewhere.
From the above equations it can be summarized as, for the same random variable $\boldsymbol{x}$ a multivariate $t$-distribution with $\nu$ degrees-of-freedom can be written as 
\begin{equation}
\phi(\boldsymbol{x};\boldsymbol{\mu},\boldsymbol{\Sigma},\nu) \sim N(\boldsymbol{x};\boldsymbol{\mu},\boldsymbol{\Sigma}/u) 
\end{equation}
where 
\begin{equation}
u \sim gamma(\nu/2,\nu/2).
\end{equation}
Then a multivariate $t$-distribution density function \cite{peel2000robust} at a point $\boldsymbol{x}$ can be represented by
\begin{equation}
\phi(\boldsymbol{x};\boldsymbol{\mu},\boldsymbol{\Sigma},\nu) = \frac{\Gamma(\frac{\nu+d}{2})\mid\boldsymbol{\Sigma} \mid^{-\frac{1}{2}}}{(\pi \nu)^{\frac {d}{2}}\Gamma(\frac {\nu}{2})[1+\nu^{-1}\delta(\boldsymbol{x},\boldsymbol{\mu};\boldsymbol{\Sigma})]^{\frac{\nu+d}{2}} }
\label{eq:mul_t_dist}
\end{equation}
where the parameter $\Gamma$ represents the Gamma function. Here the parameter $\nu$ provides a heavy-tailed  alternative to the normal distribution family (Fig. \ref{fig:t-dist}) and as $\nu$ tends to infinity, the $t$-distribution approaches the normal distribution. Hence, this parameter $\nu$ may be viewed as a robustness tuning parameter. It can be fixed in advance or it can be inferred from the
data for each component \cite{lange1989robust} based on different applications.
The Mahalanobis distance $\delta(\boldsymbol{x},\boldsymbol{\mu};\boldsymbol{\Sigma}) = (\boldsymbol{x}-\boldsymbol{\mu})^T\boldsymbol{\Sigma}(\boldsymbol{x}-\boldsymbol{\mu})$ denotes squared distance between $\boldsymbol{x}$ and $\boldsymbol{\mu}$. In the proposed novel methodology for visual servoing, student $t$-distribution functions are used to model distributions of an image instead of directly taking image intensities. 
Finding an analytical representation of tuning parameter $\boldsymbol{\mu}$ from image attributes, which controls the spread of the distribution is a main challenge and is one of the major contribution of proposed approach. An overview of the proposed approach of servoing is given in the following section.
\begin{figure*}[t]
\centering
\includegraphics[width=4.8in]{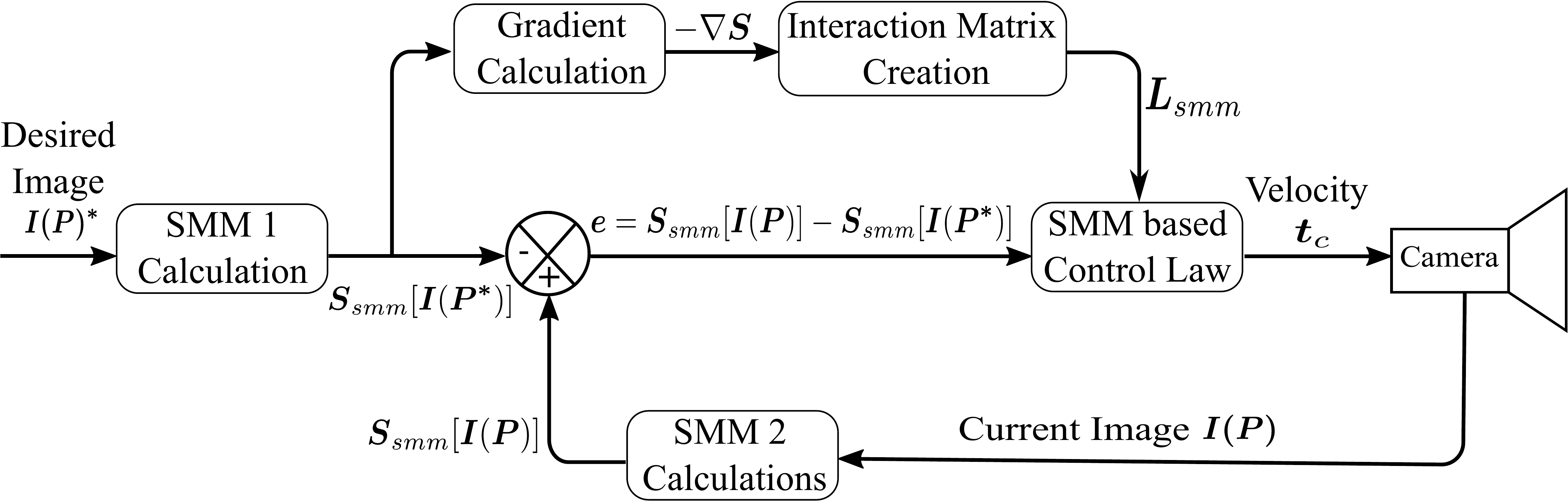}
\caption{Block diagram for SMM based visual servoing in image space: The visual servoing control law which is derived under the framework of student $t$-distribution mixture models minimizes the error $\boldsymbol{S}_{smm}[\boldsymbol{I(p)}]-\boldsymbol{S}_{smm}[\boldsymbol{I(p)^*}]$ by providing appropriate velocity to the camera.}
\label{fig:BD}
\end{figure*}
\section{Outline of proposed Methodology}\label{sec:outline}
From the literatures on visual servoing it can be observed that there is a paradigm shift from using point features to entire image to generate control signals. 
Using entire photometric image information as visual feature gives advantage of avoiding feature detection, matching or tracking steps. On the other hand, the use of image intensity values to generate control signals are sensitive to illumination variations and suffers from smaller convergence domains. The idea of using the probability distributions, which consider the effect of association between random variables, can be used to overcome these disadvantages. The successful implementation of similar concepts can be seen in other applications like image registration \cite{gerogiannis2007robust}. This becomes motivation for the proposed SMM based visual servoing. The proposed idea is to replace each image intensity by a student $t$-distribution. The heavily tailed distribution function of $t$-distribution helps to consider the effect of neighbourhood values. By modelling each image pixels by $t$-distribution the image under consideration can be represented by a mixture of distributions. This mixture model is used to design the proposed control law.

The proposed framework has been developed to control the motion of camera to reach a desired pose as shown in Fig. \ref{fig:smm_teaser} by considering whole image data ($N \times M$) as input. Fig. \ref{fig:BD} provides an overview of the proposed methodology. The algorithm is initiated by feeding the desired image $\boldsymbol{I(p)^*} \in R^k$ and the initial image $\boldsymbol{I(p)} \in R^k$ to SMM calculation modules 1 and 2, where $k = N \times M$. Here the current image is used as a feedback signal to the control law. The probabilistic models of desired image $\boldsymbol{S}_{smm}[\boldsymbol{I(p)^*}] \in R^k$ and current image $\boldsymbol{S}_{smm}[\boldsymbol{I(p)}]\in R^k$ are calculated using multivariate $t$-distributions  and produced as an output from SMM calculation modules. It may be noted that here every pixel in the image is treated as feature point. In the proposed method the visual servoing operation is posed as a minimization problem in which the error to be minimized is $\boldsymbol{S}_{smm}[\boldsymbol{I(p)}]-\boldsymbol{S}_{smm}[\boldsymbol{I(p)^*}]$.
Once the error is modeled, next the interaction matrix has to be calculated using image feature points. Besides conventional methods, in SMM based approach the interaction matrix is created using SMM of image. For this, the obtained SMM of desired image is passed through a gradient calculation module. The gradient of SMM image can be obtained by using a gradient filter over $\boldsymbol{S}_{smm}[\boldsymbol{I(p)^*}]$ or by taking gradient of SMM analytically. Since $t$-distribution is differentiable, analytical expressions can also be obtained easily. The obtained gradient is then used to calculate the interaction matrix $\boldsymbol{L}_{smm}\in R^{k\times6}$. 

Since entire image is being used, interaction matrix calculation for each iterations may take more time. Therefore in the proposed method the interaction matrix is calculated only once using desired image and used for the entire iterations. It has been shown in \cite{chaumette1998potential} that the closed loop system will remain stable even if the interaction matrix computed from the desired image is used. Next the SMM based control law module generate velocities to move the camera and generate new current image. The control law minimizes the error using an optimization framework. Eventually, the camera velocity become zero once the camera reaches the desired position which constitutes end of visual servoing operation. The SMM of an image and derivation of interaction matrix are detailed in the following sections.  

\section{Image modelling and Interaction matrix}\label{sec:img_model}
Instead of directly considering pixels of an image as features the proposed method represent it as a distribution function. This takes into account the influence of neighbouring pixels. Hence, it is more suitable for the applications in real world environments in which features can not be extracted and tracked easily. The first part of this section explains statistical modelling of image using multivariate $t$-distribution. The next part presents an analytical formulation of interaction matrix for SMM based visual servoing.     
\subsection{Image representation using SMM}
A student mixture model is a combination of $t$-distribution functions at several locations. The $t$-distribution density function (multivariate case) at a point $\boldsymbol{X}=(x,y)$ can be represented by \eqref{eq:mul_t_dist} having parameters $\boldsymbol{\mu},\boldsymbol{\Sigma},\nu$. Using this formulation, an image can be modelled as a combination of multiple $t$-distribution functions given by 
\begin{equation}
    \boldsymbol{S}_{smm}(\boldsymbol{I}) = \sum_{i=1}^{n} \pi_i \phi(\boldsymbol{X};\boldsymbol{\mu}_i,\boldsymbol{\Sigma}_i,\nu_i),
\label{eq:gen_smm}
\end{equation}
where the mixing proportions $\pi_i$ are non-negative and sum upto unity. The quantity $n$ is the total number of pixel locations in the image.
Here it should be noted that each pixel in the image contributes one component $t$-distribution to the SMM model. In general image processing application scenarios the mixture models are usually estimated from the image features in interest. Unlike in the general probabilistic mixture modelling paradigm, in the proposed method the mean of each multivariate $t$-distribution $\boldsymbol{\mu}_i \in R^{1\times2}$ is equal to its observed pixel position $\boldsymbol{X}_i=(x,y)$ and the covariance matrix $\boldsymbol{\Sigma}_i\in R^{2\times2}$ is proportional to the pixel intensity $\boldsymbol{I}_i$ at $\boldsymbol{X}_i$. 

Finding the tuning parameter $\nu$, which controls the spread of the distribution is main challenge and is one of the major part of our contribution. In order to obtain the value of each $\nu_i$ for each $t$-distribution, one of the basic property of $t$-distributions is considered. This property gives a relation between the variance and degree-of-freedom as given below:
\begin{equation}
\sigma(t_{dist}) = \frac{\nu}{\nu-2}
\label{eq:var_dof}
\end{equation} 
Since the image pixel intensity is used as variance $(\sigma = I_i)$  for modeling SMM of an image, using \eqref{eq:var_dof} the values of $\nu_i$ can be obtained from the following relation
\begin{equation}
\nu_i = \frac{2\sigma_i}{(\sigma_i-1)}.
\label{eq:inten_to_dof}
\end{equation}
\begin{figure*}[t]
\centering
\subfloat[]{\includegraphics[width=2.3in]{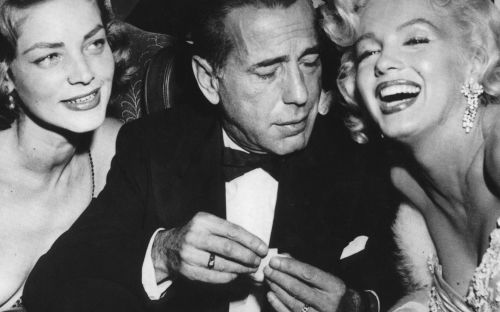}}
\hspace{0.01cm}
\subfloat[]{\includegraphics[width=2.3in]{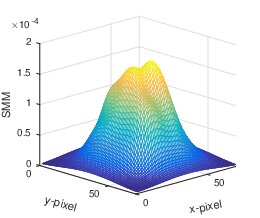}}
\caption{Image modeling by SMM: (a) Input image (b) SMM of the image. Each pixel location in the image contributes mean for $t$-distribution and intensity at that position for covariance matrix. The aggregation of all such $t$-distributions constitute student mixture model for the image }
    \label{fig:smm_img_rep}
\end{figure*}

This extra parameter $\nu$, for modeling distribution (SMM) yields more smoothness to the density function with a probability density function having heavier tail. Using student $t$ -distribution for modeling image is first of its kind to be used in visual servoing context. Fig. \ref{fig:smm_img_rep} shows an example of an image modeled by SMM using \eqref{eq:mul_t_dist} and \eqref{eq:gen_smm}. Each pixel locations of the given image Fig. \ref{fig:smm_img_rep}(a) serves as mean for its SMM (Fig. \ref{fig:smm_img_rep}(b)) and intensity values as variance. The degree-of-freedom $\nu$ for each student $t$-distribution in SMM is calculated using the above formulation which relates degree-of-freedom of distribution to image pixel intensity given in \eqref{eq:inten_to_dof}. From the SMM function modeled for the given image it can be observed that the surface is smooth and continuous, and inherently considers feature
position uncertainties.

Once the SMM is modeled for current and desired images, an error between these two distributions are generated and applied to visual servo controller. The following section explains the detailed development of interaction matrix for SMM based visual servoing.
\subsection{Development of interaction matrix}
Once the feature models are obtained, next task is to compute the interaction matrix. The interaction matrix $(\boldsymbol{L})$ maps feature rate $(\boldsymbol{\dot{s}})$ into camera velocity $(\boldsymbol{t}_c)$, i.e., 
\begin{equation}
\boldsymbol{\dot{s}}=\boldsymbol{L} \boldsymbol{t}_c.
\end{equation} 
Since all the pixels in the image are considered while development of SMM, the interaction matrix is obtained by arranging respective interaction matrices at locations $\boldsymbol{x}$ as given below.
\begin{equation}
\boldsymbol{L}_{smm} = \begin{bmatrix}
\vdots  \\
\boldsymbol{L}_{smm}(\boldsymbol{x})  \\
\vdots 
\end{bmatrix}.
\label{eq:L_matrix_stack}
\end{equation}
It can be deduce from \eqref{eq:gen_smm} that $\boldsymbol{L}_{smm}(\boldsymbol{x} )$ is the sum of every interaction matrices related to each student $t$-distribution expressed
from all $i$ locations to location $\boldsymbol{x}$ as given below
\begin{equation}
\boldsymbol{L}_{smm}(\boldsymbol{x}) = \sum_{i=1}^{n} \boldsymbol{L}_{smm}(\boldsymbol{x},i).
\end{equation}
In order to obtain individual $\boldsymbol{L}_{smm}(\boldsymbol{x},i)$, it is required to find the mapping between the time derivative of SMM $(\boldsymbol{\dot{S}})$ and 2D image pixel velocity $(\boldsymbol{\dot{x}})$. By using similar concept from temporal luminance constancy equation for optical flow \cite{horn1981determining}, we can write 
\begin{equation}
\boldsymbol{S}(\boldsymbol{x}+\delta \boldsymbol{x},t+\delta t) = \boldsymbol{S}(\boldsymbol{x},t).
\label{eq:optical_flow}
\end{equation}
Here assumption is made that SMM image also follow temporal luminance constancy hypothesis. First order Taylor series expansion on \eqref{eq:optical_flow} provides
\begin{equation}
\begin{split}
\boldsymbol{S}(\boldsymbol{x},t) = \boldsymbol{S}(\boldsymbol{x}+\delta \boldsymbol{x},t+\delta t) \ \ \ \  \\+\frac{\partial \boldsymbol{S}(\boldsymbol{x}+\delta \boldsymbol{x},t+\delta t)}{\partial \boldsymbol{x}} \frac{\mathrm{d} \boldsymbol{x}}{\mathrm{d} t}\\+\frac{\partial \boldsymbol{S}(\boldsymbol{x}+\delta \boldsymbol{x},t+\delta t)}{\partial t} \frac{\mathrm{d} t}{\mathrm{d} t}
\end{split},
\end{equation}
and we get
\begin{equation}
\frac{\partial \boldsymbol{S}(\boldsymbol{x}+\delta \boldsymbol{x},t+\delta t)}{\partial \boldsymbol{x}} \frac{\mathrm{d} \boldsymbol{x}}{\mathrm{d} t}+\frac{\partial S(\boldsymbol{x}+\delta \boldsymbol{x},t+\delta t)}{\partial t}=0.
\end{equation}
In simple terms the above formula can be written into the following form 
\begin{equation}
\nabla \boldsymbol{S}^T \boldsymbol{\dot{x}} + \boldsymbol{\dot{S}} = 0 
\label{eq:opti_flow_new}
\end{equation}
Upon rearranging \eqref{eq:opti_flow_new} an equation similar to optical flow constraint equation is obtained which is given by
\begin{equation}
\boldsymbol{\dot{S}} = - \nabla \boldsymbol{S}^T \boldsymbol{\dot{x}}.
\label{eq:opti_flow_constr}
\end{equation}
It is well known that the image pixel velocity $\boldsymbol{\dot{x}}$ is related to the camera velocity $\boldsymbol{t}_c$ \cite{chaumette2006visual} in the following manner
\begin{equation}
\boldsymbol{\dot{x}} = \boldsymbol{L}_x \boldsymbol{t}_c,
\label{eq:pixel_vel}
\end{equation}
where $\boldsymbol{L}_x\in R^{2\times6}$ is known as basic interaction matrix related to pixel locations $x$ . Using \eqref{eq:pixel_vel} in \eqref{eq:opti_flow_constr} gives the relation between $\boldsymbol{\dot{S}}$ and $\boldsymbol{t}_c$, from which the interaction matrix 
\begin{equation}
\boldsymbol{L}_{smm}(\boldsymbol{x},i) = - \nabla \boldsymbol{S}^T \boldsymbol{L}_x
\end{equation}   
where
\begin{equation}
\nabla \boldsymbol{S} = \begin{bmatrix}
\nabla S_u  \\
\nabla S_v
\end{bmatrix}.
\end{equation}
is obtained. To compute the gradients $\nabla \boldsymbol{S}\in R^{1\times2}$, methods similar to horizontal and vertical image gradient calculations or analytical formulations can be used.
\section{SMM based visual servoing}\label{sec:smm_vs}
The aim of a vision-based control schemes is to minimize an
error which is given in \eqref{eq:error}. In the proposed method the error function is modeled as difference between current and desired SMMs and, visual servoing is considered as an optimization problem. The error between current and desired SMMs act as a cost function to be minimized for optimization. This section presents the underlying optimization frame work for SMM based visual servoing. 

For a robot, the end effector pose $\boldsymbol{P}$ can be represented by rigid transformation $\boldsymbol{[R, T ]}$ with respect to world coordinate system. Here, $\boldsymbol{R}$ is a $3\times3$ rotation matrix and $\boldsymbol{T} = [t_x \ t_y \ t_z ]^T $ is the $3\times1$ translation vector. The rotation matrix $\boldsymbol{R}$ can be parametrized in terms of $3$ Euler
angles $(\alpha, \beta \ and \ \gamma)$  such that $\boldsymbol{R} = \boldsymbol{R}_x(\alpha) \boldsymbol{R}_y (\beta) \boldsymbol{R}_z (\gamma)$. In visual servoing it is required to move the robot end-effector from an initial pose $\boldsymbol{P}  \in  \boldsymbol{R}^3 \times SO(3)$ to reach a desired pose $\boldsymbol{P^*}$ . To achieve this goal, a cost function $f()$ has to be defined. Most of the time this cost function is an error which needs to be minimized. Thus, a visual-servoing problem can be written as an optimization problem given by
\begin{equation}
 \hat{\boldsymbol{P}} = \operatornamewithlimits{argmin}_{\boldsymbol{P}} f(\boldsymbol{P},\boldsymbol{P^*}) 
\end{equation}
where $\hat{\boldsymbol{P}}$ is the pose closest possible to $\boldsymbol{P^*}$, reached after the optimization. In classical visual-servoing, the cost function is defined as the distance between geometrical features extracted from the image, and the corresponding minimization is
\begin{equation}
 \hat{\boldsymbol{P}} = \operatornamewithlimits{argmin}_{\boldsymbol{P}} ||(\boldsymbol{s}(\boldsymbol{P})-\boldsymbol{s}(\boldsymbol{P^*}))||. 
\end{equation}
For direct approaches like Kernel and Photometric Visual Servoing $f()$ is the sum of squared differences of the image intensities as shown below.
\begin{equation}
 \hat{\boldsymbol{P}} = \operatornamewithlimits{argmin}_{\boldsymbol{P}} ||(\boldsymbol{I}(\boldsymbol{P})-\boldsymbol{I}(\boldsymbol{P^*}))|| 
\end{equation}
\begin{figure}[!htp]
\centering
\subfloat[]{\includegraphics[width=0.5in]{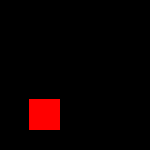}}
\hspace{1.1cm}
\subfloat[]{\includegraphics[width=2.5in]{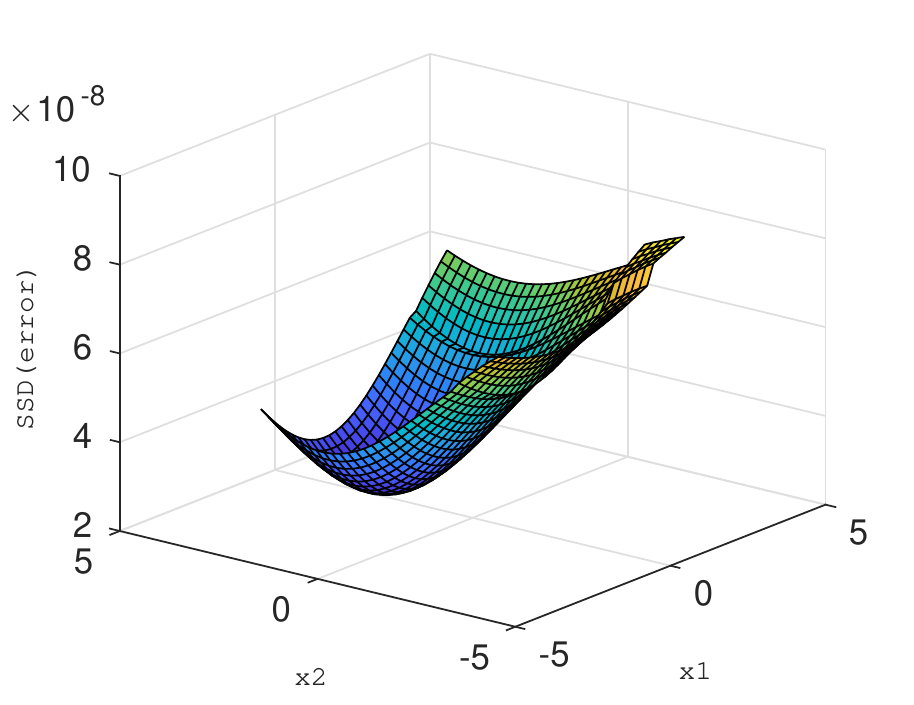}}\\
\subfloat[]{\includegraphics[width=0.5in]{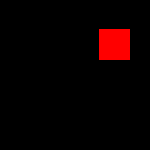}}
\hspace{1.1cm}
\subfloat[]{\includegraphics[width=2.5in]{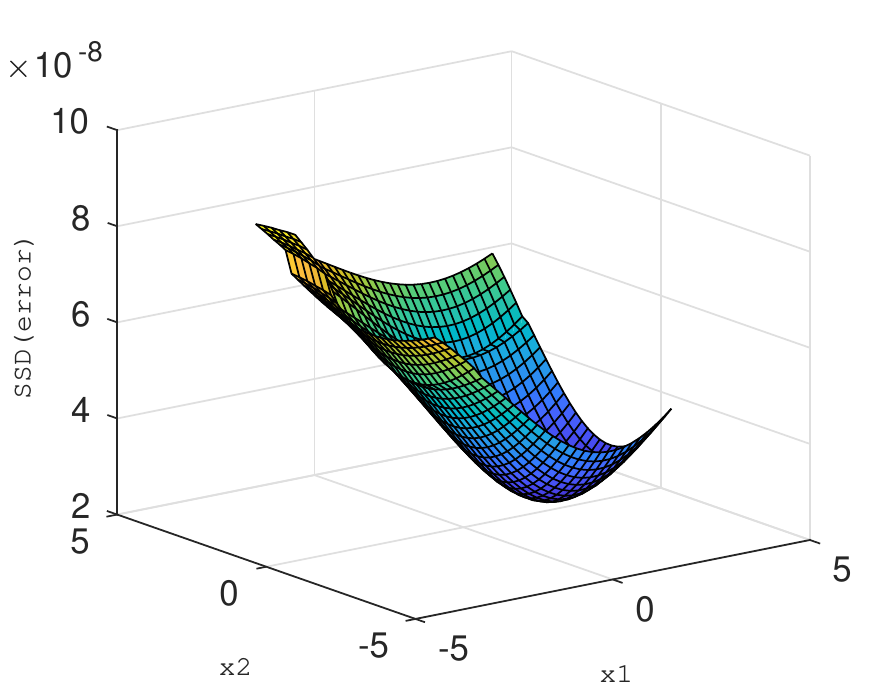}}\\
\subfloat[]{\includegraphics[width=0.5in]{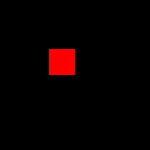}}
\hspace{1.1cm}
\subfloat[]{\includegraphics[width=2.5in]{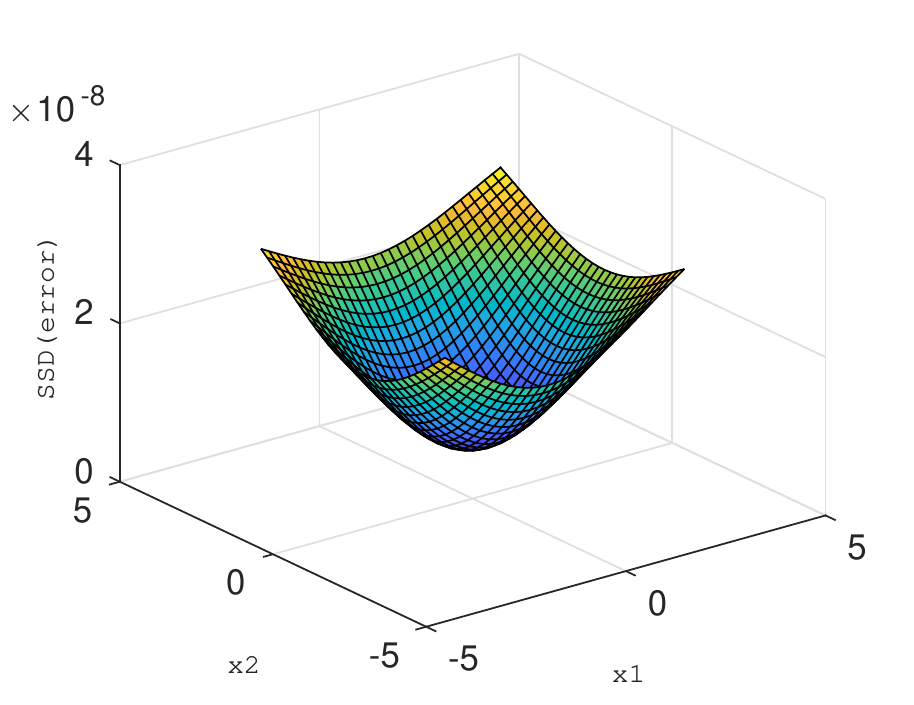}}
\caption{Cost functions for different test images having different point locations using SMM approach. It can be observed that the SMM based
cost functions are smooth and differentiable.  This makes the optimization framework to work easily and guarantees convergence to servoing operation.}
    \label{fig:cost_func}
\end{figure}
It must be noted that the proposed method also uses the entire photometric information
(\ref{eq:gen_smm}) provided by the images as in the other dense visual servoing approaches. Therefore in the proposed framework the objective is to regulate the following error function to zero: 
\begin{equation}
\boldsymbol{e}(\boldsymbol{P}) = \boldsymbol{S}_{smm}[\boldsymbol{I}(\boldsymbol{P})]- \boldsymbol{S}_{smm}[\boldsymbol{I}(\boldsymbol{P^*})],
\label{eq:smm_error}
\end{equation}
where term $\boldsymbol{S}_{smm}[ \ ]$ is calculated using \eqref{eq:gen_smm} for current and desired images. In other words, the problem is to minimize an error vector $\boldsymbol{e}(\boldsymbol{P})$ of visual features $\boldsymbol{S}_{smm}(\boldsymbol{P})$ extracted from the initial/current image by finding a vector  $\boldsymbol{t}_c$ that incrementally minimizes a cost function $E(\boldsymbol{S}_{smm}(\boldsymbol{P}))$. 
Viewing the problem as a non-linear least squares minimization, the cost function is reformulated as
\begin{equation}
E(\tilde{\boldsymbol{S}}(\boldsymbol{P})) = \frac{1}{2} [\tilde{\boldsymbol{S}}(\boldsymbol{P})- \tilde{\boldsymbol{S}}(\boldsymbol{P^*})]^T [\tilde{\boldsymbol{S}}(\boldsymbol{P})- \tilde{\boldsymbol{S}}(\boldsymbol{P^*})]
\label{eq:smm_cost}
\end{equation}
where $\tilde{\boldsymbol{S}}$ represents $\boldsymbol{S}_{smm}(\boldsymbol{I}())$.

Fig. \ref{fig:cost_func} depicts various cost functions for 2D cases with different point images. It was observed that the SMM based cost functions are smooth and also makes the surface differentiable without using any extension parameters as required for GMM \cite{crombez2015photometric}. This makes the modeling easier and guarantees convergence to servoing operation.  

Since first order optimization methods are known to be slow, methods that are based on the second order Taylor series approximation of the cost function like Gauss-Newton method are considered here. With the proposed novel error function \eqref{eq:smm_error} in-terms of SMM and interaction matrix in (18)
the required camera velocity is obtained as
\begin{equation}
\boldsymbol{t}_c = -\lambda \boldsymbol{L}_{smm}^+ \boldsymbol{S}_{smm}[\boldsymbol{I}(\boldsymbol{P})]- \boldsymbol{S}_{smm} [\boldsymbol{I}(\boldsymbol{P^*})] 
\end{equation}
where $\lambda$ is a positive scalar and $\boldsymbol{L}_{smm}^+$ is the pseudo inverse of SMM based interaction matrix given in \eqref{eq:L_matrix_stack}. 

It will be shown in the following section that the above controller results in (a) exponential decrease of error function and (b) convergence of camera and joint velocities are regulated to zero at the desired pose. 
\begin{figure}[t]
\centering
\subfloat[]{\includegraphics[width=2in]{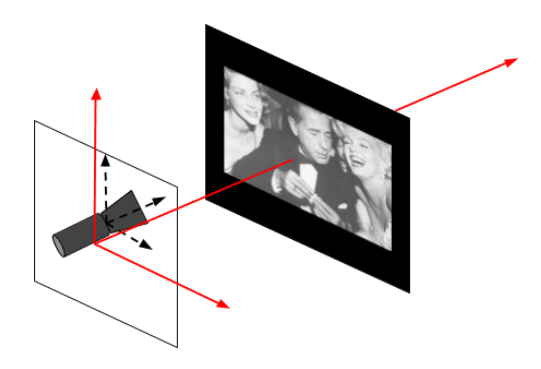}}
\subfloat[]{\includegraphics[width=3in]{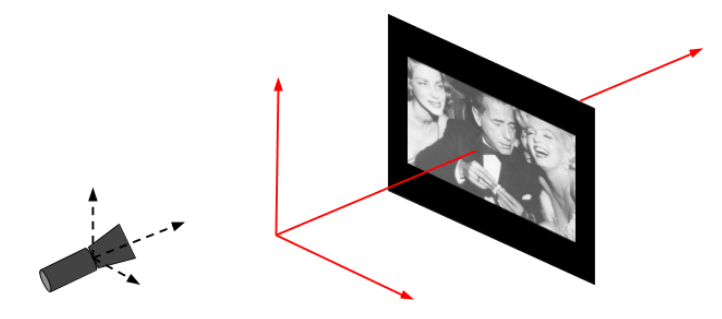}}\\
\subfloat[]{\includegraphics[width=2in]{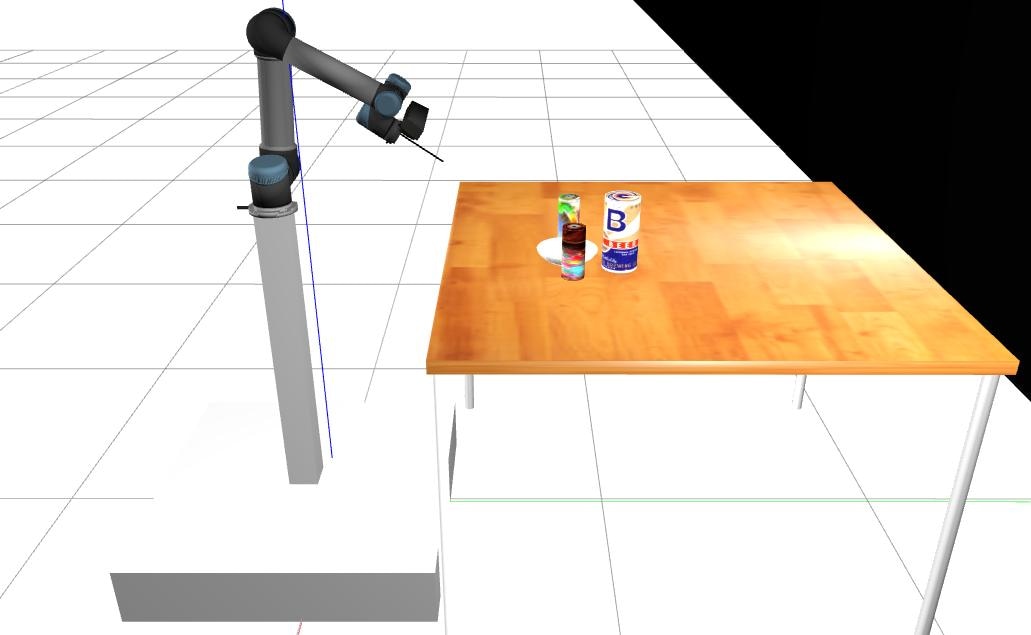}}
\caption{SMM based visual servoing simulation environment:(a) 2D positioning task with planar camera motion (b) 3D positioning task using 6-DOF camera motion(c) Realistic environment in Gazebo with UR5 manipulator and 3D objects.}
    \label{fig:sim_env}
\end{figure} 
\section{Results and Discussion}\label{sec:result}
This section discusses simulation results of visual servoing using the proposed framework. An eye in hand system is considered and the camera reaching desired pose is considered as convergence of an experiment. Number of simulated experiments are performed to decipher performance of the proposed methodology. Initial part of this section shows the behaviour of the proposed method by simple 2D (Fig. \ref{fig:sim_env}a) and 3D (Fig. \ref{fig:sim_env}b) camera movements which is servoing towards a planar image object. For verifying the effectiveness of the proposed algorithm in real scenario, visual servoing experiments are carried out   in Gazebo environment with using UR5 manipulator and 3D objects  as shown in Fig. \ref{fig:sim_env}(c). Finally to show the robustness of the framework, its performance under different image sizes, image contents and occlusions are investigated.
\begin{table}[t]
\caption{Test cases for $2D$ positioning task with fixed depth and desired position at $(t_x,t_y,\theta_z) = (0,0,0)$. Camera positioning error are shown for the start and end of SMM visual servoing}
\vspace{-.3cm}
\centering
\begin{adjustbox}{max width=6cm}
\begin{tabular}{|l|ccc|ccc|}
\hline
 & \multicolumn{3}{c|}{Initial error} & \multicolumn{3}{c|}{Final error} \\ 
\hline 
Exp. & $t_x(m)$ & $t_y(m)$ & $\theta_z$ & $t_x(m)$ & $t_y(m)$ & $\theta_z$ \\ \hline \hline
$1$ & $0.3$ & $0.3$ & $0^{\circ}$ & $0.0021$ & $0.0083$ & $0^{\circ}$ \\
$2$ & $0.25$ & $0.25$ & $-15^{\circ}$ & $-0.0002$ & $0.0050$ & $-0.0892^{\circ}$ \\
$3$ & $0.25$ & $0.25$ & $10^{\circ}$ & $0.0007$ & $0.0058$ & $-0.01^{\circ}$ \\
$4$ & $-0.3$ & $0.25$ & $-18^{\circ}$ & $-0.0012$ & $0.0077$ & $-0.0081^{\circ}$ \\ 
$5$ & $0.4$ & $0.35$ & $5^{\circ}$ & $0.0015$ & $0.0076$ & $0.054^{\circ}$ \\\hline 
\end{tabular}
\end{adjustbox}
\vspace{-.3cm}

\label{tb:2D_pos}
\end{table}
\begin{figure}[]
\centering
\subfloat[Initial image]{\includegraphics[width=1.2in]{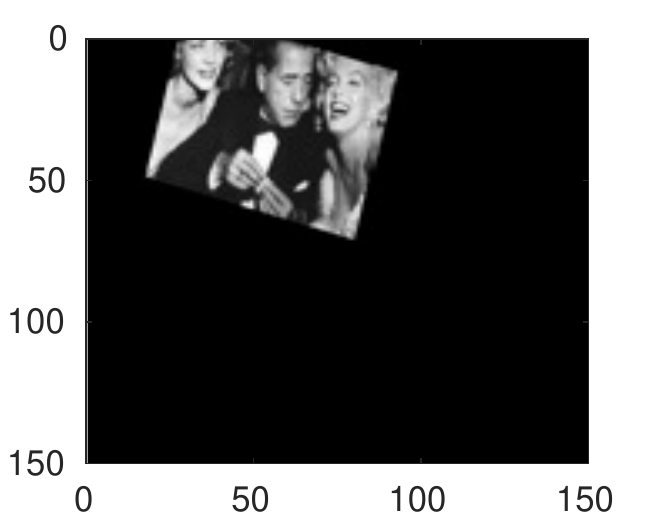}}
\subfloat[Final image]{\includegraphics[width=1.2in]{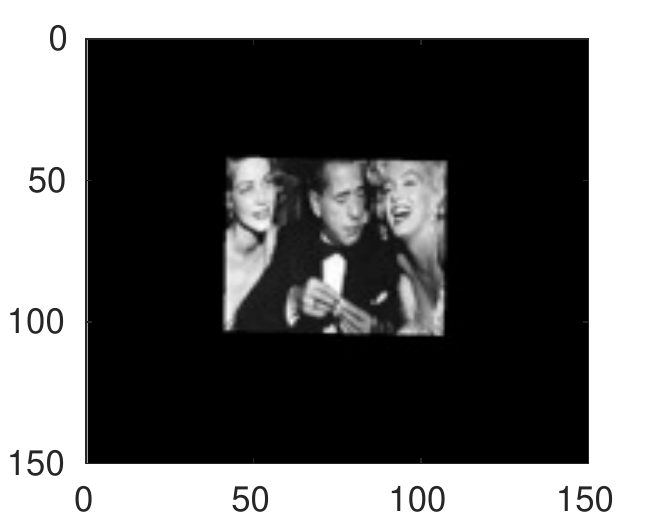}}
\subfloat[Camera velocities]{\includegraphics[width=1.2in]{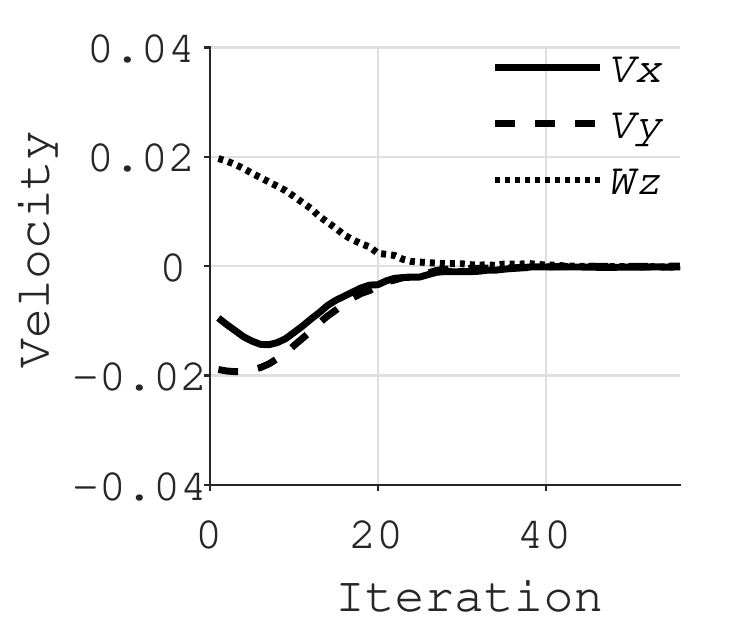}}
\subfloat[Error]{\includegraphics[width=1.2in]{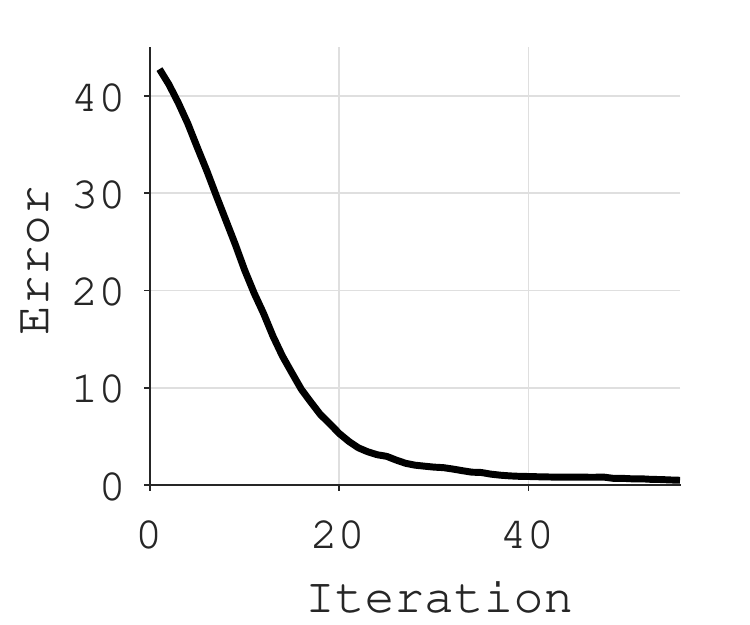}}\\
\vspace{-.3cm}
    \caption{One of the test case for $2D$ SMM visual servoing.}
    \label{fig:smm_vs_2D}
\end{figure}
\subsection{$2D$ and $3D$ positioning task}
First a 2D case was considered where the objective is to control linear and rotational motion $(v_x,v_y,\omega_z)$ along a plane (see Fig. \ref{fig:sim_env}(a)) exhibited by a 3-DOF camera to reach the desired pose. In this experiment a planar image is considered in the workspace and desired camera position is taken as $(t_x,t_y,\theta_z) = (0,0,0)$. To evaluate the validity of the proposed framework different initial poses have been chosen as shown in Table \ref{tb:2D_pos}. The end of the simulation is considered when the  norm of error, pertaining to \eqref{eq:smm_cost}, approaches to zero. From the error in final pose  shown in Table \ref{tb:2D_pos}, it can be concluded that the motion exhibited by proposed framework made the camera to successfully reach the desired pose with negligible errors. The initial image, final image, camera velocities and error plots are shown in Fig. \ref{fig:smm_vs_2D} for Experiment 3 in Table \ref{tb:2D_pos}  as an illustration. It can be seen that the velocities decay to zero and the norm of error shows the convergence of cost function to minimum. 
\begin{table}[t]
\caption{Test cases for $3D$ positioning task with 6-DOF camera for reaching desired pose 
$(p_x, p_y, p_z, \theta_x, \theta_y, \theta_z) = (0,0,0,0,0,0)$. Camera positioning error are shown for the start and end of SMM visual servoing}
\vspace{-.3cm}
\centering
\begin{adjustbox}{max width=\textwidth}
\begin{tabular}{|l|cccccc|cccccc|}
\hline
 & \multicolumn{6}{c|}{Initial error} & \multicolumn{6}{c|}{Final error} \\ 
\hline 
Exp. & $p_x(m)$ & $p_y(m)$ &$p_z(m)$ & $\theta_x$ &$\theta_y$ & $\theta_z$ & $p_x(m)$ & $p_y(m)$ &$p_z(m)$ & $\theta_x$ &$\theta_y$ & $\theta_z$ \\ \hline \hline
$1$ & $-0.4$ & $-0.4$ & $-0.50$ &$0.5^{\circ}$ &$1^{\circ}$ &$-10^{\circ}$ & $-0.0071$ & $-0.1220$ & $-0.0500$ & $0.0156^{\circ}$ &$0.0023^{\circ}$&$-1.8307^{\circ}$\\
$2$ & $-0.4$ & $-0.4$ & $-0.53$ &$1^{\circ}$ &$0.3^{\circ}$ &$-20^{\circ}$ & $-0.1193$ & $-0.0035$ & $-0.0013$ & $0.0089^{\circ}$ &$0.0171^{\circ}$&$-1.1296^{\circ}$\\
$3$ & $0.36$ & $0.38$ & $-0.48$ &$0^{\circ}$ &$0.3^{\circ}$ &$-12^{\circ}$ & $0.0286$ & $0.0249$ & $-0.0028$ & $0.0269^{\circ}$ &$0.0311^{\circ}$&$-0.6179^{\circ}$\\
$4$ & $0.35$ & $0.37$ & $-0.55$ &$0^{\circ}$ &$0^{\circ}$ &$8^{\circ}$ & $0.0210$ & $0.0322$ & $-0.0012$ & $-0.0080^{\circ}$ &$0.0046^{\circ}$&$0.1668^{\circ}$\\
$5$ & $0.22$ & $-0.32$ & $-0.51$ &$0.1^{\circ}$ &$1.2^{\circ}$ &$20^{\circ}$ & $0.0274$ & $0.0231$ & $-0.0164$ & $0.0885^{\circ}$ &$0.7329^{\circ}$&$0.4003^{\circ}$\\
$6$ & $0.37$ & $0.42$ & $-0.46$ &$0.8^{\circ}$ &$0.5^{\circ}$ &$8^{\circ}$ & $0.0204$ & $0.0315$ & $-0.0050$ & $0.1583^{\circ}$ &$0.0519^{\circ}$&$0.9347^{\circ}$\\
$7$ & $0.30$ & $-0.45$ & $-0.48$ &$1^{\circ}$ &$0.5^{\circ}$ &$-20^{\circ}$ & $0.0274$ & $-0.0125$ & $-0.0081$ & $0.4828^{\circ}$ &$0.1682^{\circ}$&$-0.8595^{\circ}$\\
$8$ & $-0.37$ & $0.28$ & $-0.52$ &$-0.6^{\circ}$ &$-0.1^{\circ}$ &$-20^{\circ}$ & $-0.0123$ & $0.0352$ & $-0.0021$ & $-0.1498^{\circ}$ &$-0.0718^{\circ}$&$0.0590^{\circ}$\\
$9$ & $0.35$ & $0.38$ & $-0.53$ &$0.4^{\circ}$ &$0.3^{\circ}$ &$-20^{\circ}$ & $0.0181$ & $0.0136$ & $-0.0210$ & $-0.0602^{\circ}$ &$0.0348^{\circ}$&$-1.3683^{\circ}$\\
$10$ & $0.27$ & $-0.35$ & $-0.49$ &$0.4^{\circ}$ &$1^{\circ}$ &$14^{\circ}$ & $0.0112$ & $0.0833$ & $-0.0089$ & $0.0951^{\circ}$ &$0.0743^{\circ}$&$0.9961^{\circ}$\\
\hline 
\end{tabular}
\end{adjustbox}
\vspace{-.3cm}

\label{tb:exp_3D}
\end{table}
\begin{figure}[!htp]
\centering
\subfloat[Initial image]{\includegraphics[width=1.5in]{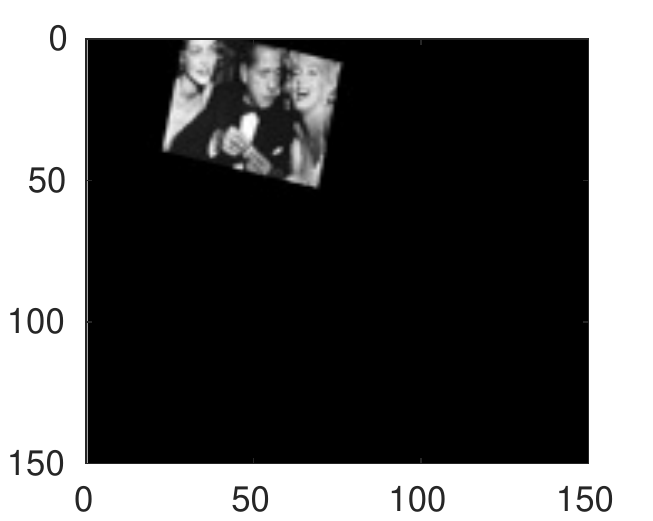}}
\hspace{0.01cm}
\subfloat[Final image]{\includegraphics[width=1.5in]{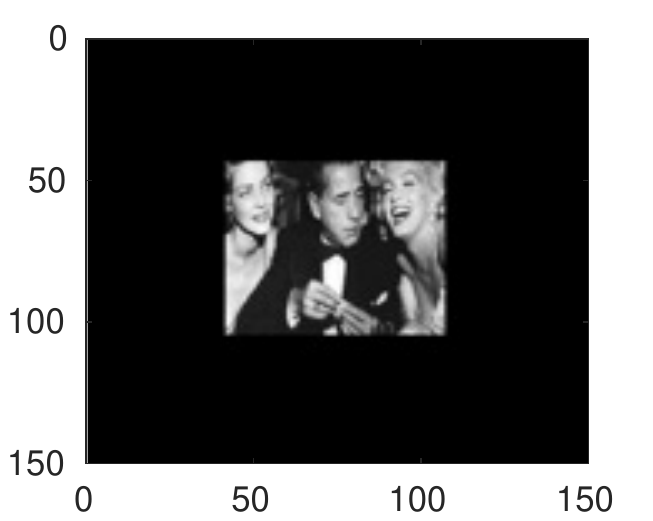}}
\hspace{0.01cm}
\subfloat[$I-I^*$ at final pose]{\includegraphics[width=1.5in]{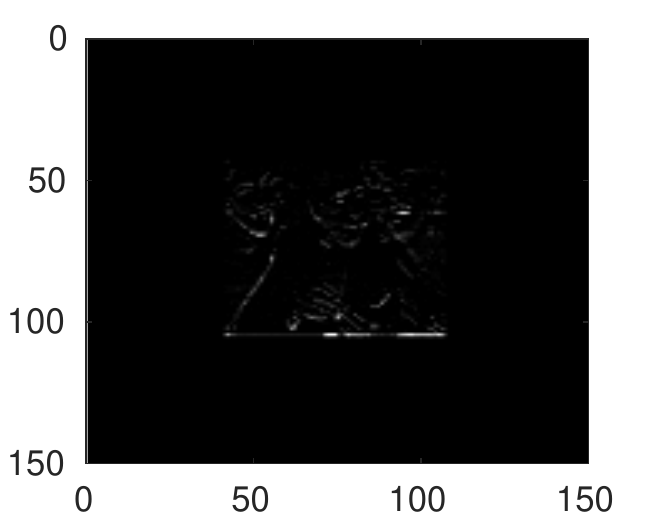}}\\
\vspace{-.5cm}
\subfloat[Linear velocity]{\includegraphics[width=1.5in]{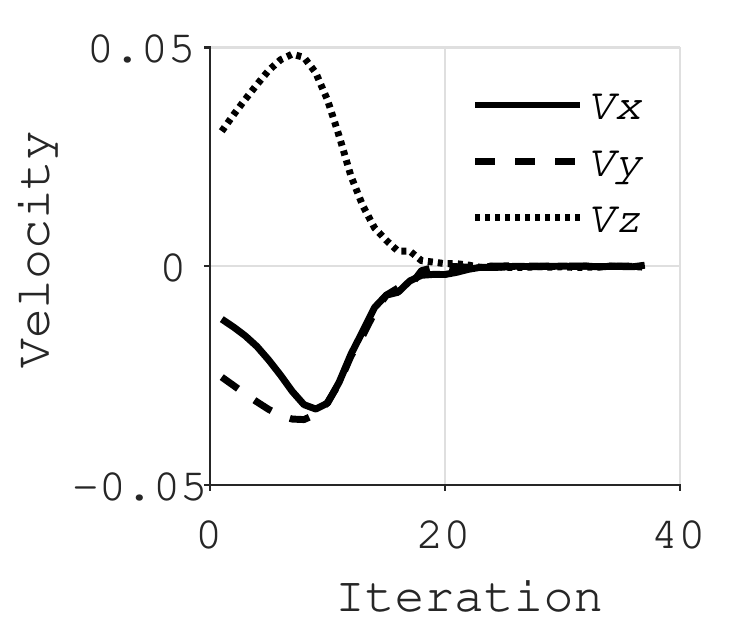}}
\hspace{0.01cm}
\subfloat[Angular velocity]{\includegraphics[width=1.5in]{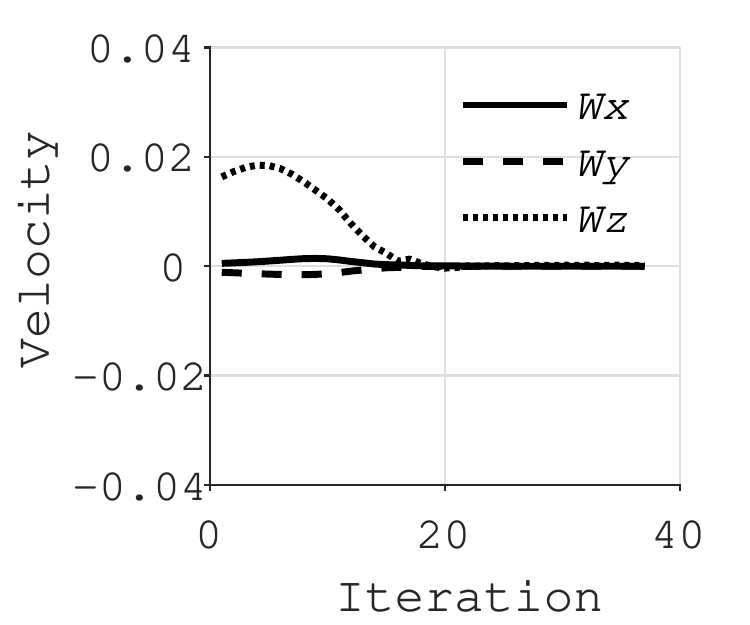}}
\hspace{0.01cm}
\subfloat[error]{\includegraphics[width=1.5in]{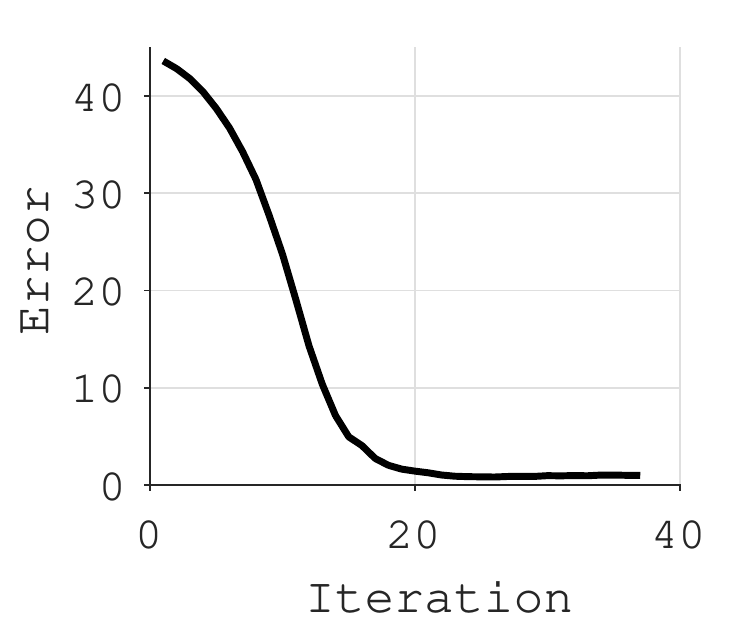}}
\vspace{-.45cm}
\caption{Test case for 3D SMM visual servoing.}
\label{fig:smm_exp_3D}
\end{figure}

Next, a spatial case with a $6$-DOF camera and a planar object placed at a fixed distance as depicted in Fig. \ref{fig:sim_env}(b) is considered. For experimental purposes, initial depth is considered as $-0.5m$ and initial camera plane is assumed parallel to the desired one. Figure \ref{fig:smm_exp_3D} shows the results for SMM based visual servoing for one of the $3D$ test case from Table \ref{tb:exp_3D}  (i.e., Experiment $3$). It can be observed that the camera converges to the desired pose with final errors close to zero. From the final error in camera pose, in Table \ref{tb:exp_3D}, and the pixel level error between final and desired images ($\boldsymbol{I}-\boldsymbol{I^*}$),  in Fig. \ref{fig:smm_exp_3D}(c), it is evident that desired pose is reached. This validates the convergence of proposed framework in 3D case. Plots of linear and angular velocity, and norm of error in Fig.\ref{fig:smm_exp_3D}(d-f) depict that the proposed algorithm converges successfully to desired values. In order to validate robustness of proposed approach, servoing using different camera configurations were performed as presented inTable \ref{tb:exp_3D}. It can be seen from the error in position that the algorithm is able to perform successful visual servoing for all the cases. This validates the robustness to different camera poses.  

\subsection{SMM visual servoing using a robot manipulator}
In this subsection, effectiveness of the proposed framework is demonstrated for 3D objects. For this purpose, a 3D spatial environment was created in ROS Gazebo as shown in  Fig.\ref{fig:sim_env}(c). The environment consists of an UR5 manipulator (6-DOF), a depth camera (Microsoft Kinnect) mounted at the end-effector and objects such as a cereal bowls and coke cans placed randomly on a table. A detailed description and results of the experiments are discussed in following sections.
\subsubsection{Visual servoing towards 3D objects in gazebo}
\begin{figure}[!htp]
\centering
\captionsetup[subfigure]{labelformat=empty}
\subfloat[]{\includegraphics[width=1.3in]{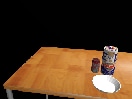}}
\hspace{0.4cm}
\subfloat[]{\includegraphics[width=1.3in]{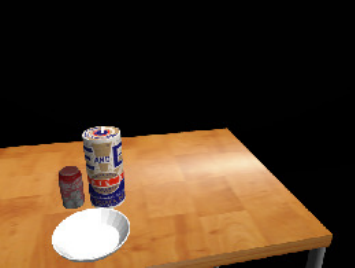}}
\hspace{0.4cm}
\subfloat[]{\includegraphics[width=1.3in]{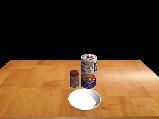}}\\
\vspace{-1.55\baselineskip}

\subfloat[]{\includegraphics[width=1.3in]{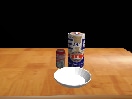}}
\hspace{0.4cm}
\subfloat[]{\includegraphics[width=1.3in]{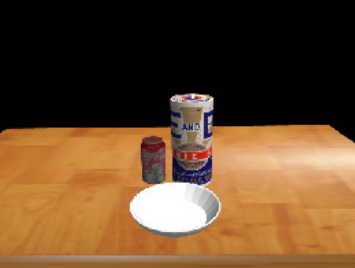}}
\hspace{0.4cm}
\subfloat[]{\includegraphics[width=1.3in]{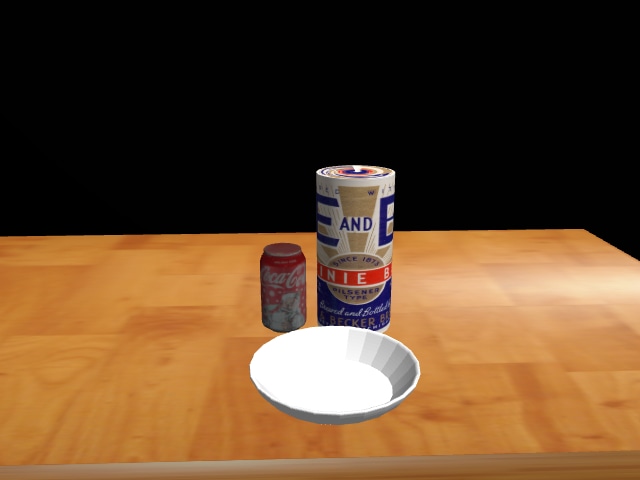}}\\
\vspace{-1.6\baselineskip}

\subfloat[]{\includegraphics[width=1.3in]{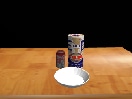}}
\hspace{0.4cm}
\subfloat[]{\includegraphics[width=1.3in]{figures/Desired_3D_posn1.jpg}}
\hspace{0.4cm}
\subfloat[]{\includegraphics[width=1.3in]{figures/Desired_3D_posn1.jpg}}\\
\vspace{-1.6\baselineskip}

\subfloat[]{\includegraphics[width=1.5in]{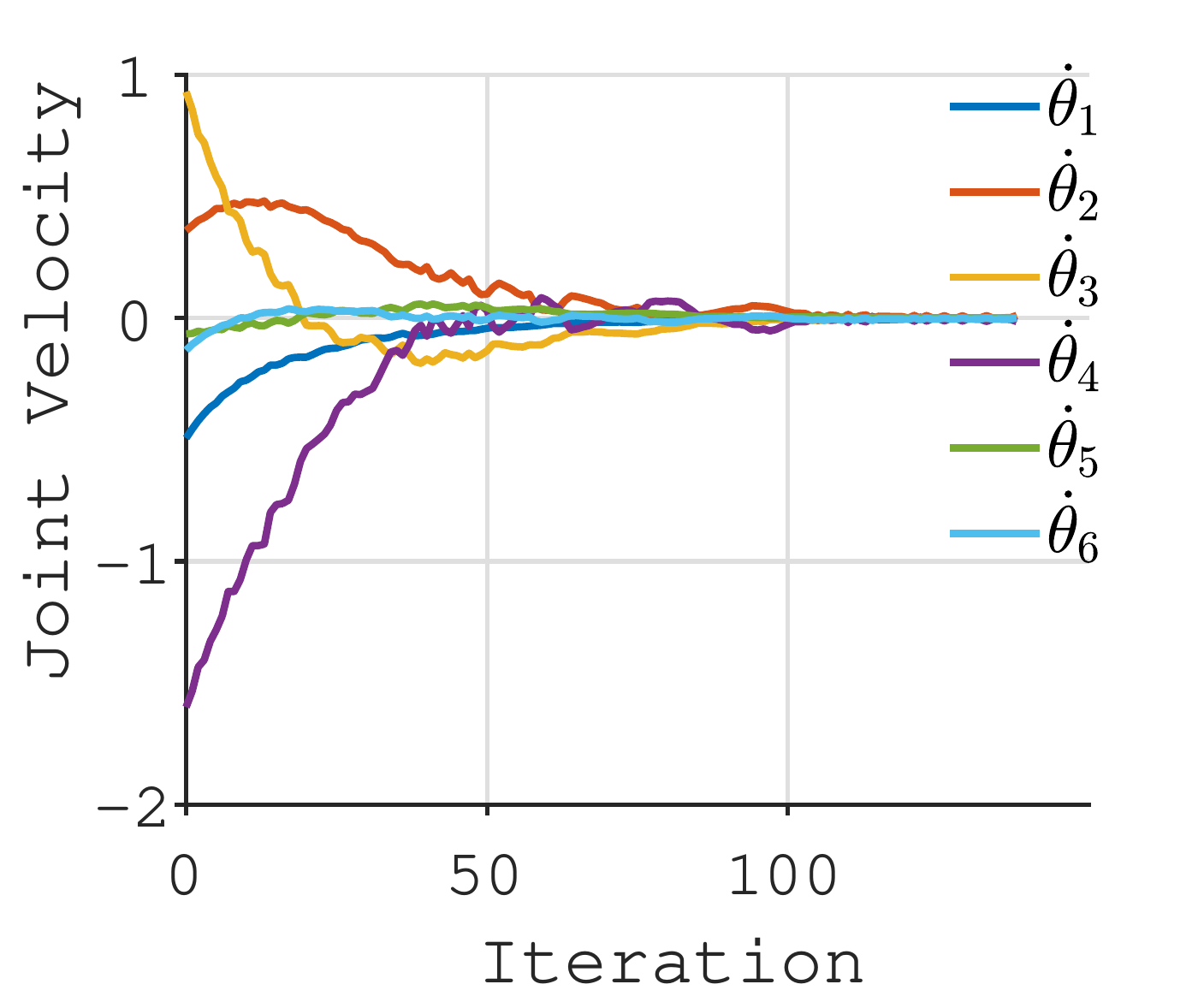}}
\hspace{0.01cm}
\subfloat[]{\includegraphics[width=1.5in]{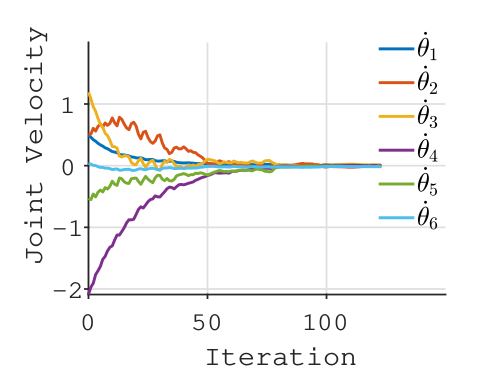}}
\hspace{0.01cm}
\subfloat[]{\includegraphics[width=1.5in]{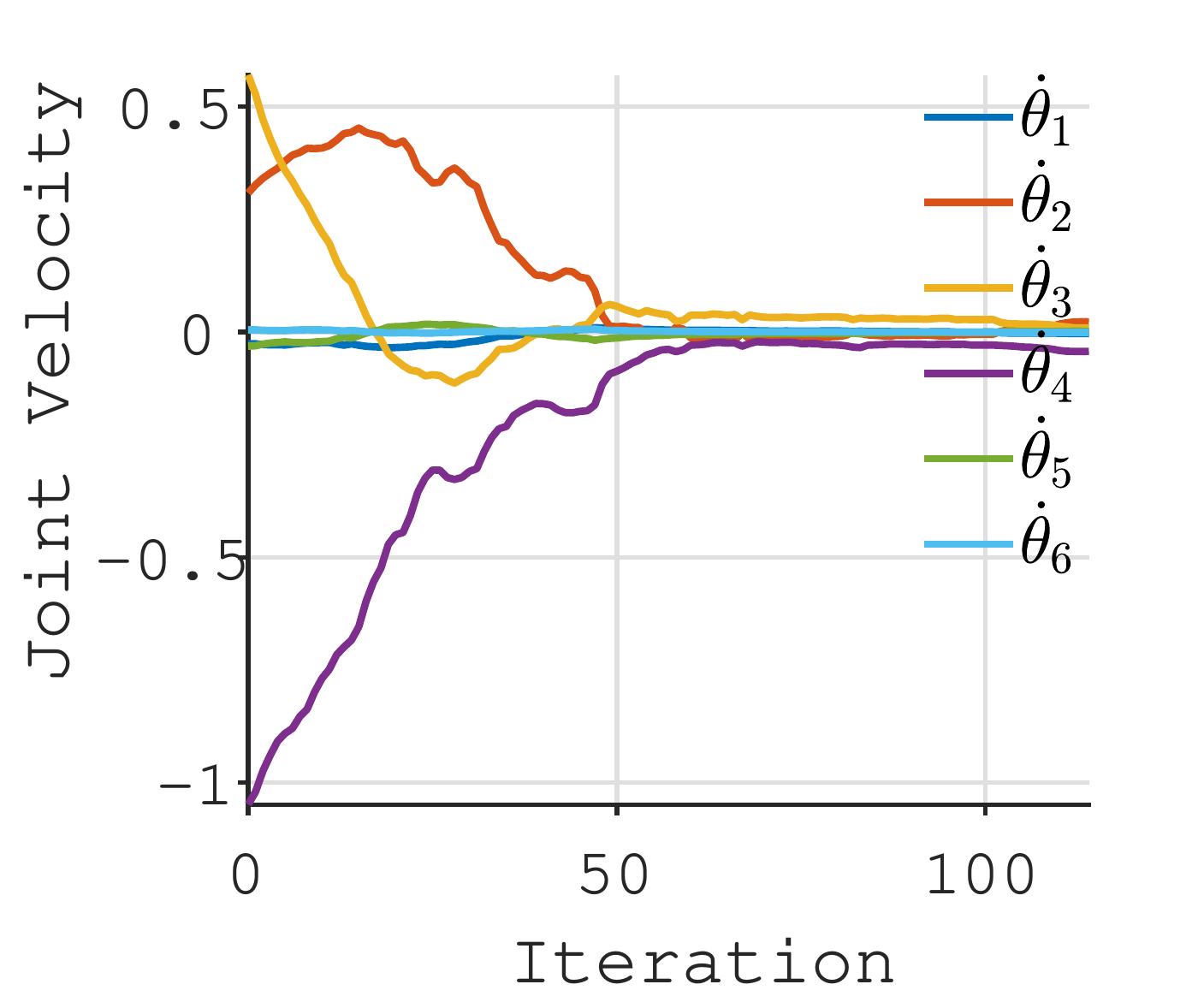}}\\
\vspace{-1.65\baselineskip}

\subfloat[]{\includegraphics[width=1.5in]{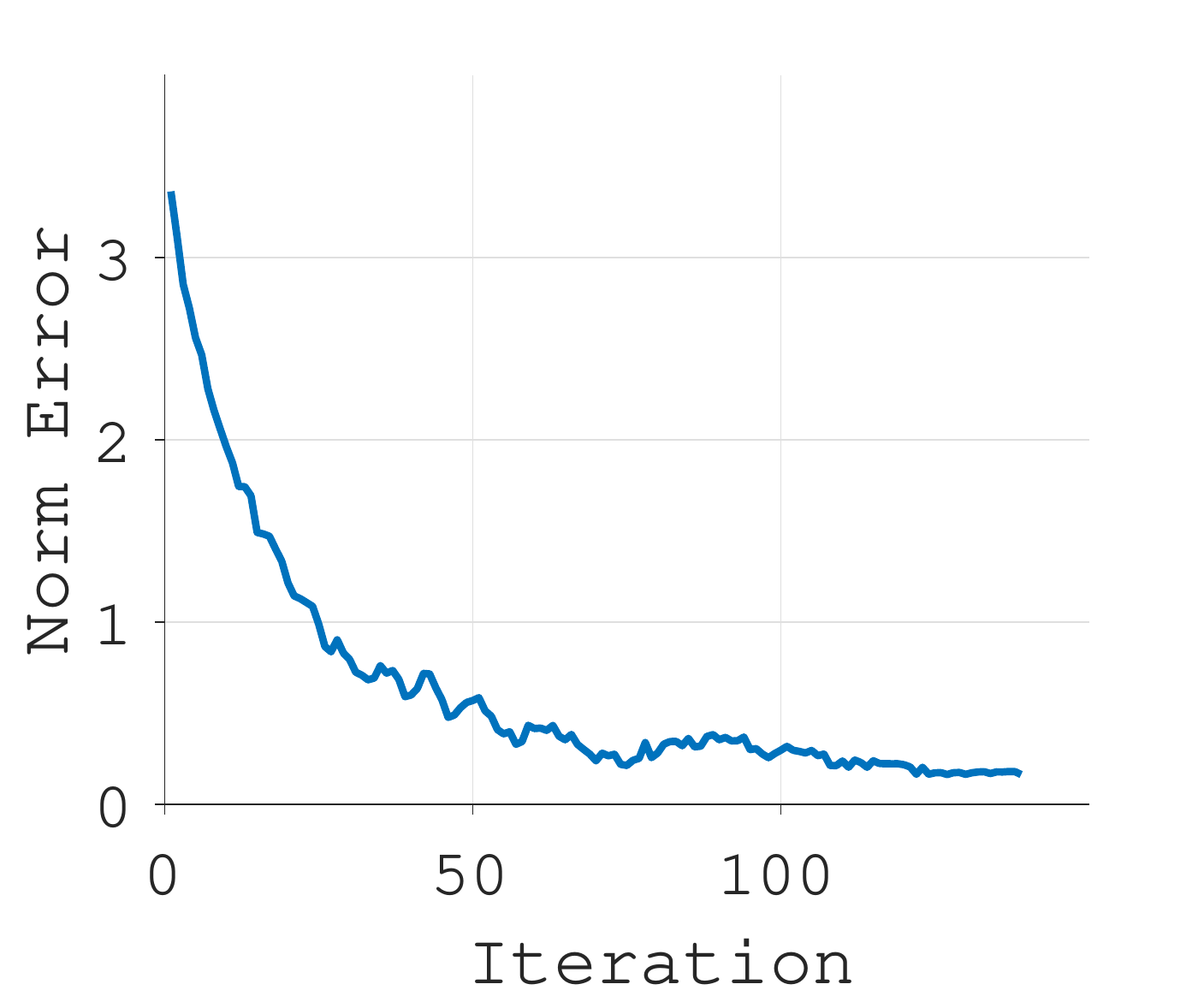}}
\hspace{0.01cm}
\subfloat[]{\includegraphics[width=1.5in]{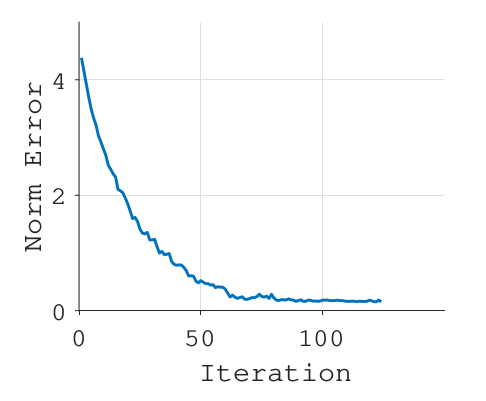}}
\hspace{0.01cm}
\subfloat[]{\includegraphics[width=1.5in]{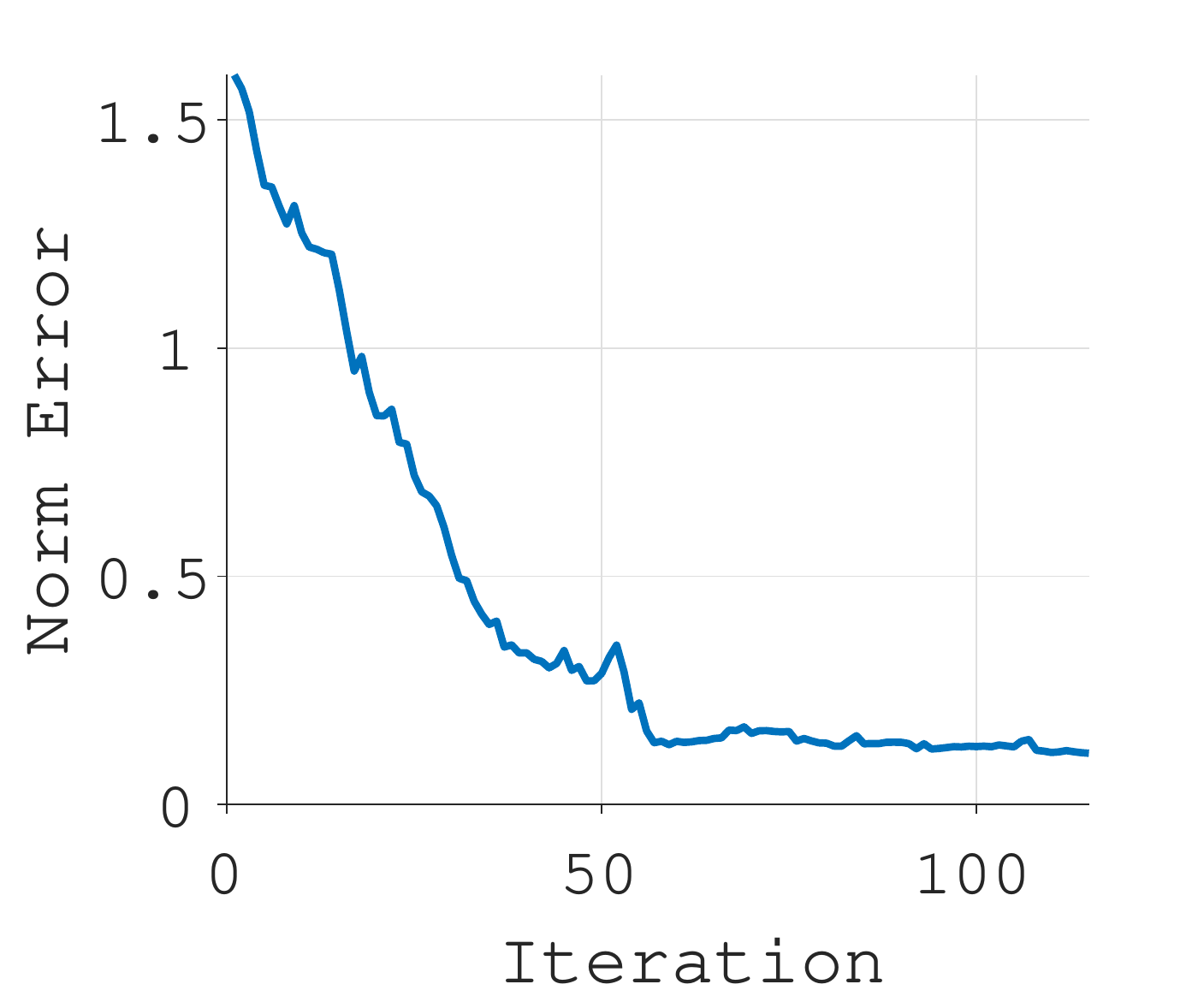}}\\
\vspace{-1.6\baselineskip}

\subfloat[]{\includegraphics[width=1.55in]{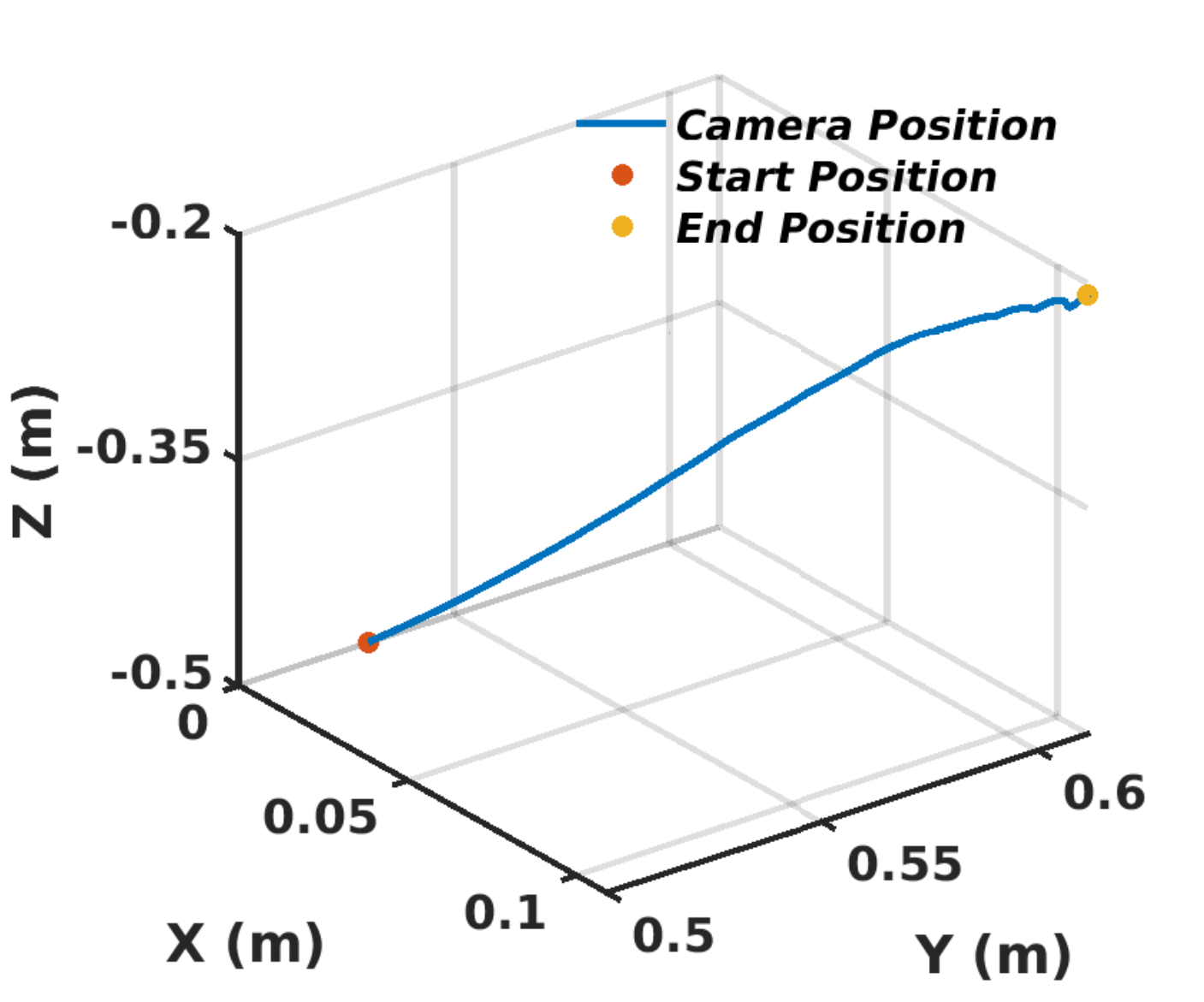}}
\hspace{0.01cm}
\subfloat[]{\includegraphics[width=1.55in]{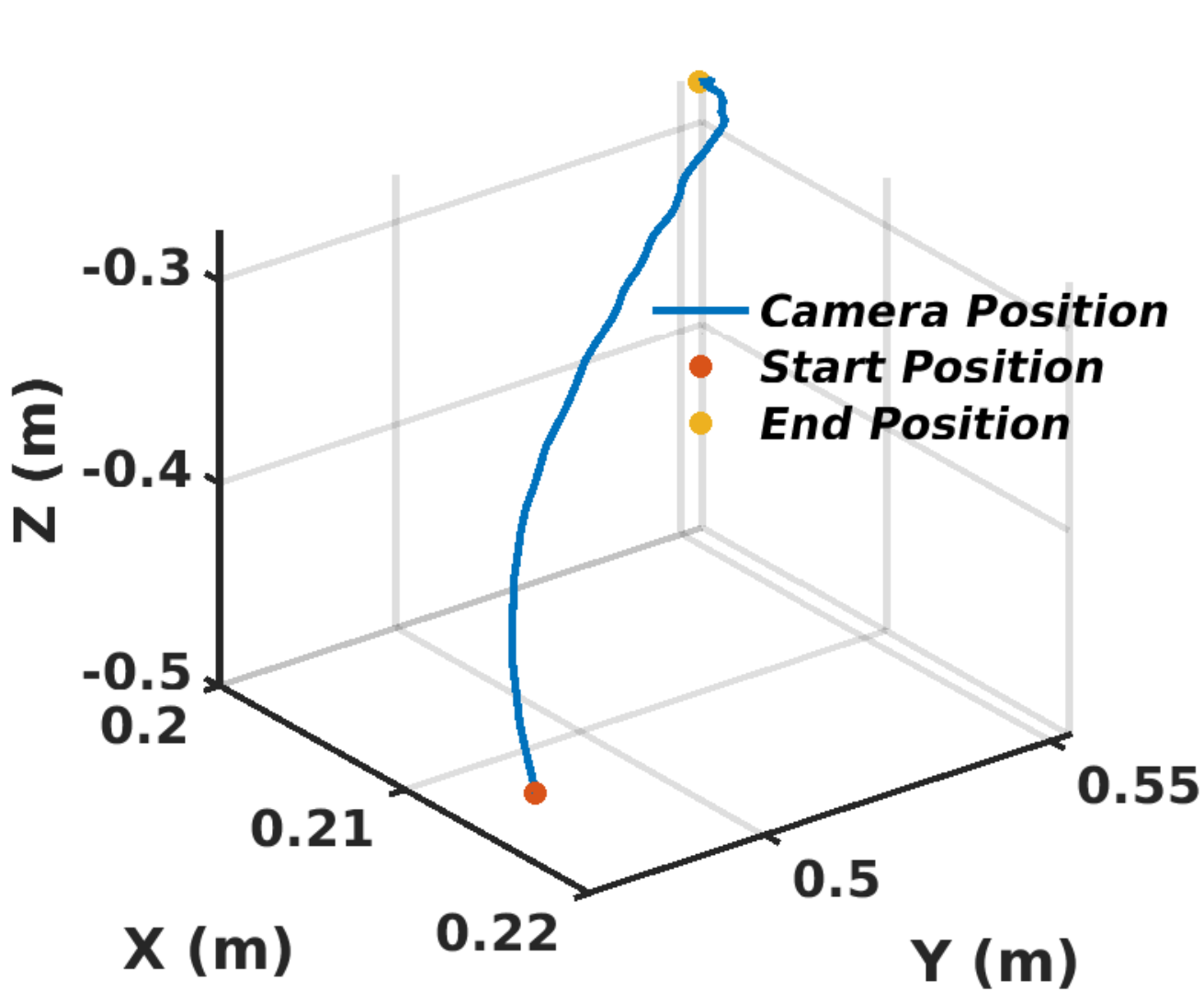}}
\hspace{0.01cm}
\subfloat[]{\includegraphics[width=1.55in]{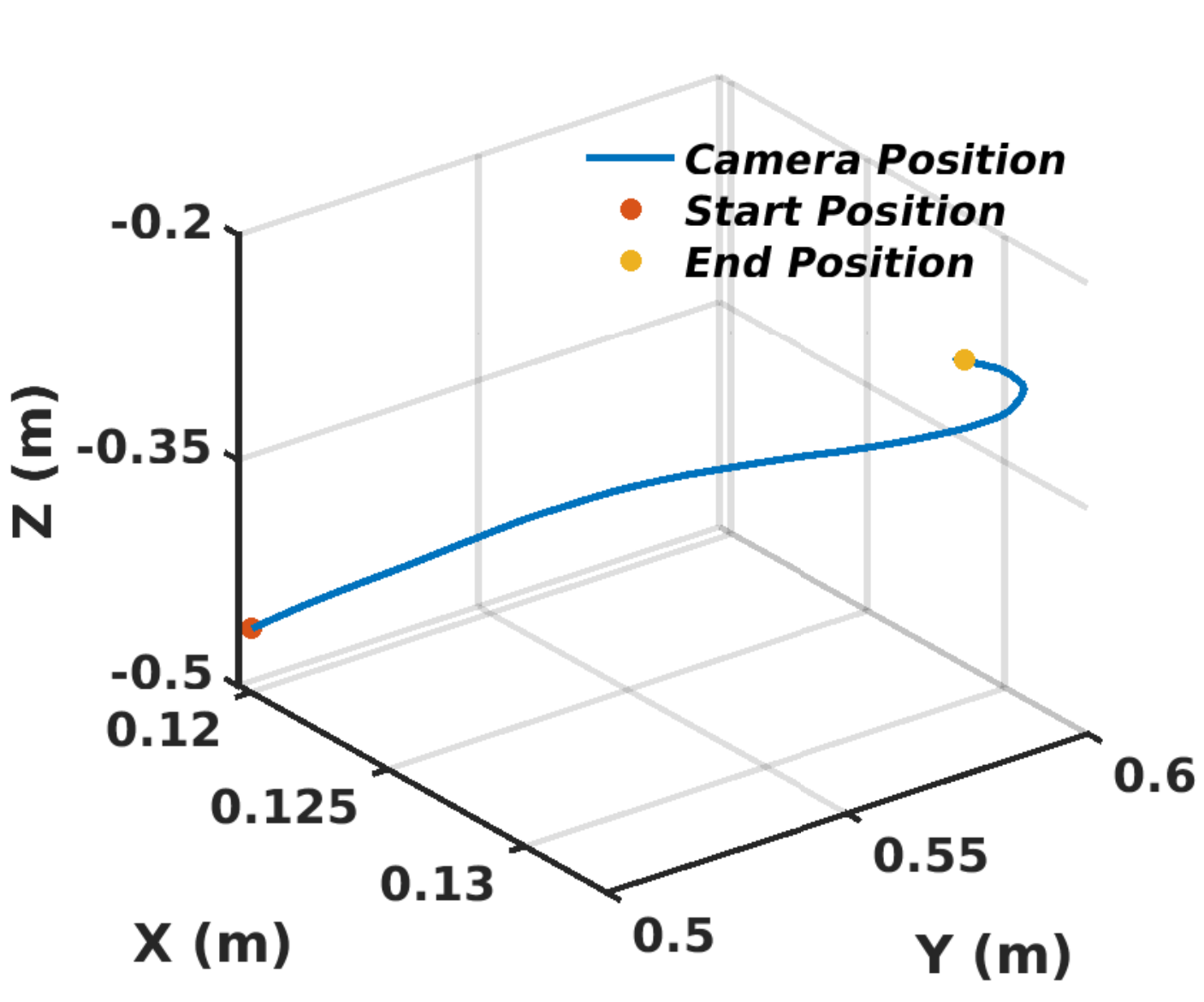}}\\
\caption{Results of SMM visual servoing for three different start positions in gazebo using UR5 manipulator. Rows 1-6 shows initial image, final obtained image, desired image, joint velocities (rad/s), norm error (pixels) and camera trajectory plots respectively.}
    \label{fig:3D_posn1_0ld}
\end{figure}

The UR5 manipulator mounted on a stand is positioned in front of a wooden table as seen in Fig. \ref{fig:sim_env}(c). In this setup, experiments were performed to test the algorithm for reaching a fixed desired position, row 3 in Fig. \ref{fig:3D_posn1_0ld}, from different initial positions, row 1 in Fig. \ref{fig:3D_posn1_0ld}. The actual images obtained are shown in the row 2 in Fig. \ref{fig:3D_posn1_0ld} which match with the desired images in row 3. From the plots of joint velocity in row 4 and norm error in row 5, it can be concluded that the algorithm converges satisfactorily with exponential decay in norm of error and joint velocities tending to zero. The camera positions in row 6 shows the path traced by the end-effector reaching the desired position. A number of experiments were preformed as depicted in Table \ref{tb:3D_Gazebo} which shows that final error in pose is reduced close to zero for all cases similar  to Fig. \ref{fig:3D_posn1_0ld}. This validates the convergence and robustness of the proposed framework in a realistic environment using a robotic manipulator.
\begin{table}
\caption{Positioning error : $3D$ positioning task in gazebo}
\centering
\begin{adjustbox}{max width=\textwidth}
\begin{tabular}{|l|cccccc|cccccc|}
\hline
 & \multicolumn{6}{c|}{Initial error} & \multicolumn{6}{c|}{Final error} \\ 
\hline 
Exp. & $p_x(m)$ & $p_y(m)$ &$p_z(m)$ & $\theta_x$ &$\theta_y$ & $\theta_z$ & $p_x(m)$ & $p_y(m)$ &$p_z(m)$ & $\theta_x$ &$\theta_y$ & $\theta_z$ \\ \hline \hline
$1$ & $0.1$ & $0.07$ & $0.23$ &$-5.73^{\circ}$ &$0.46^{\circ}$ &$-11.46^{\circ}$ & $0.0122$ & $-0.0257$ & $-0.0242$ & $1.19^{\circ}$ &$-0.51^{\circ}$&$0.54^{\circ}$\\
$2$ & $-0.1$ & $0.12$ & $0.23$ &$-5.73^{\circ}$ &$0.46^{\circ}$ &$11.46^{\circ}$ & $-0.078$ & $0.033$ & $0.046$ & $-1.66^{\circ}$ &$1.03^{\circ}$&$-2.29^{\circ}$\\
$3$ & $0.0018$ & $0.082$ & $0.23$ &$-5.73^{\circ}$ &$0.46^{\circ}$ &$0^{\circ}$ & $-0.0105$ & $0.0048$ & $0.011$ & $-0.04^{\circ}$ &$0.28^{\circ}$&$0.49^{\circ}$\\
$4$ & $-0.052$ & $0.36$ & $0.21$ &$-6.87^{\circ}$ &$0^{\circ}$ &$8.59^{\circ}$ & $-0.0037$ & $-0.0094$ & $0.015$ & $-1.54^{\circ}$ &$0.47^{\circ}$&$3.78^{\circ}$\\
$5$ & $0.17$ & $-0.4$ & $0.24$ &$-22.9^{\circ}$ &$0.02^{\circ}$ &$-20.05^{\circ}$ & $-0.0014$ & $-0.0011$ & $0.0016$ & $-0.12^{\circ}$ &$-0.034^{\circ}$&$-0.057^{\circ}$\\
$6$ & $0.16$ & $0.28$ & $0.15$ &$-5.73^{\circ}$ &$0^{\circ}$ &$-14.32^{\circ}$ & $0.0012$ & $-0.0012$ & $-0.0011$ & $-0.085^{\circ}$ &$0.47^{\circ}$&$0.25^{\circ}$\\
\hline 
\end{tabular}
\end{adjustbox}

\label{tb:3D_Gazebo}
\end{table}
\subsubsection{Close object pursuit using SMM}
\begin{figure}[t]
\centering
\captionsetup[subfigure]{labelformat=empty}
\subfloat[]{\includegraphics[width=2.7in]{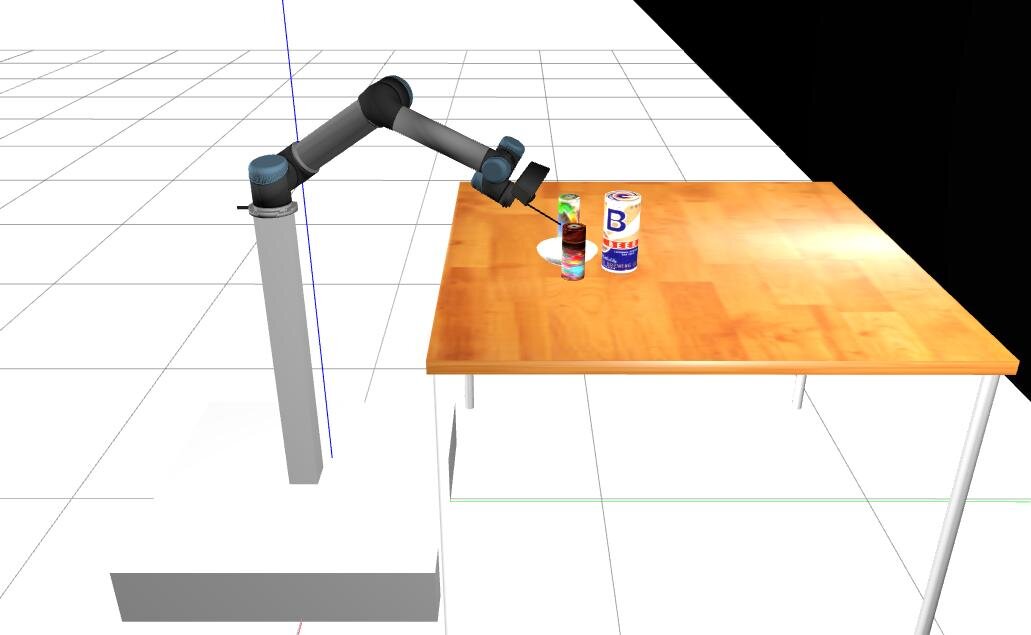}}
\hspace{0.05cm}
\subfloat[]{\includegraphics[width=2in]{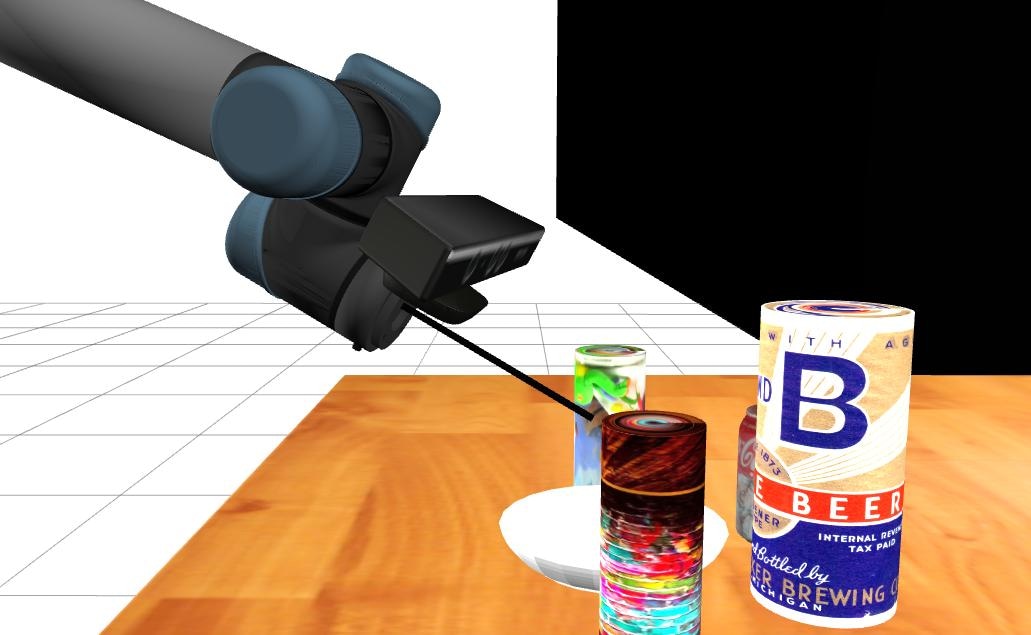}}
\caption{UR5 manipulator positioning at the end of SMM Visual Servoing}
	\label{fig:Gazebo_Env}	
\end{figure}
\begin{figure}
\centering
\captionsetup[subfigure]{labelformat=empty}
\subfloat[]{\includegraphics[width=1.3in]{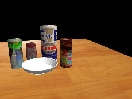}}
\hspace{0.4cm}
\subfloat[]{\includegraphics[width=1.3in]{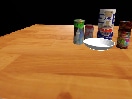}}
\hspace{0.4cm}
\subfloat[]{\includegraphics[width=1.3in]{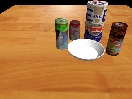}}\\
\vspace{-1.5\baselineskip}

\subfloat[]{\includegraphics[width=1.3in]{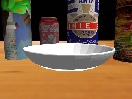}}
\hspace{0.4cm}
\subfloat[]{\includegraphics[width=1.3in]{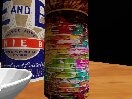}}
\hspace{0.4cm}
\subfloat[]{\includegraphics[width=1.3in]{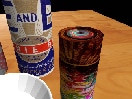}}\\
\vspace{-1.5\baselineskip}

\subfloat[]{\includegraphics[width=1.3in]{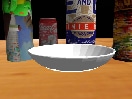}}
\hspace{0.4cm}
\subfloat[]{\includegraphics[width=1.3in]{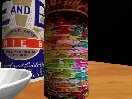}}
\hspace{0.4cm}
\subfloat[]{\includegraphics[width=1.3in]{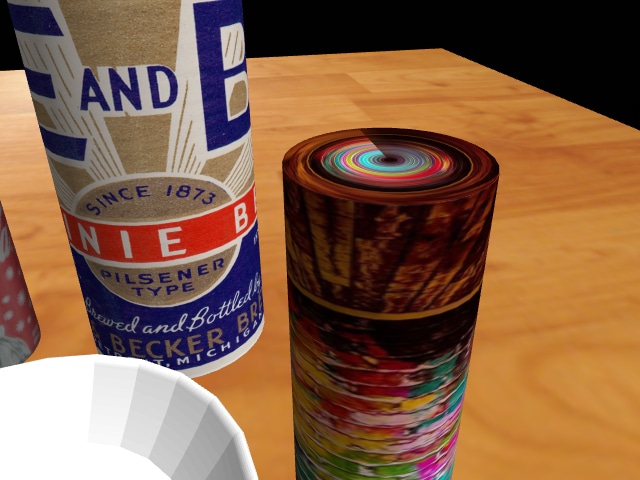}}\\
\vspace{-1.5\baselineskip}

\subfloat[]{\includegraphics[width=1.5in]{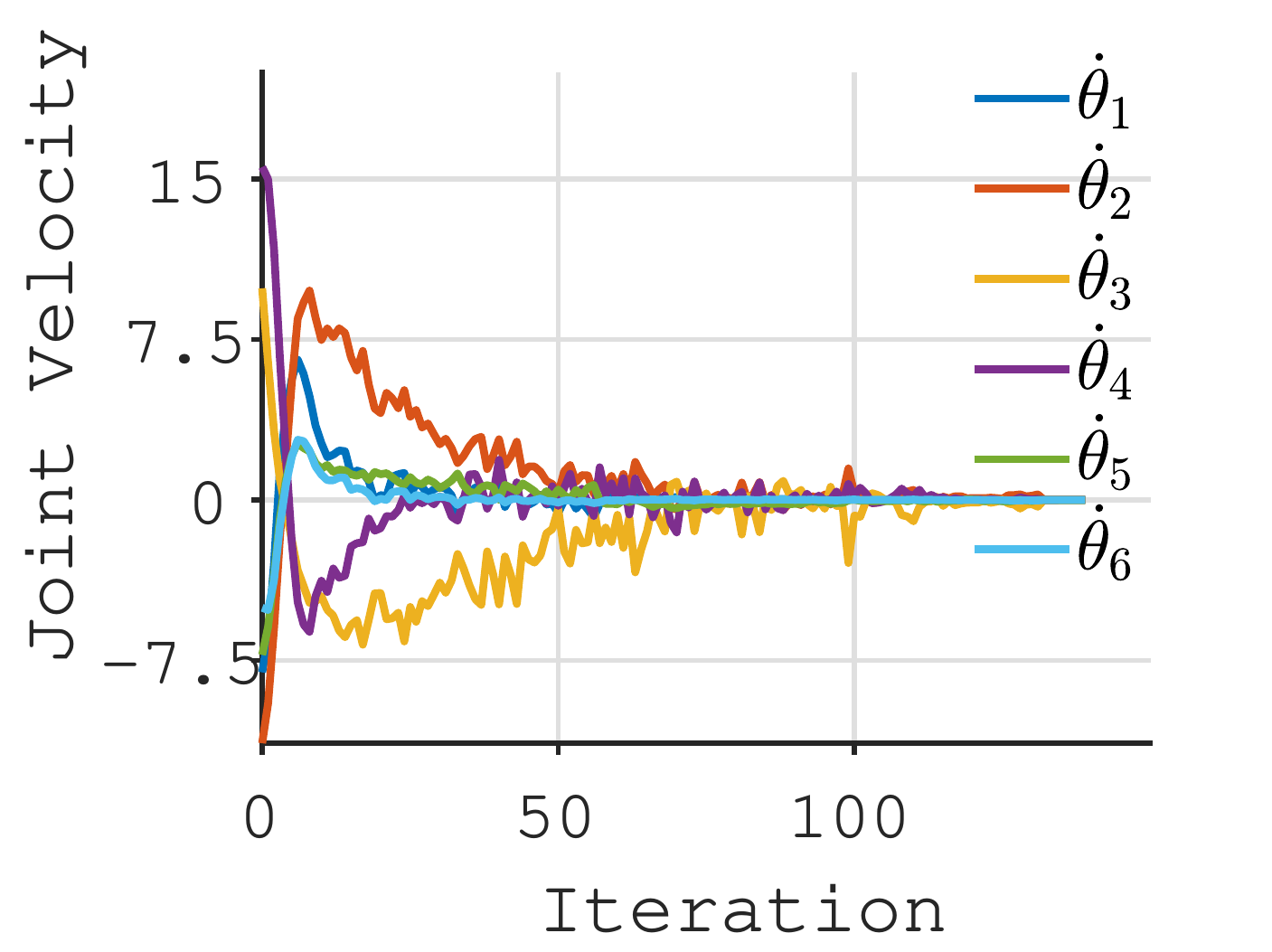}}
\hspace{0.01cm}
\subfloat[]{\includegraphics[width=1.5in]{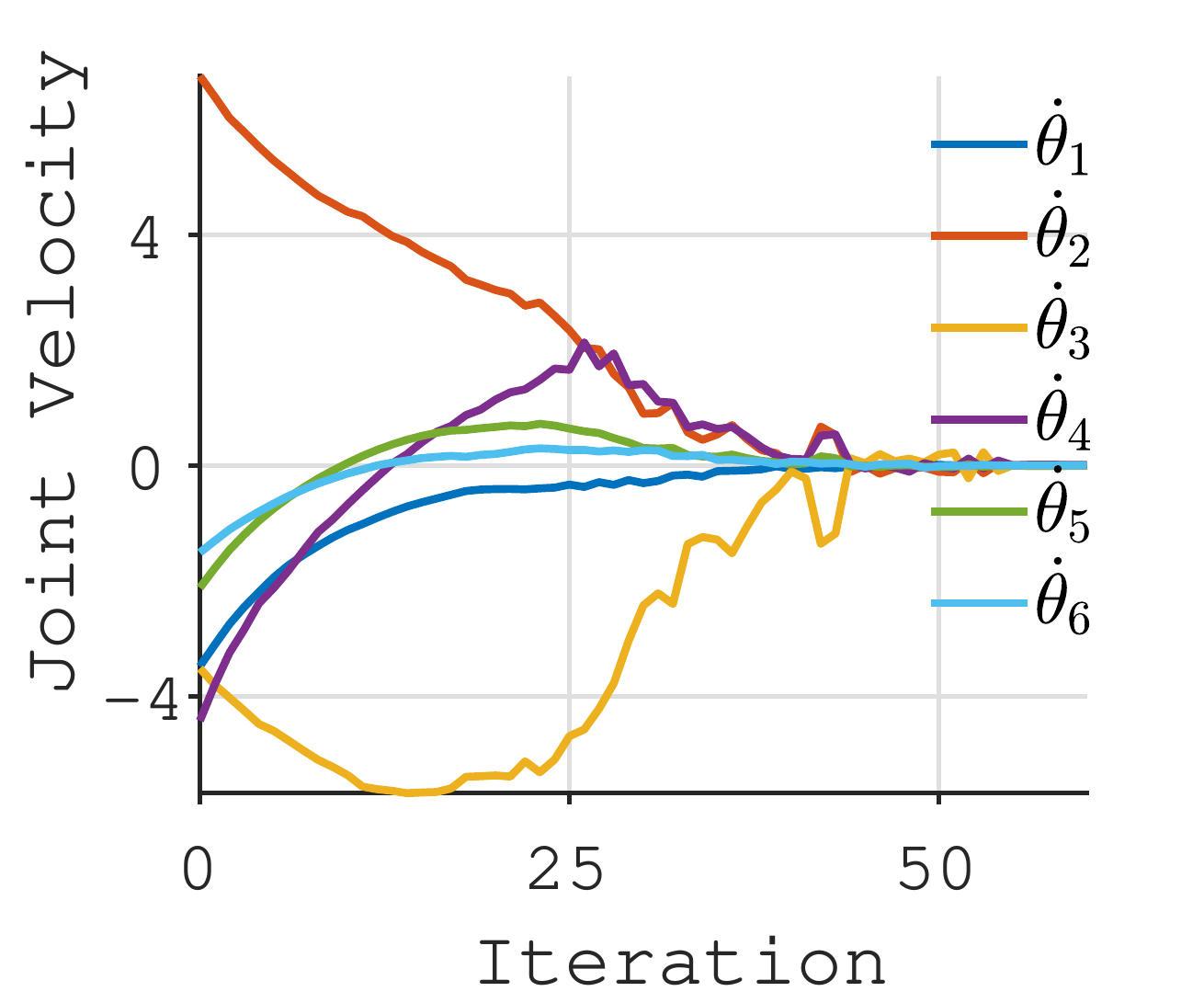}}
\hspace{0.01cm}
\subfloat[]{\includegraphics[width=1.5in]{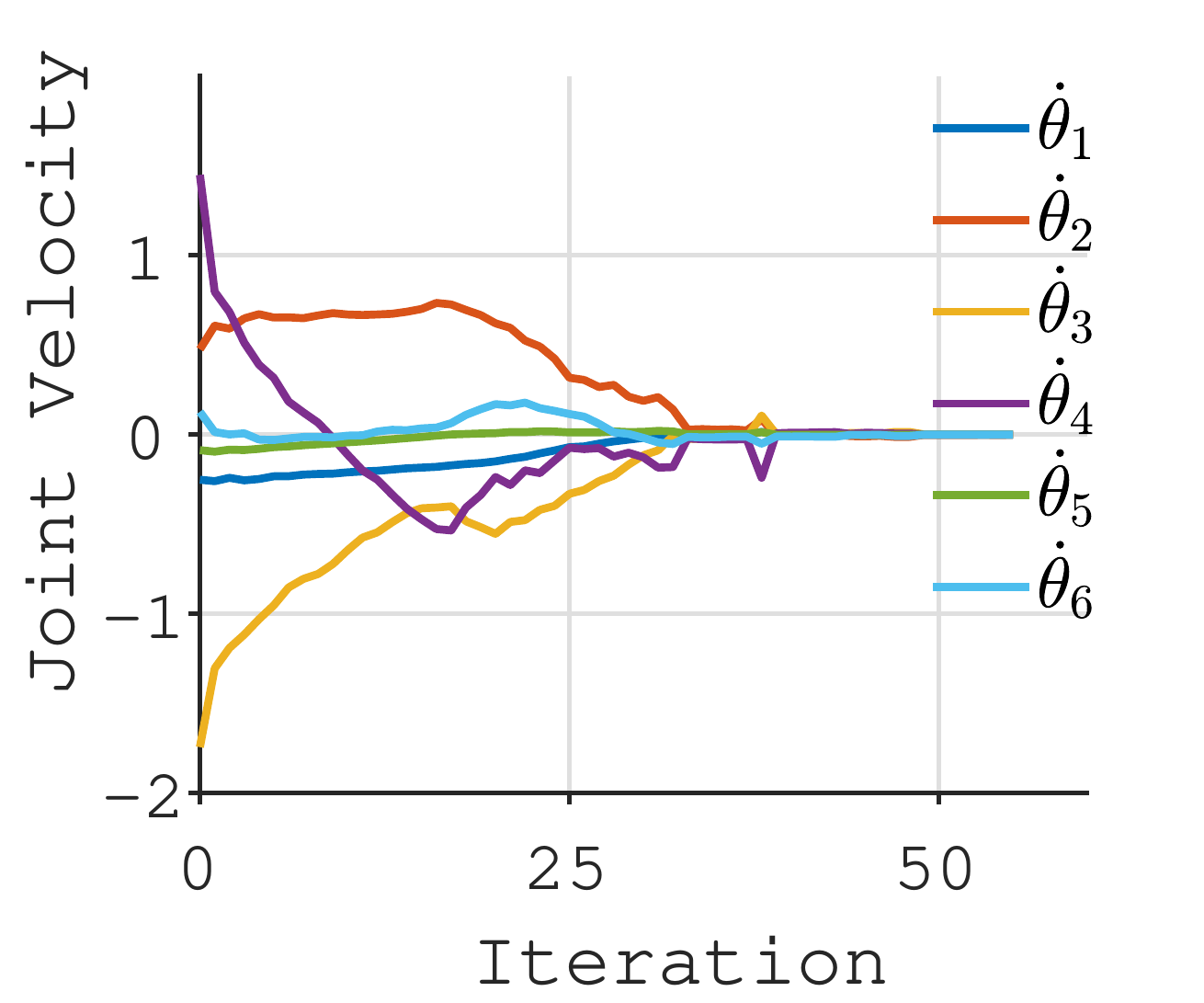}}\\
\vspace{-1.5\baselineskip}

\subfloat[]{\includegraphics[width=1.5in]{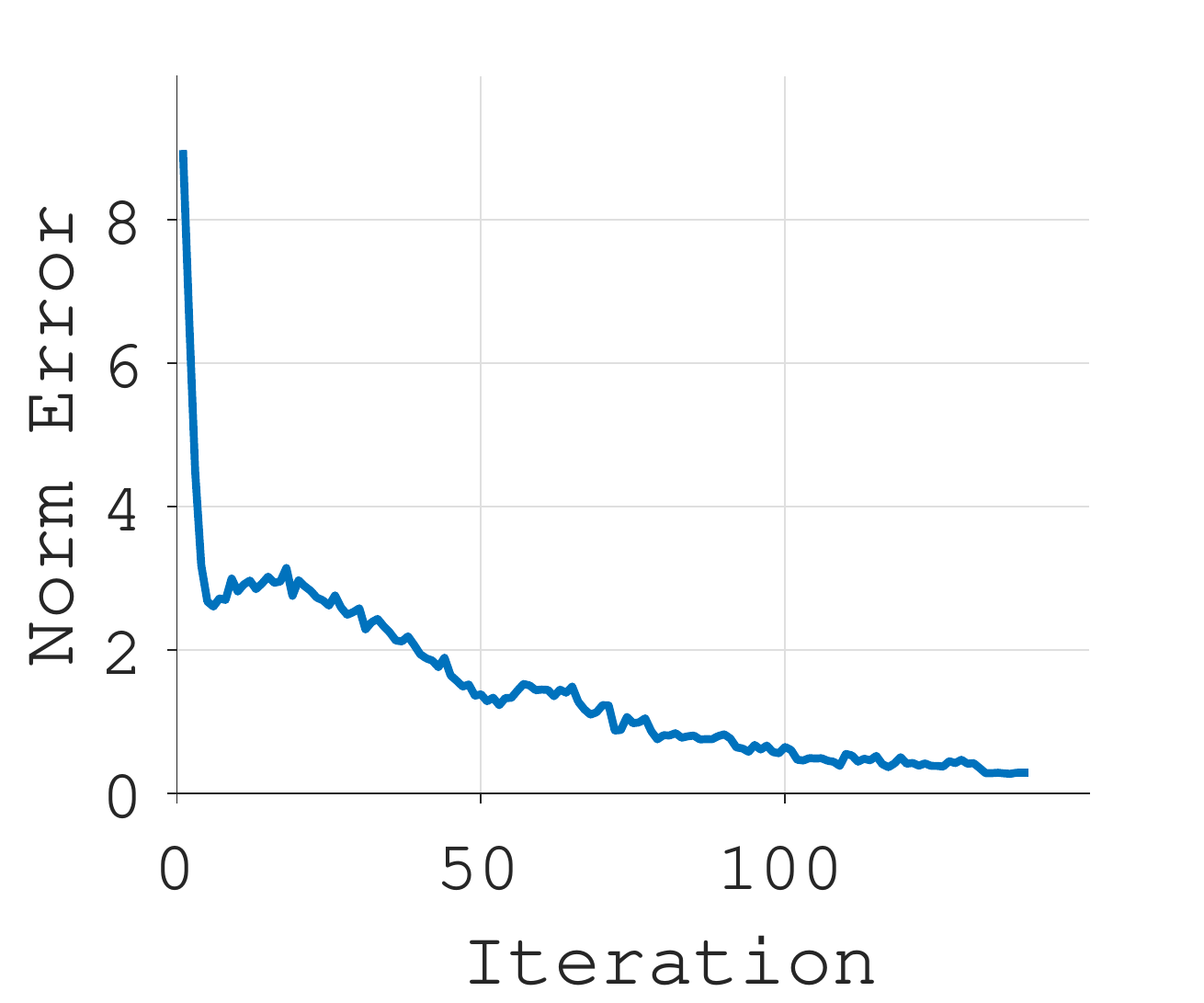}}
\hspace{0.01cm}
\subfloat[]{\includegraphics[width=1.5in]{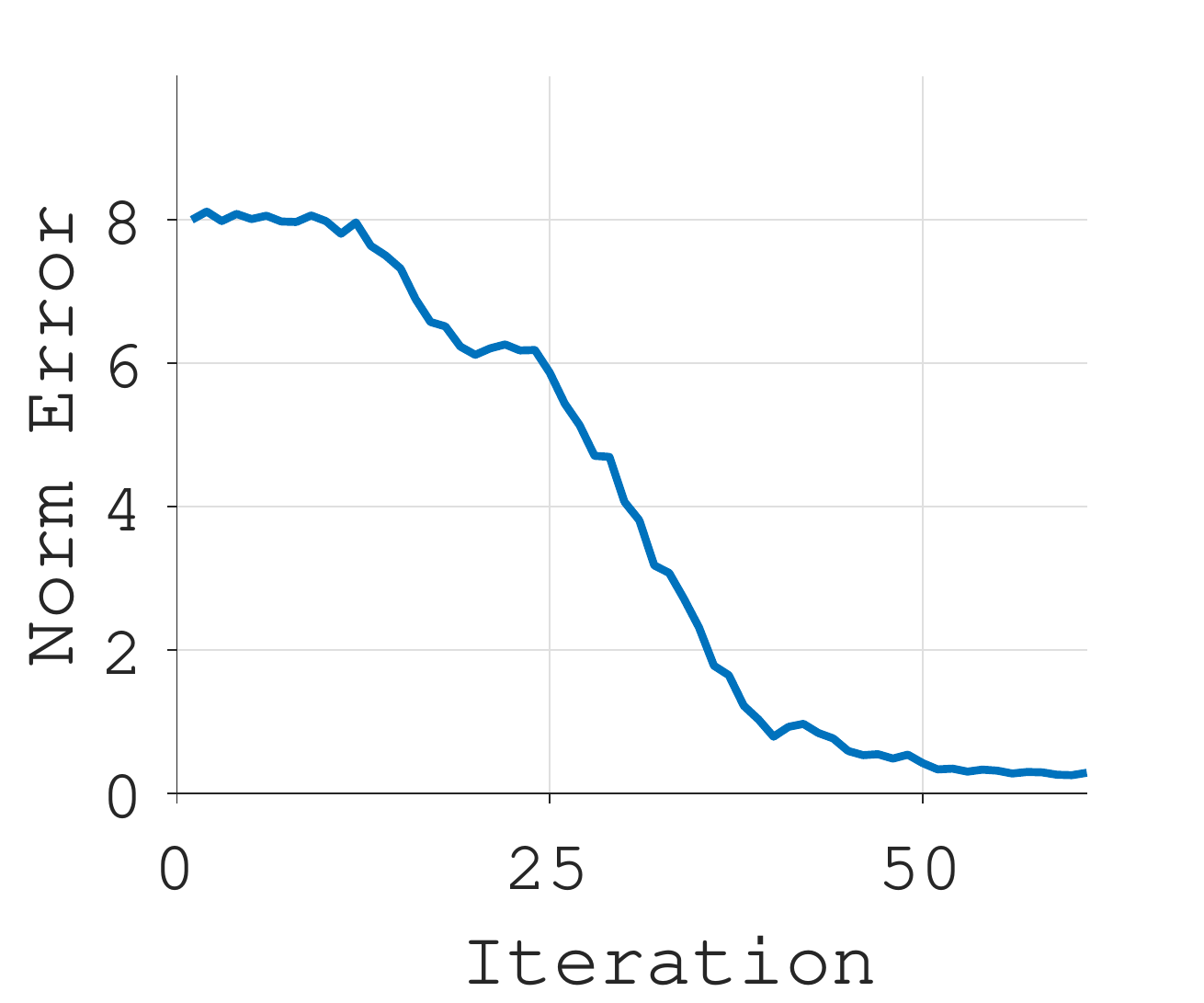}}
\hspace{0.01cm}
\subfloat[]{\includegraphics[width=1.5in]{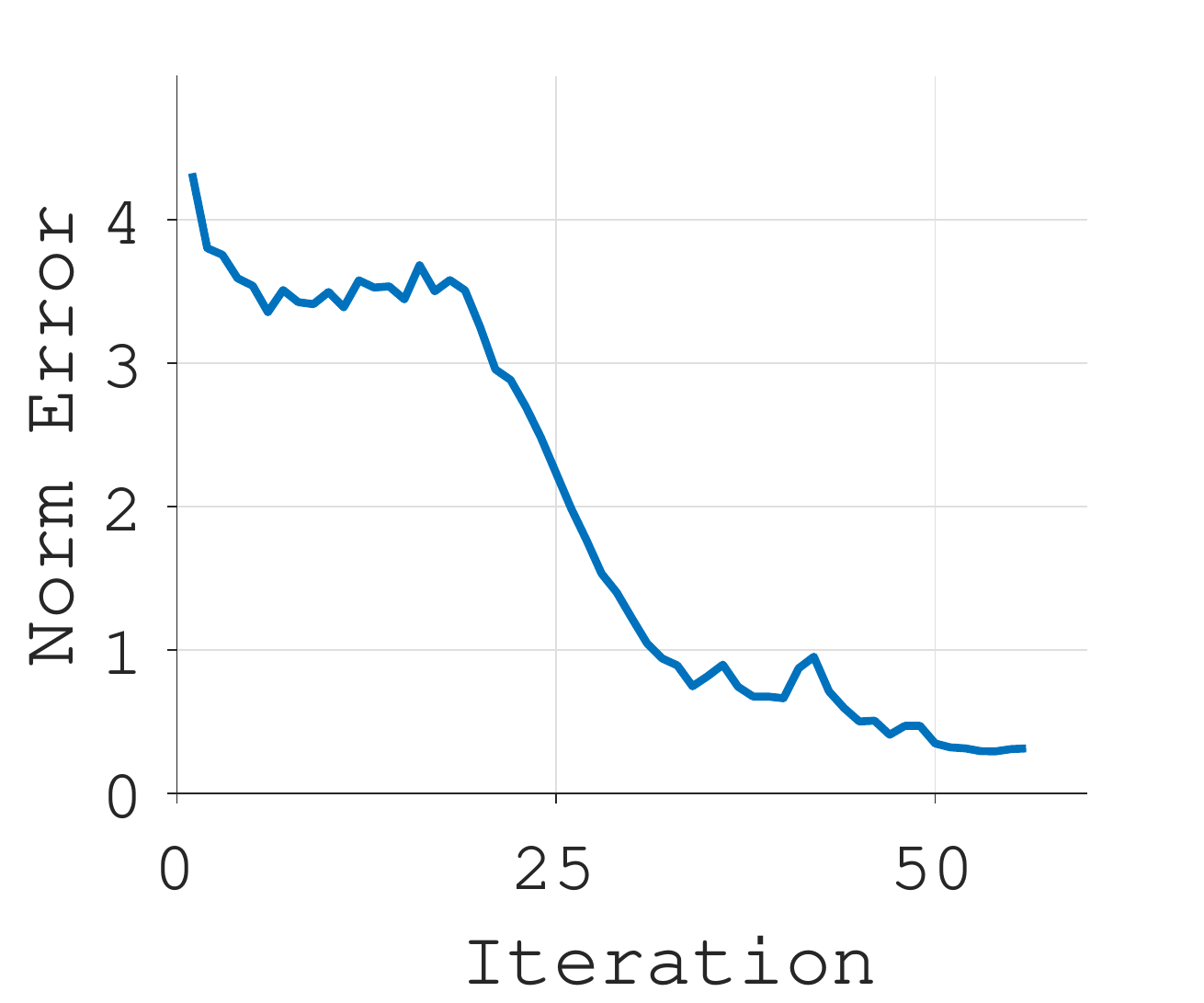}}\\
\vspace{-1.5\baselineskip}

\subfloat[]{\includegraphics[width=1.55in]{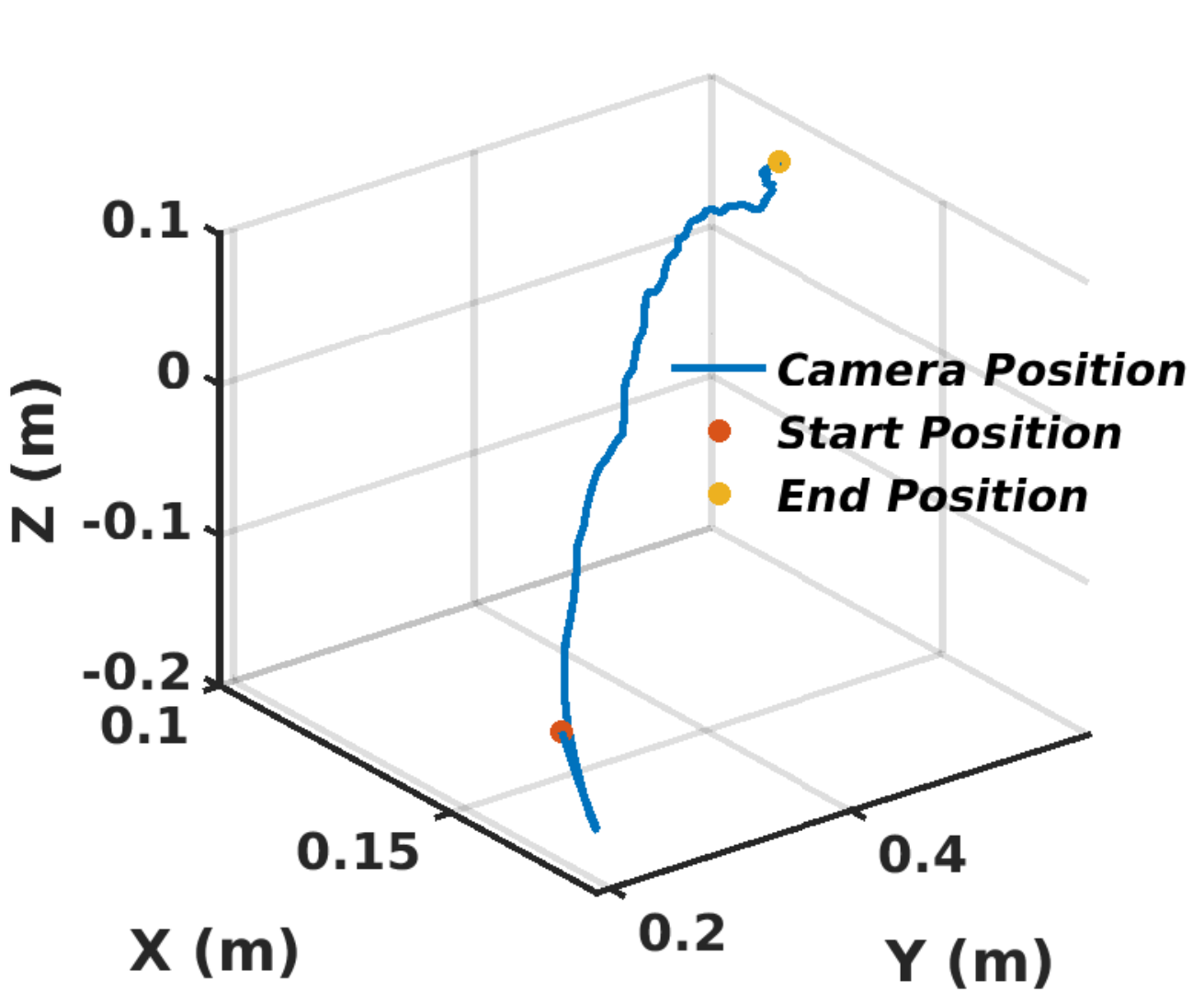}}
\hspace{0.01cm}
\subfloat[]{\includegraphics[width=1.55in]{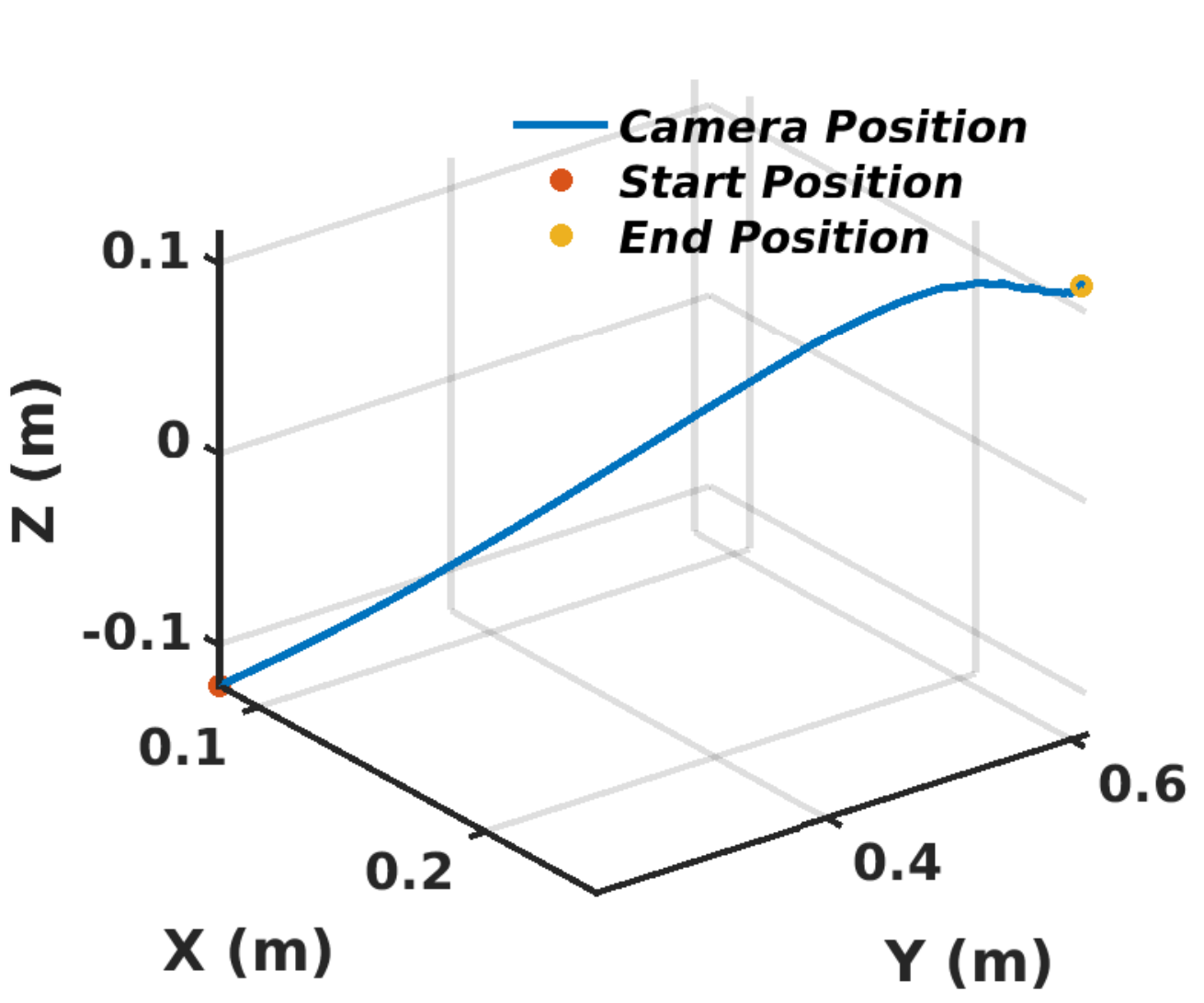}}
\hspace{0.01cm}
\subfloat[]{\includegraphics[width=1.55in]{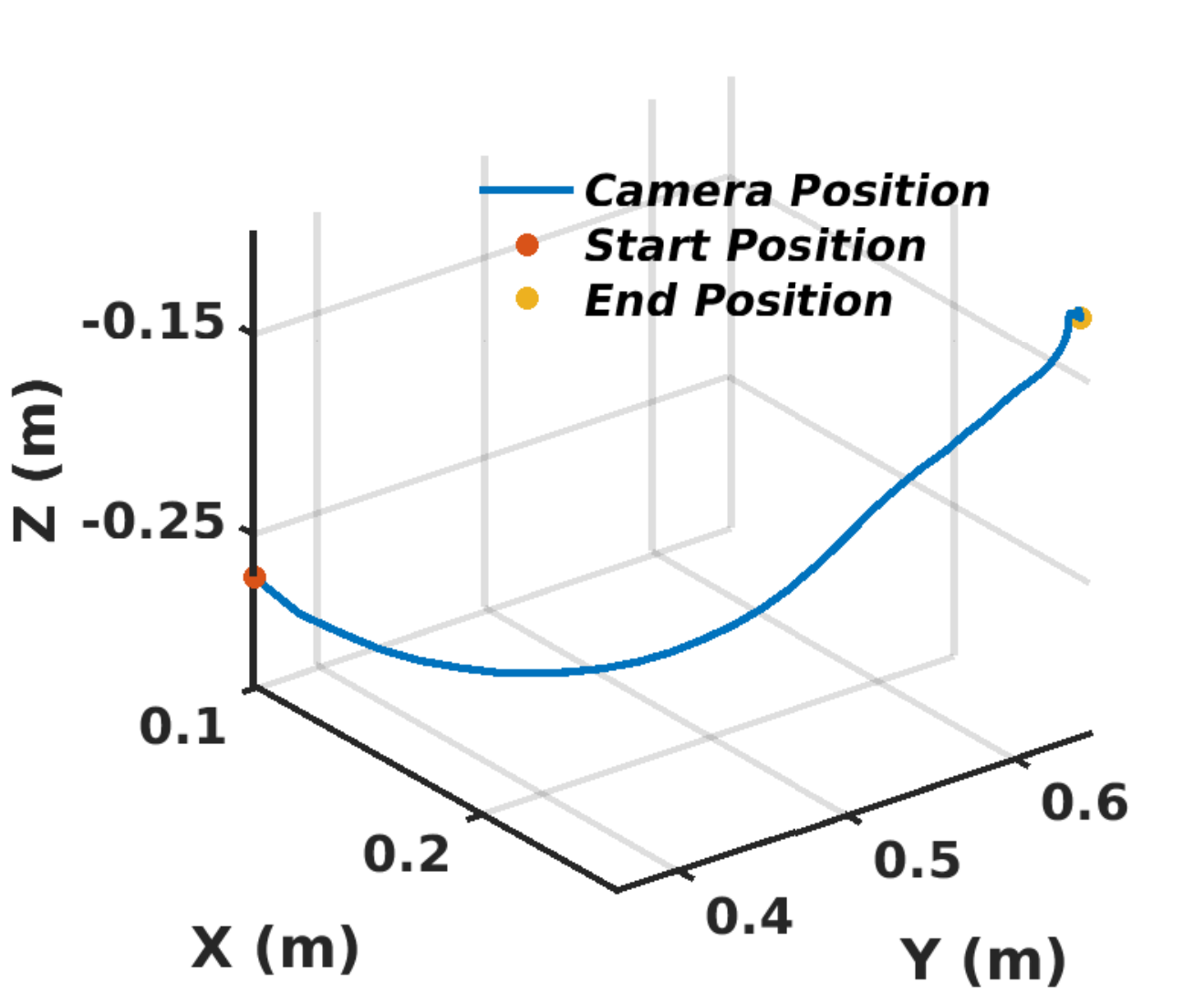}}\\
\vspace{-1\baselineskip}
\caption{Results of SMM visual servoing for three different start and desired positions in gazebo using UR5 manipulator. Rows 1-6 shows initial image, desired image, final obtained image, joint velocities (rad/s), norm error (pixels) and camera trajectory plots respectively. }
    \label{fig:3D_closeup_1}
\end{figure}

In the previous section visual servoing was performed towards a group of 3D objects requiring small camera motion. In this subsection, visual servoing is performed towards a particular object among a cluster of objects, i.e., close object pursuit, requiring large camera motion. The pose of the manipulator reached at the end of visual servoing is shown in Fig. \ref{fig:Gazebo_Env}. Here, an additional link depicting gripper is attached to the end-effector of the UR5 manipulator that touches the object once the manipulator reaches its final position at the end of servoing. A number of experiments were performed to show efficacy of the proposed framework in close object manipulation requiring large camera motion. It can be seen from row 1 in Fig. \ref{fig:3D_closeup_1} that the manipulator starts from a position where the entire cluster of objects is visible. 
The final image corresponds to closer view of the object under consideration as seen in row 3 in Fig. \ref{fig:3D_closeup_1}, which clearly depicts that the manipulator has moved close enough to see the desired image in row 2. This shows the effectiveness of the proposed method for object manipulations requiring large camera motion. Plots in row 4 and row 5 in Fig.\ref{fig:3D_closeup_1} show that the joint velocities and norm error values are reduced exponentially from the start to end as the algorithm converges. The path traced by the end-effector during the pursuit can also be seen in row 6.
\subsection{Robustness Evaluation}
This section discusses the robustness of the proposed visual servoing framework through a larger set of simulated experiments. This includes experiments with different image scenes, image resolutions and occlusions. Results are verified for both 2D image in Fig. \ref{fig:sim_env}(b) and 3D object in Fig.  \ref{fig:sim_env}(c) in gazebo  environment. 
\subsubsection{Effect to image resolution } 
\begin{figure}[h]
\centering
\captionsetup[subfigure]{labelformat=empty}
\subfloat[]{\includegraphics[width=1.2in]{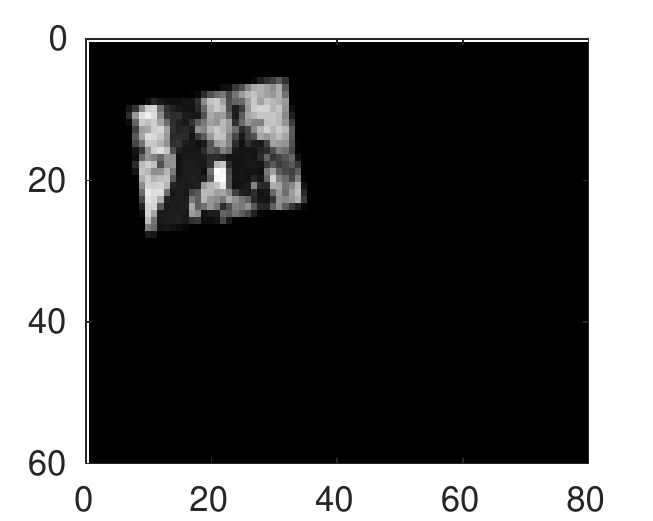}}
\subfloat[]{\includegraphics[width=1.2in]{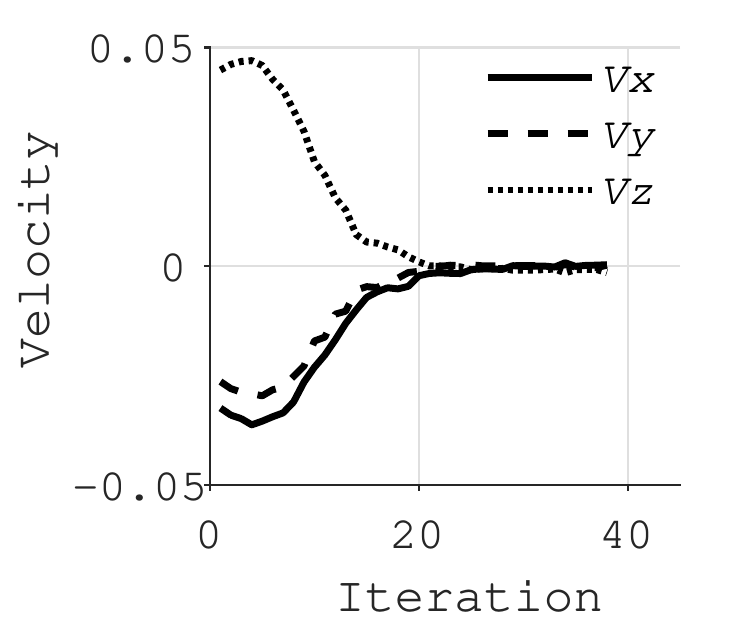}}
\subfloat[]{\includegraphics[width=1.2in]{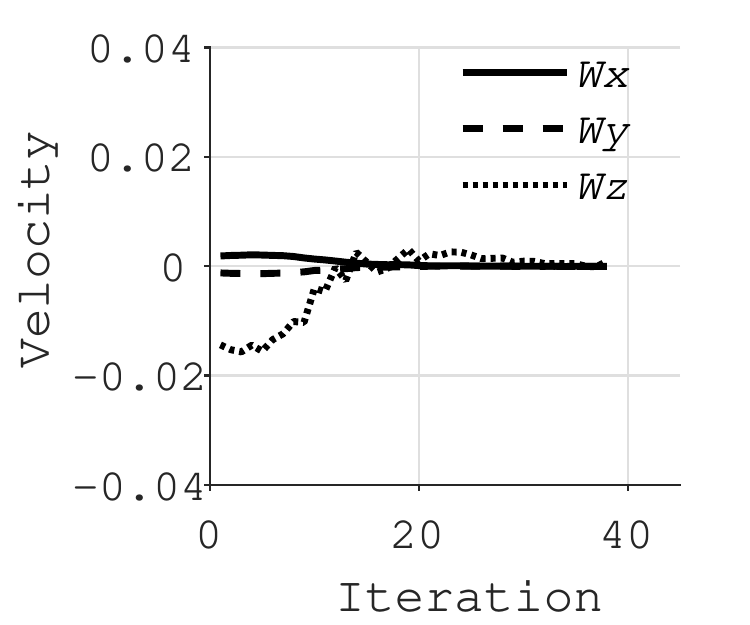}}
\subfloat[]{\includegraphics[width=1.2in]{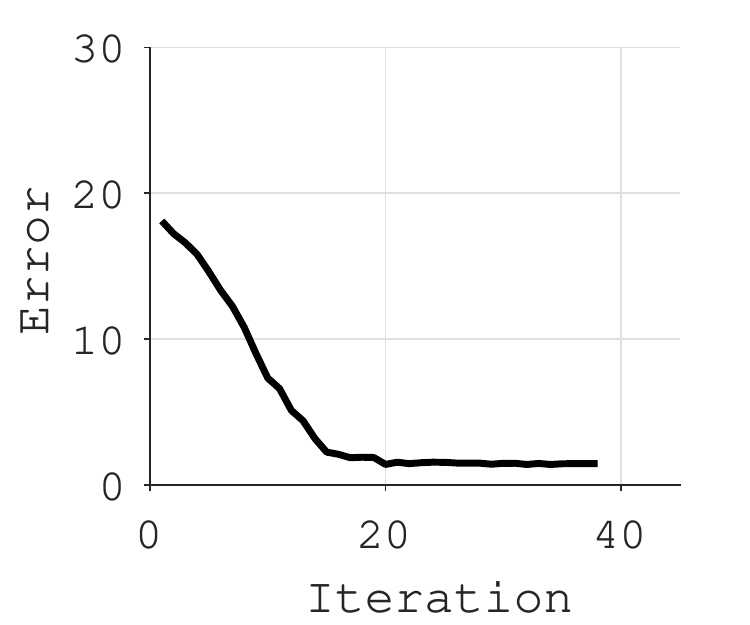}}\\
\vspace{-1.5\baselineskip}

\subfloat[]{\includegraphics[width=1.2in]{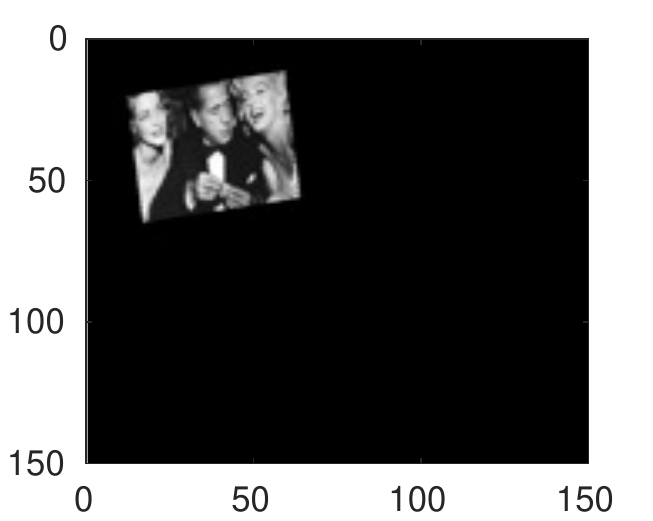}}
\subfloat[]{\includegraphics[width=1.2in]{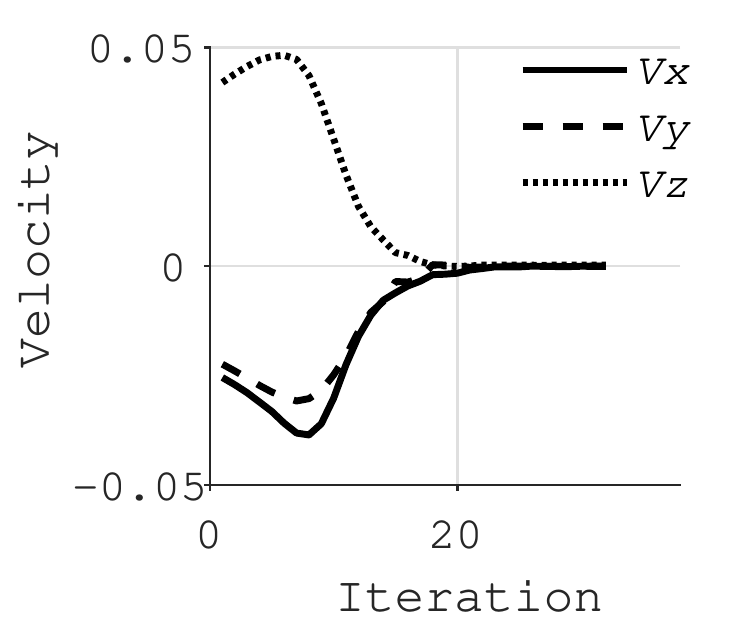}}
\subfloat[]{\includegraphics[width=1.2in]{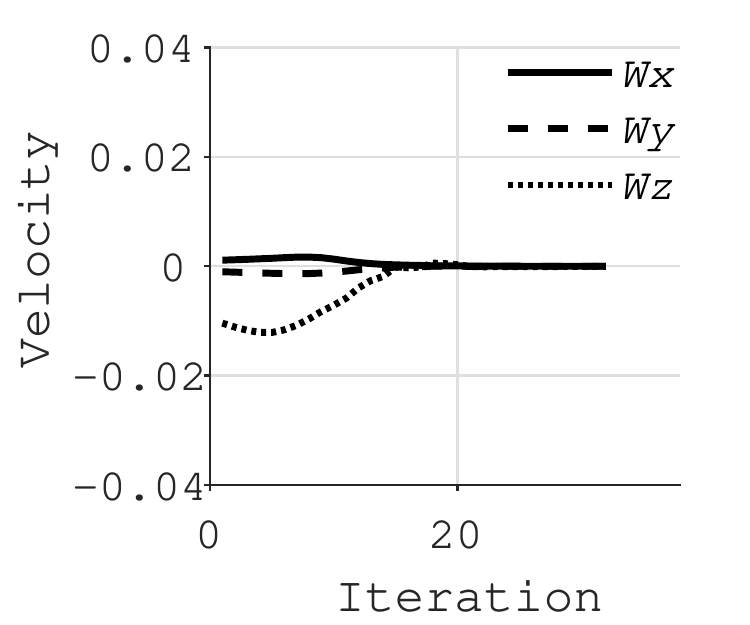}}
\subfloat[]{\includegraphics[width=1.2in]{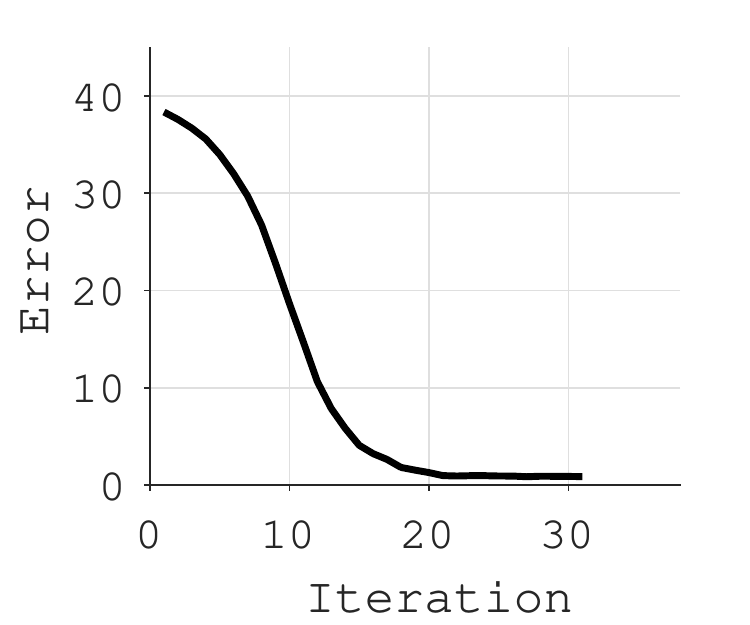}}\\
\vspace{-0.6cm}
    \caption{SMM visual servoing 6-DOF case using different image resolutions. First and second row uses $(50\times50)$ and $(100\times100)$ pixels respectively. The columns are initial image, linear velocities, angular velocity and norm error.}
    \label{fig:smm_img_size}
\end{figure}

\begin{figure}[htp!]
\centering
\captionsetup[subfigure]{labelformat=empty}
\subfloat[]{\includegraphics[width=1.5in]{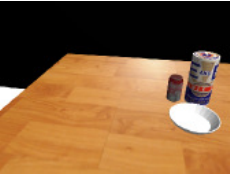}}
\hspace{0.1cm}
\subfloat[]{\includegraphics[width=1.5in]{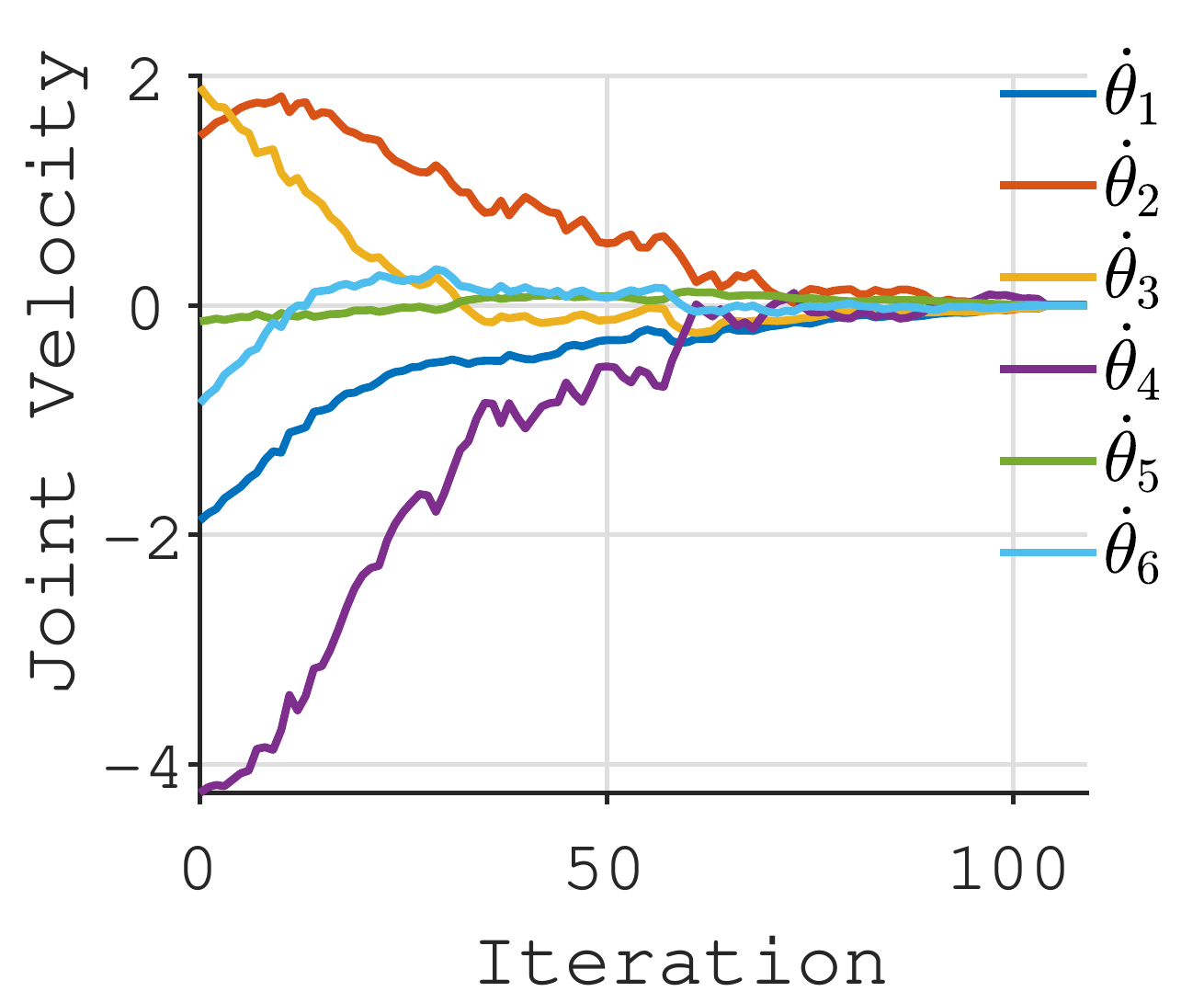}}
\hspace{0.1cm}
\subfloat[]{\includegraphics[width=1.5in]{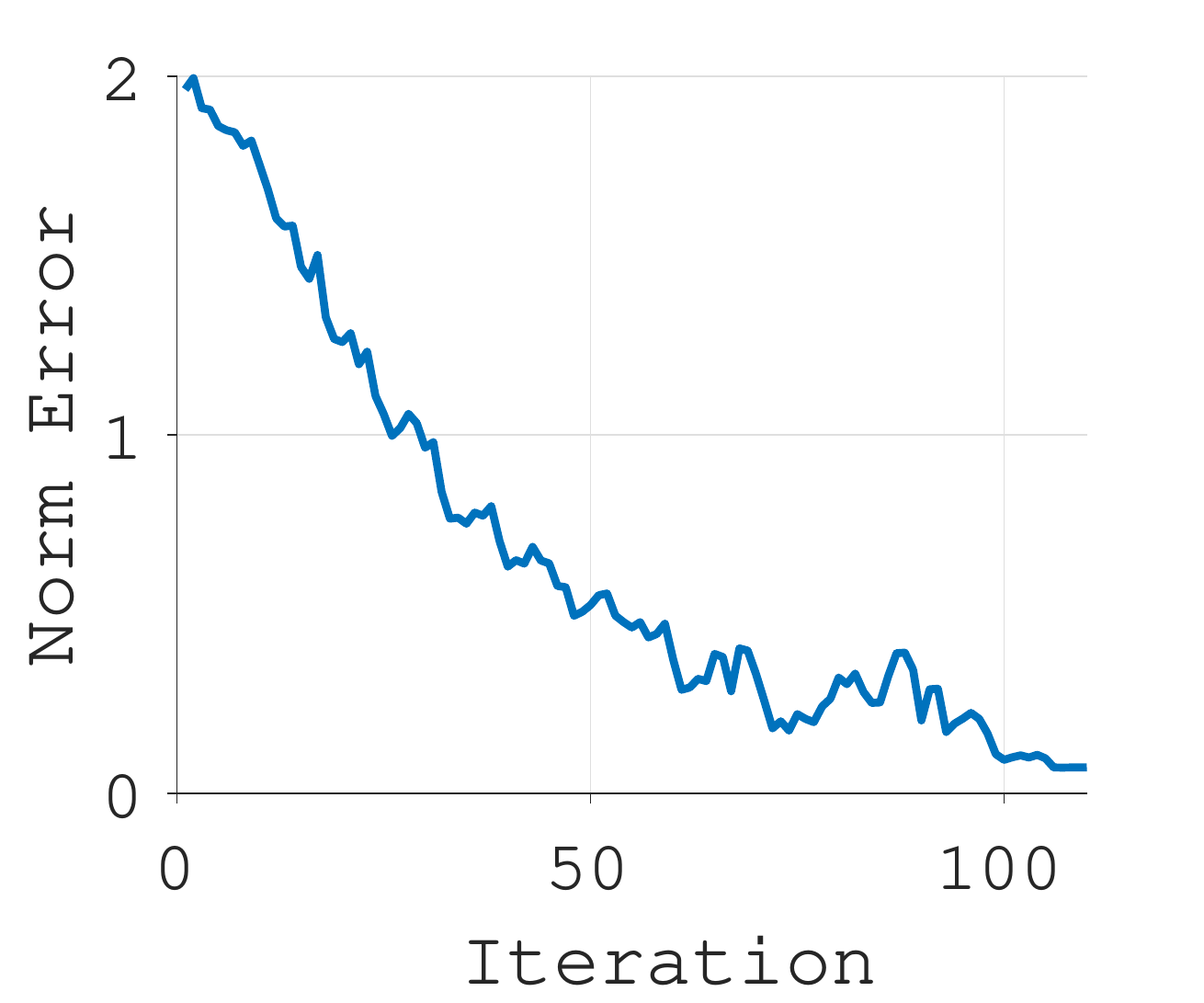}}\\
\vspace{-1.5\baselineskip}
\subfloat[]{\includegraphics[width=1.5in]{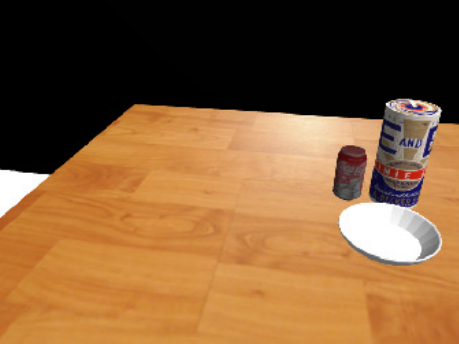}}
\hspace{0.1cm}
\subfloat[]{\includegraphics[width=1.5in]{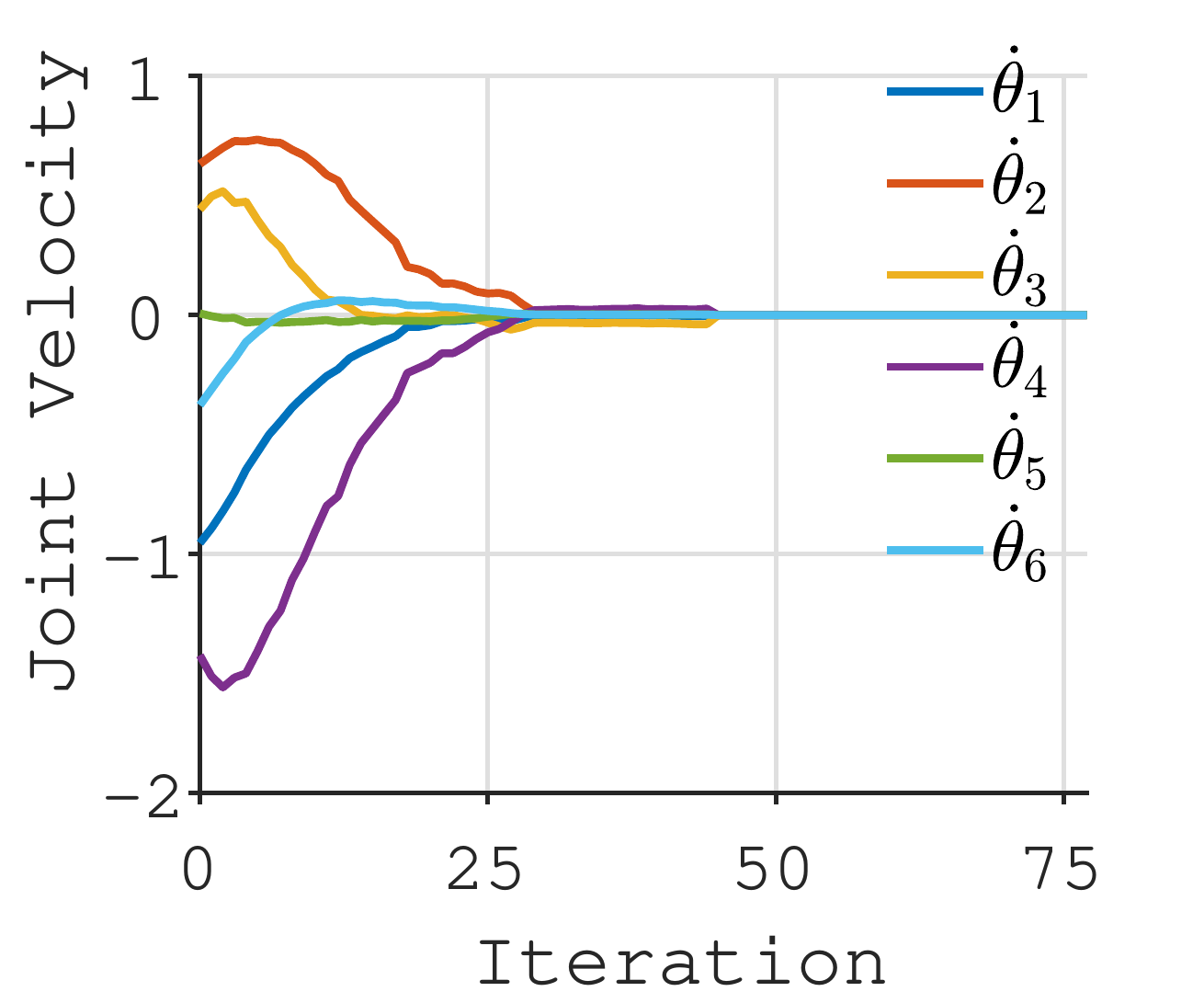}}
\hspace{0.1cm}
\subfloat[]{\includegraphics[width=1.5in]{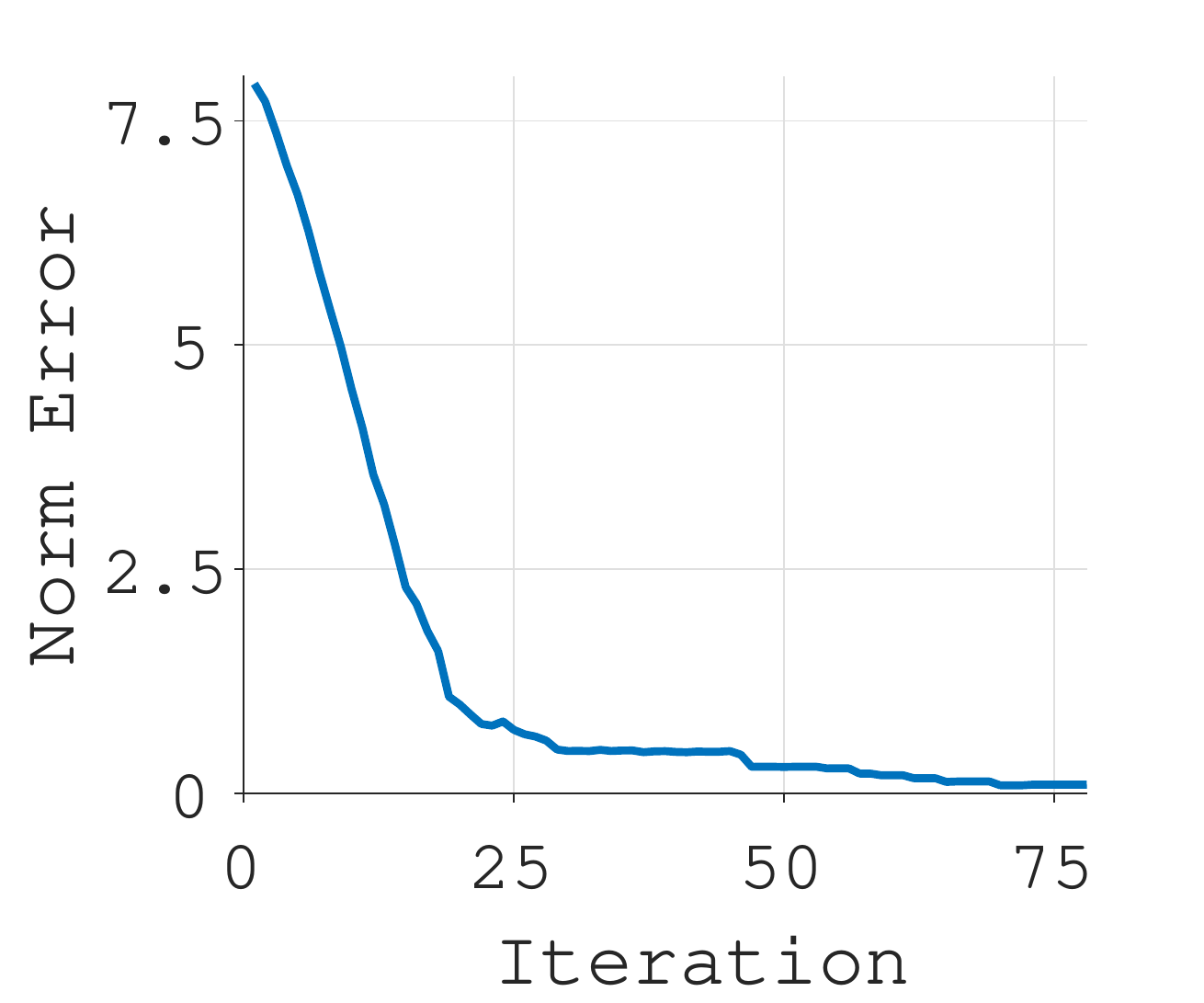}}\\
\vspace{-0.6cm}
\caption{SMM visual servoing with UR5 manipulator using different image resolutions. First and second row uses $(50\times50)$ and $(100\times100)$ pixels respectively. The columns are initial image, joint velocities and norm error}
    \label{fig:3D_50x50}
\end{figure}

Here, two test cases are considered in order to compare the execution of proposed approach. Case A uses small image size of $50\times50$ pixels and Case B uses larger image size of $100\times100$ pixel. First 3D positioning task was considered where for both cases, the initial configurations of camera were taken same as given by experiment 4 in Table \ref{tb:exp_3D}. The row 1 and row 2 of Fig. \ref{fig:smm_img_size} show results for Case A and Case B, respectively. Similarly, results were obtained using UR5 manipulator for 3D object case in Fig. \ref{fig:3D_50x50} where rows 1 and 2 show the results for Case A and Case B, respectively. 

From these plots it can be observed that even though both cases converge and follow exponential decay, error in Fig.\ref{fig:smm_img_size} (Column 4) and Fig. \ref{fig:3D_50x50} (Column 3), Case B have slightly better convergence than Case A. It can also be observed that velocity profiles in Fig.\ref{fig:smm_img_size} (Column 2-3) and Fig.\ref{fig:3D_50x50} (Column 2) are smoother when $(100\times100)$ image is used. This infers that the increase in image size results in smoothness and better convergence. But as image size increases the number of calculations for computation of SMM also increases.  Hence, there is a trade off between performance and cost of operation for a particular servoing application. 
\subsubsection{Robustness to image content} 
\begin{figure}[t]
\centering
\captionsetup[subfigure]{labelformat=empty}
\subfloat[]{\includegraphics[width=1.5in]{figures/Final_3D_posn2-eps-converted-to.pdf}}
\hspace{0.2cm}
\subfloat[]{\includegraphics[width=1.55in]{figures/jvel_3D_2.png}}
\hspace{0.01cm}
\subfloat[]{\includegraphics[width=1.55in]{figures/er_3D_2.png}}\\
\vspace{-1.6\baselineskip}

\subfloat[]{\includegraphics[width=1.5in]{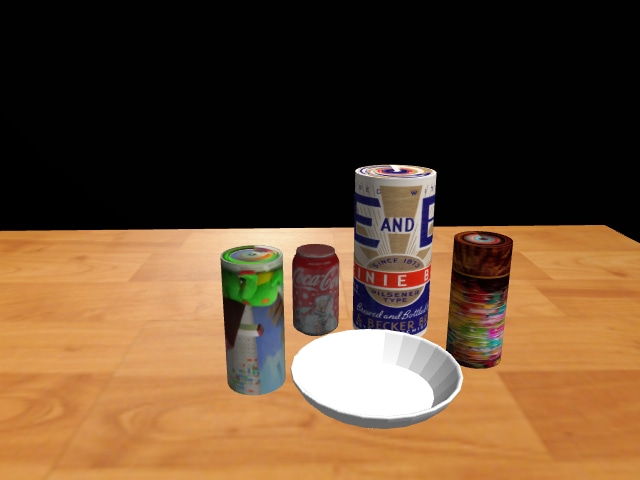}}
\hspace{0.2cm}
\subfloat[]{\includegraphics[width=1.5in]{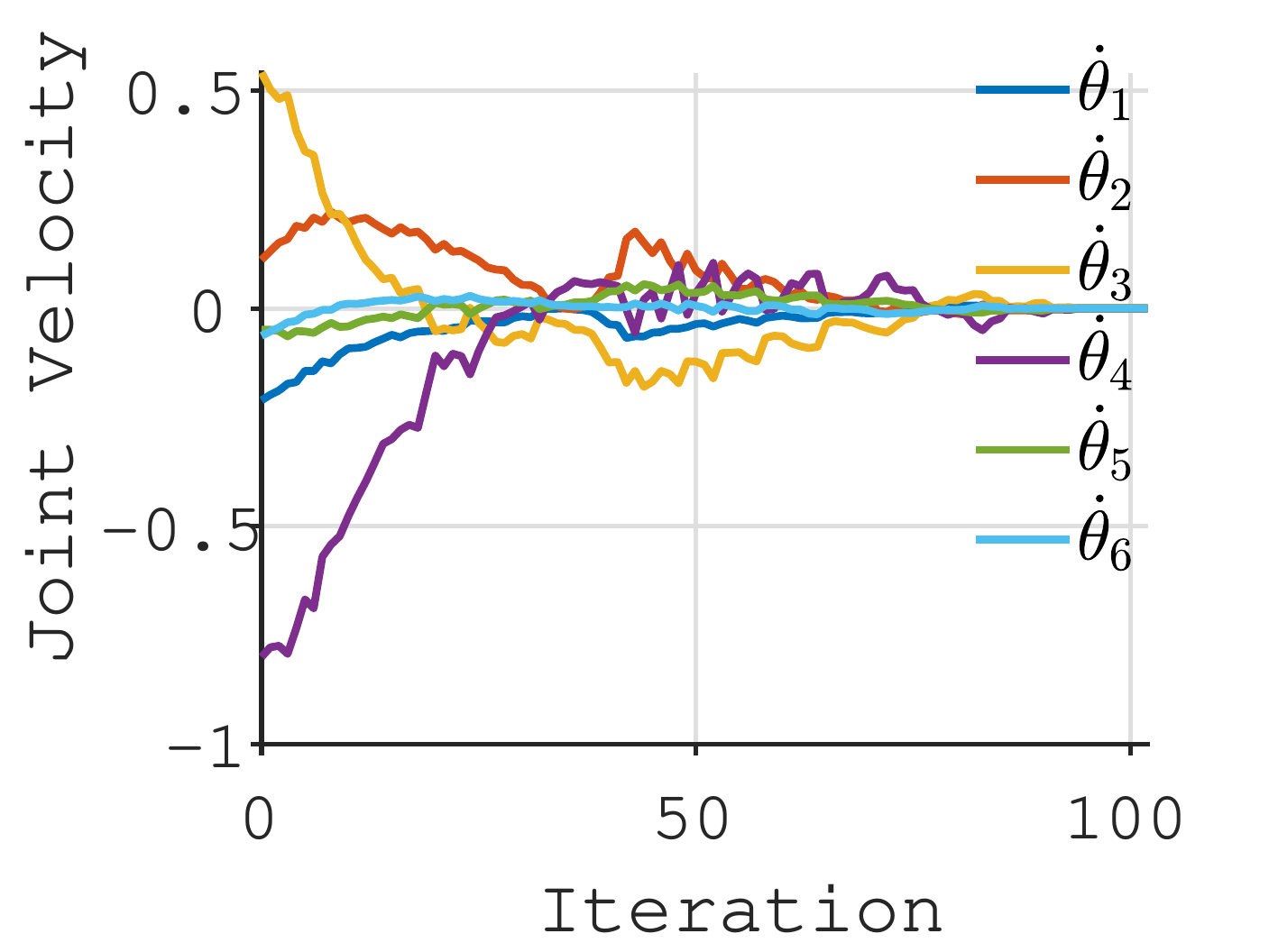}}
\hspace{0.03cm}
\subfloat[]{\includegraphics[width=1.5in]{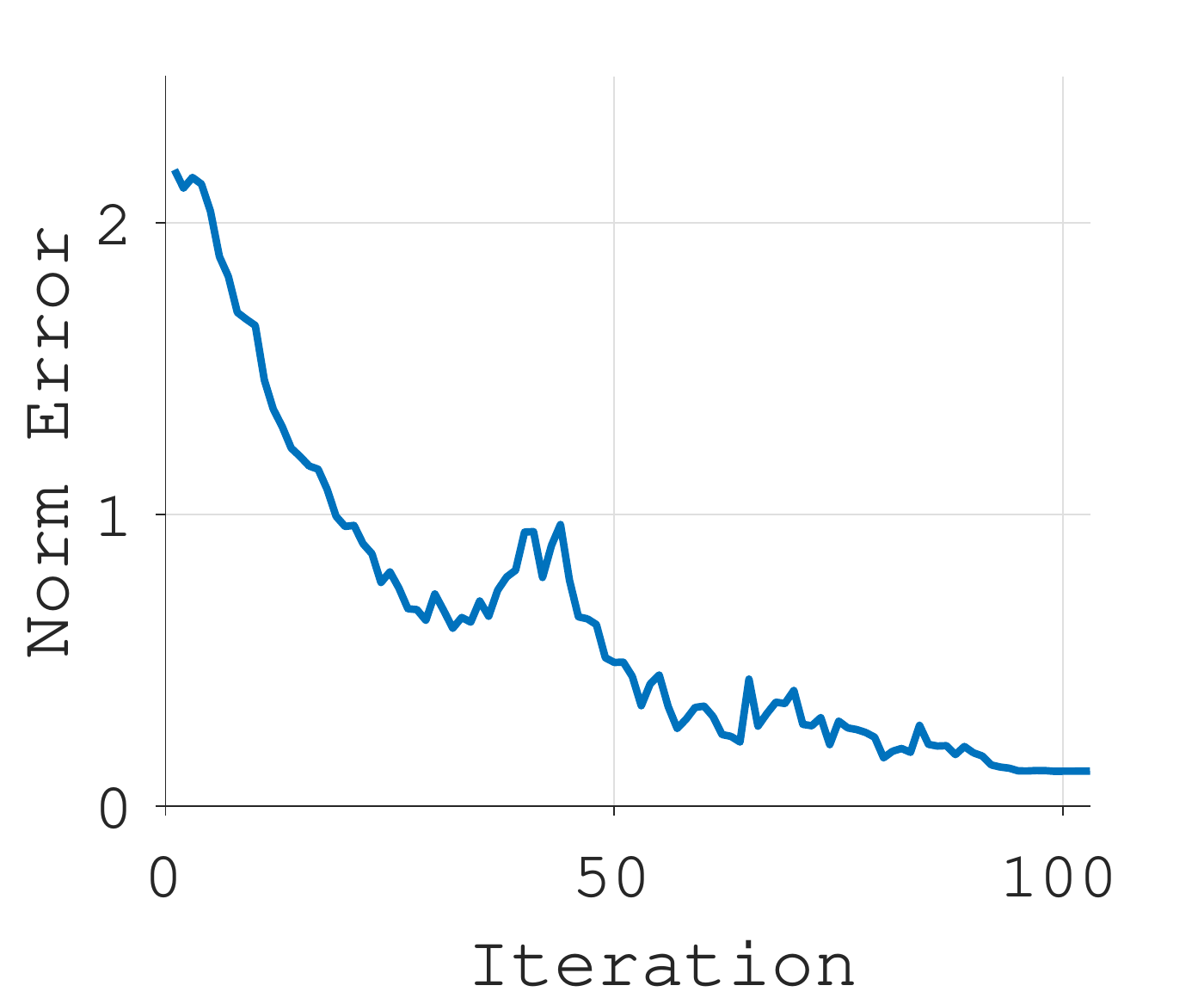}}\\
\vspace{-1.6\baselineskip}

\subfloat[]{\includegraphics[width=1.5in]{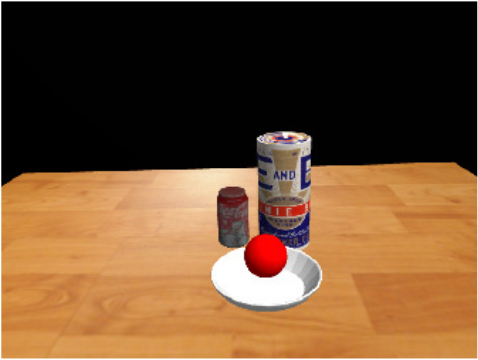}}
\hspace{0.15cm}
\subfloat[]{\includegraphics[width=1.5in]{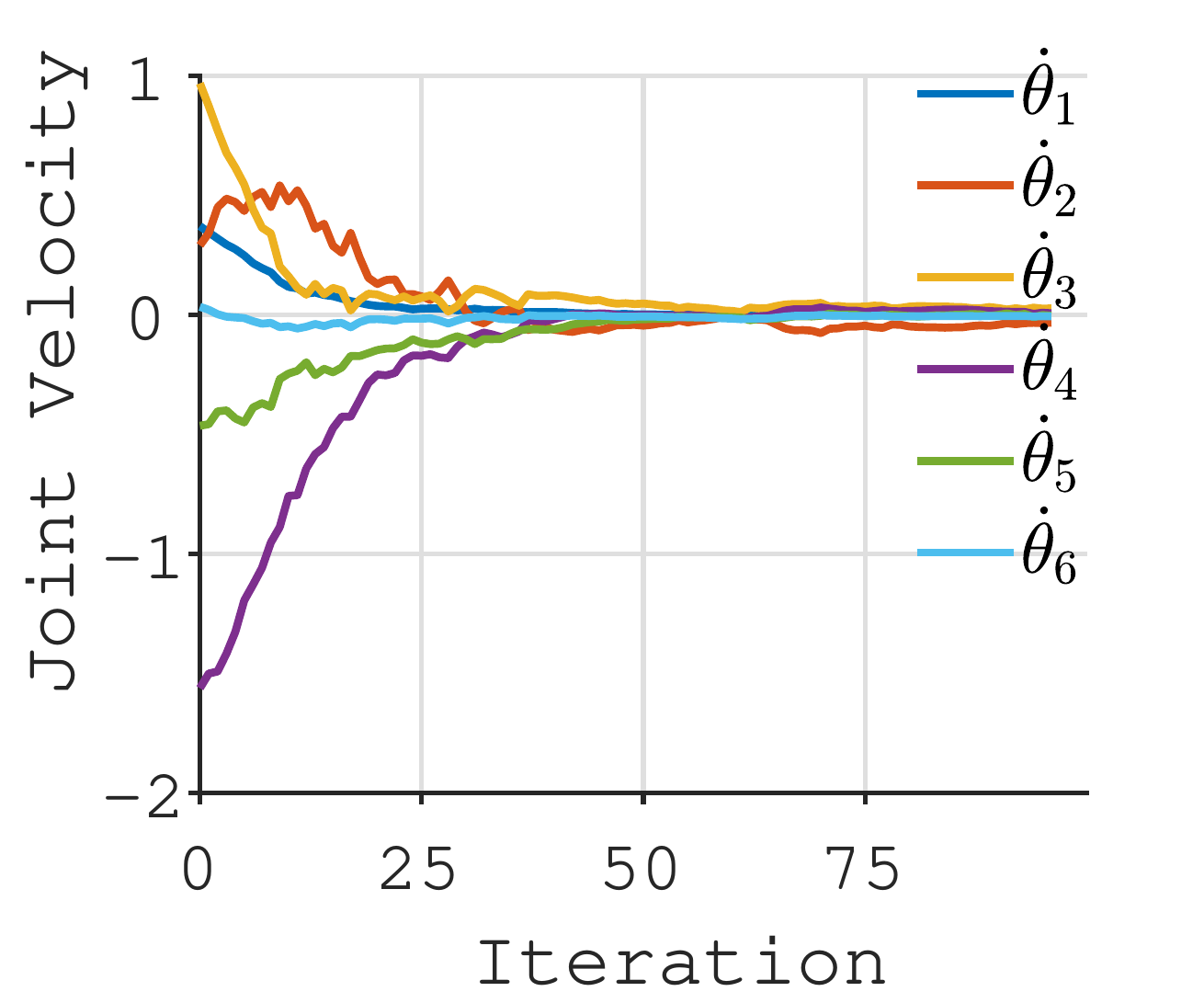}}
\hspace{0.01cm}
\subfloat[]{\includegraphics[width=1.5in]{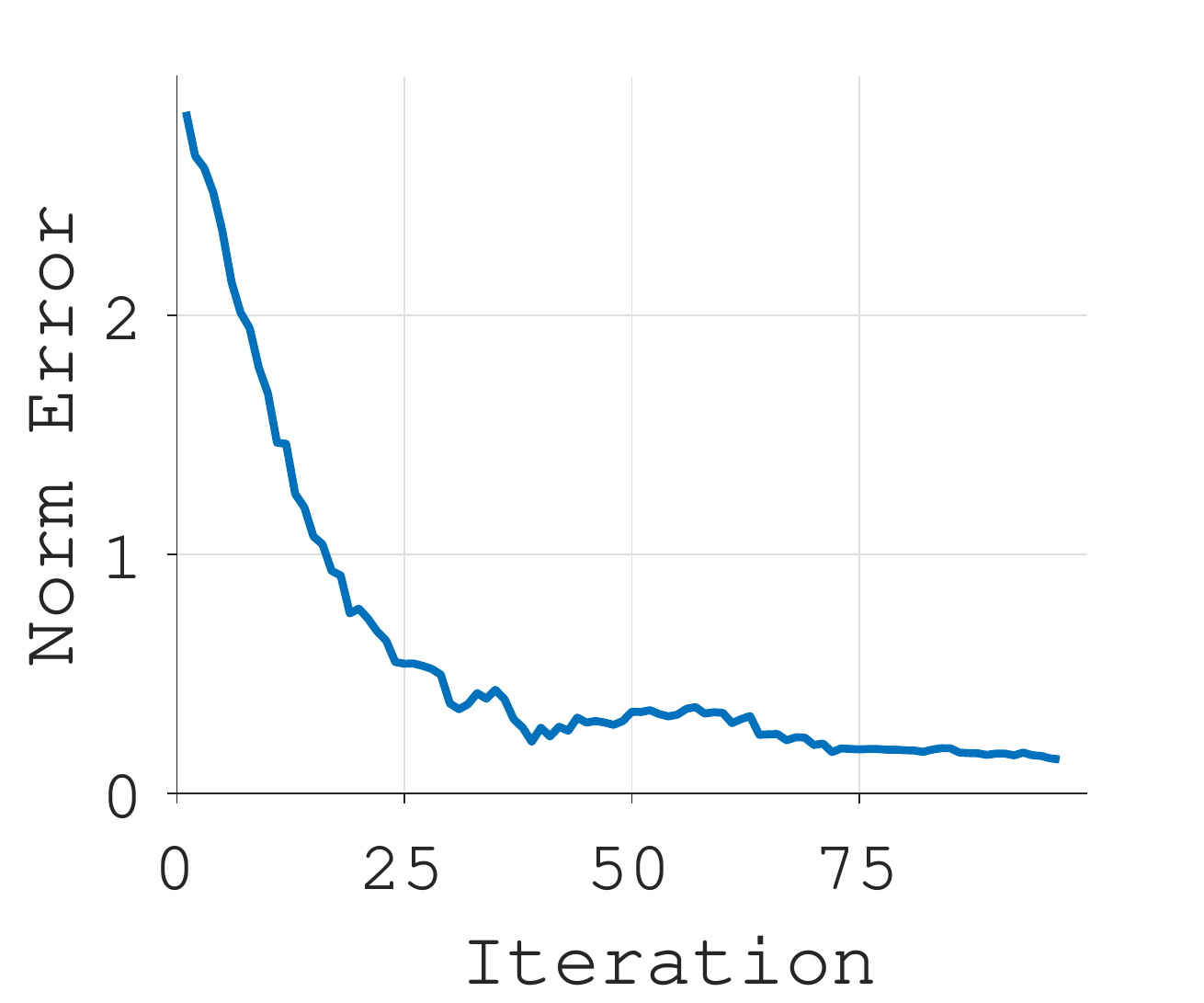}}\\
\vspace{-1\baselineskip}
\caption{SMM visual servoing of UR5 manipulator using different image content. Column 1-4 shows the image, manipulator joint velocities (rad/s) and norm error (pixels) plot.}
    \label{fig:3D_posn1}
\end{figure}

In this subsection, further investigations are carried out to demonstrate robustness in terms of different images and image contents.
\begin{figure}[t]
\centering
\captionsetup[subfigure]{labelformat=empty}
\subfloat[]{\includegraphics[width=1.3in,height=0.8in]{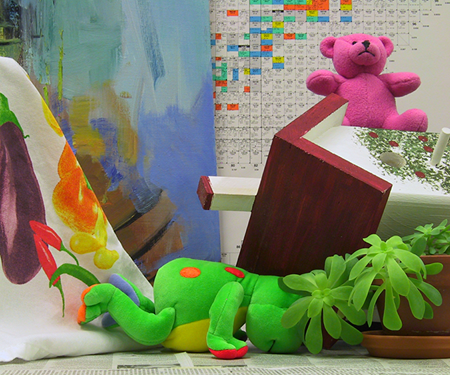}}
\subfloat[]{\includegraphics[width=1.2in]{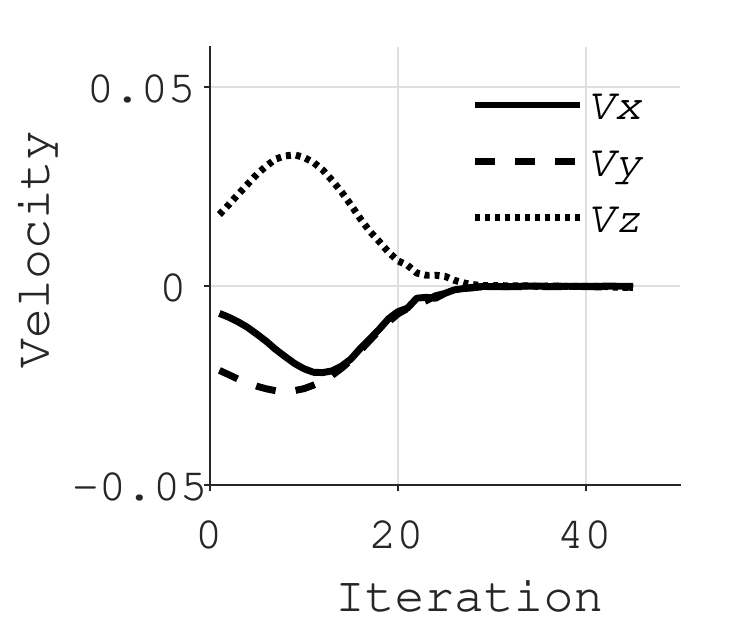}}
\subfloat[]{\includegraphics[width=1.2in]{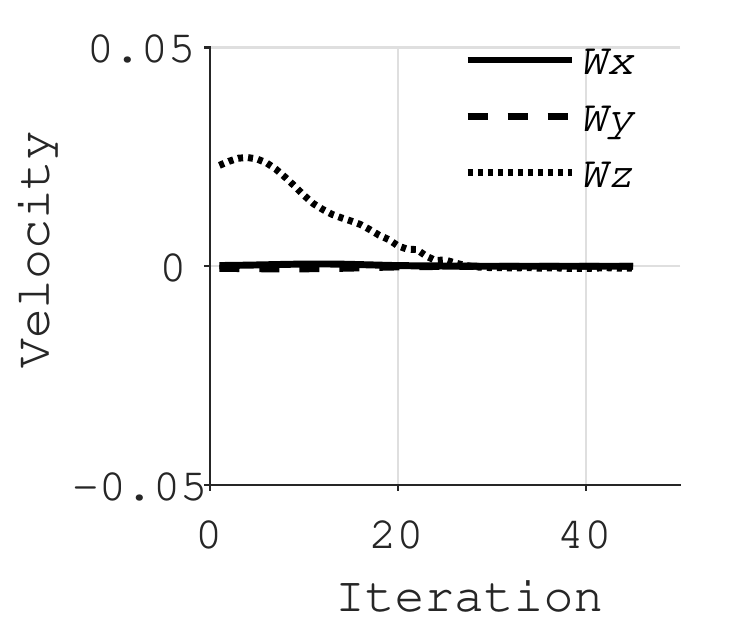}}
\subfloat[]{\includegraphics[width=1.2in]{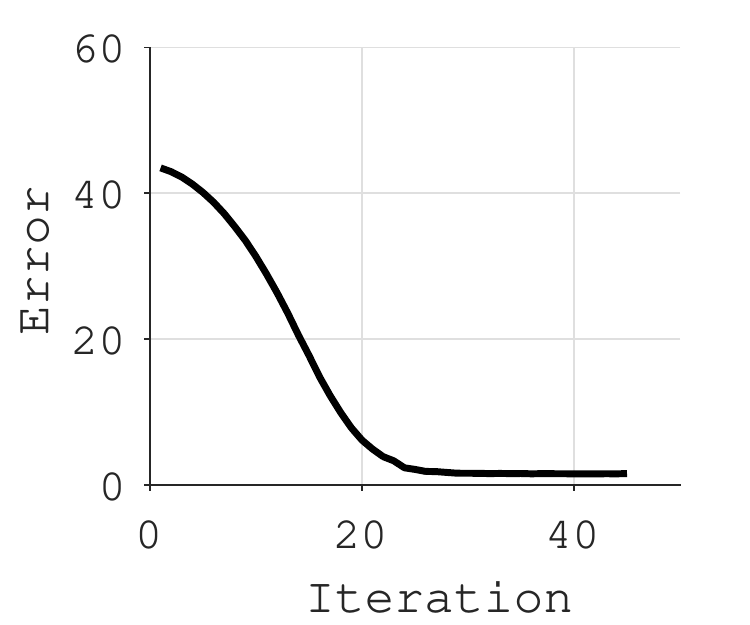}}\\
\vspace{-1.9\baselineskip}
\subfloat[]{\includegraphics[width=1.3in,height=0.9in]{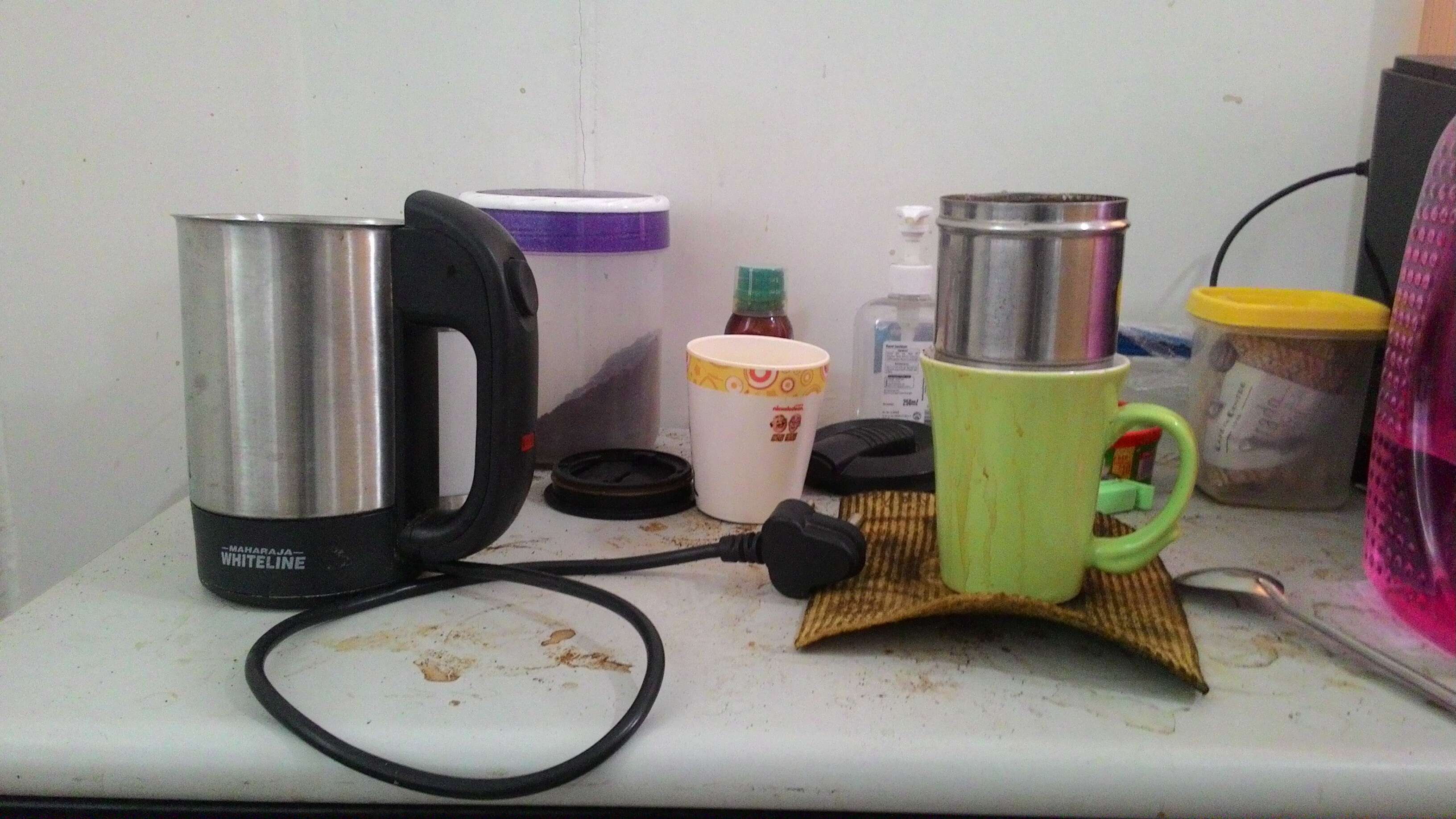}}
\subfloat[]{\includegraphics[width=1.2in]{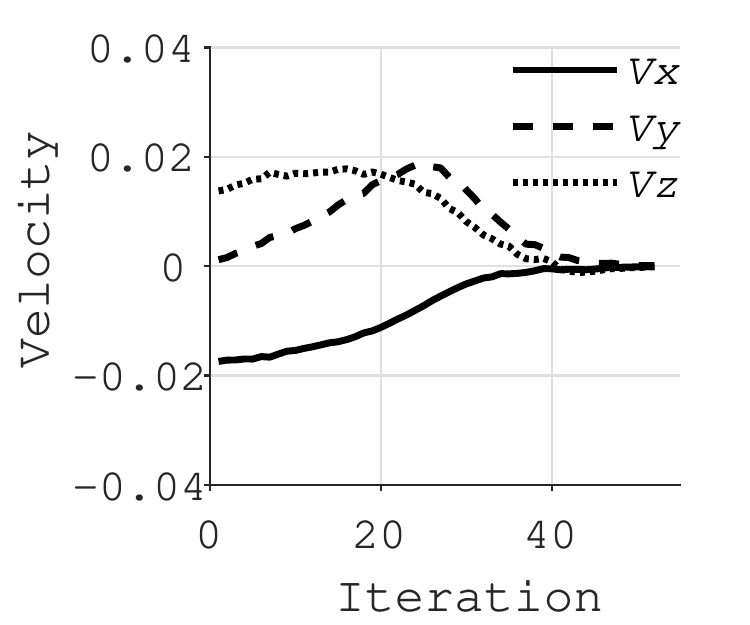}}
\subfloat[]{\includegraphics[width=1.2in]{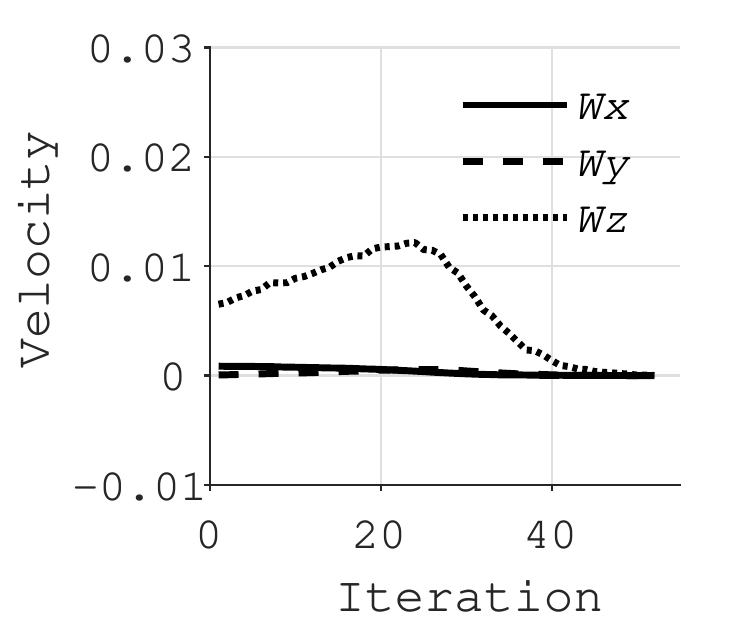}}
\subfloat[]{\includegraphics[width=1.2in]{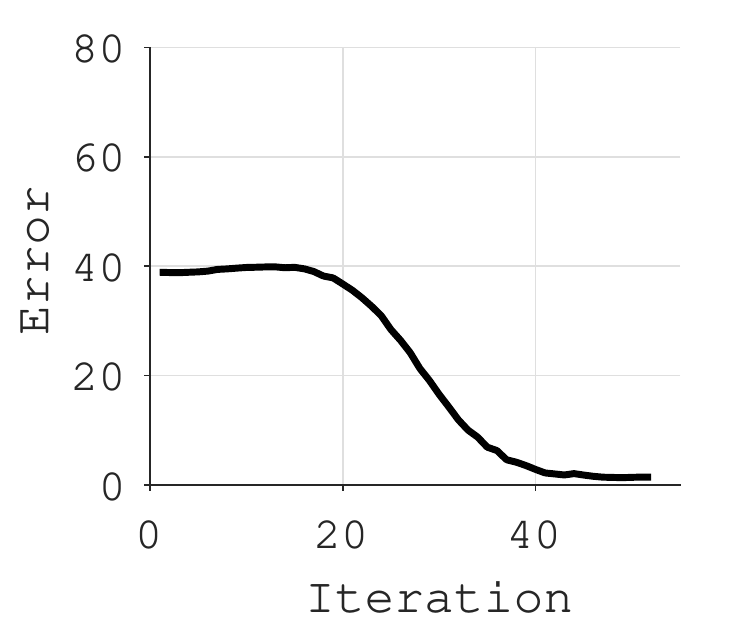}}\\
    \caption{SMM visual servoing 6-DOF case using using different image content. Column 1-4 shows the image, camera linear and angular velocities, and error plot.}
    \label{fig:diff_img_cont}
\end{figure}
The experiments were conducted to depicts performance of the proposed framework for different textured images from different poses. In Fig. \ref{fig:diff_img_cont} two different images have been considered. The first include standard test image (Fig. \ref{fig:diff_img_cont}(Row 1)) and the second image presents real objects ((Fig. \ref{fig:diff_img_cont}(Row 2)) using a web cam.  Another set of experiments was conducted in gazebo using a UR5 manipulator wherein different objects were used in the environment along with different poses as shown in Fig.\ref{fig:3D_posn1}. 

The results of the experiments in Fig. \ref{fig:diff_img_cont}  and Fig.\ref{fig:3D_posn1} depict that norm of error and  joint velocities for different objects converges to the desired values. It is worth noting that for the latter case, the control law converges even in the case of a low-textured scenes and luminance variations. This shows efficacy of the proposed framework when using different shape and structure of the objects in the scene.

\subsubsection{Robustness to partial occlusions}
\begin{figure}[t]
\centering
\subfloat[]{\includegraphics[width=1.45in]{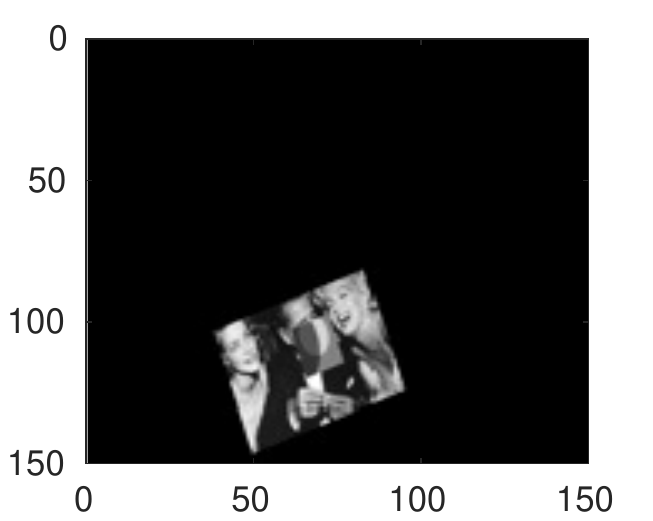}}
\hspace{0.01cm}
\subfloat[]{\includegraphics[width=1.45in]{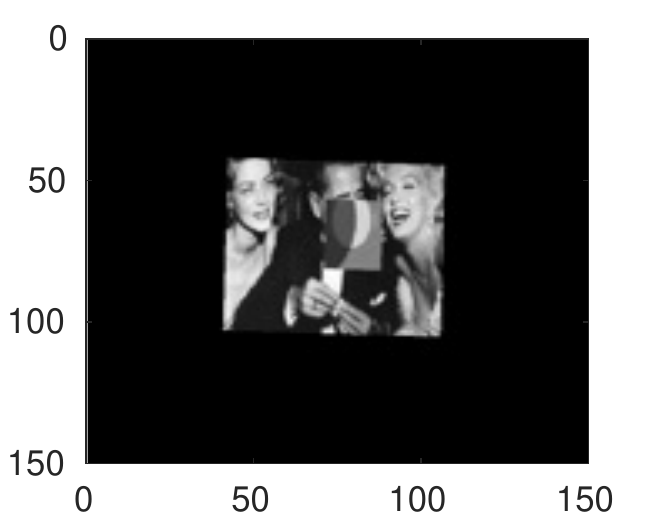}}
\hspace{0.01cm}
\subfloat[]{\includegraphics[width=1.55in]{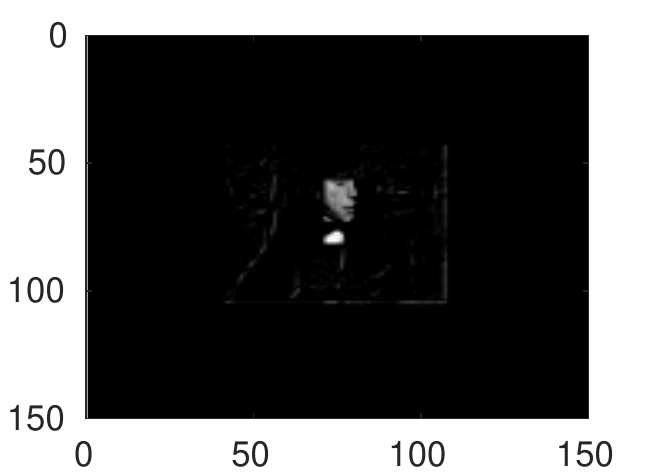}}\\
\vspace{-0.45cm}
\subfloat[]{\includegraphics[width=1.45in]{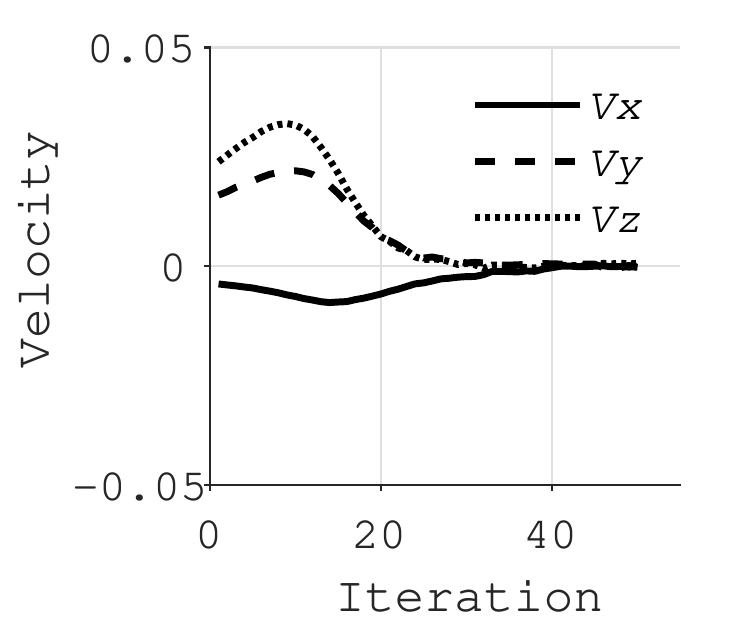}}
\hspace{0.01cm}
\subfloat[]{\includegraphics[width=1.45in]{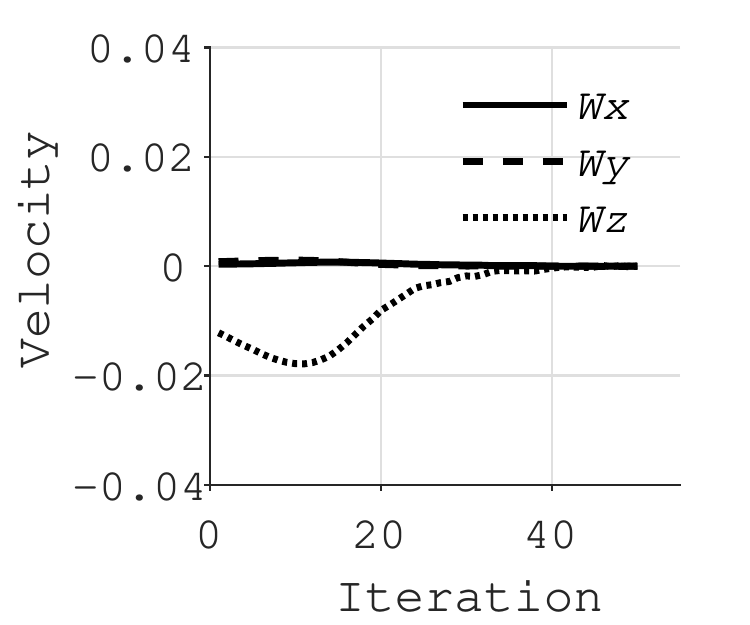}}
\hspace{0.01cm}
\subfloat[]{\includegraphics[width=1.55in]{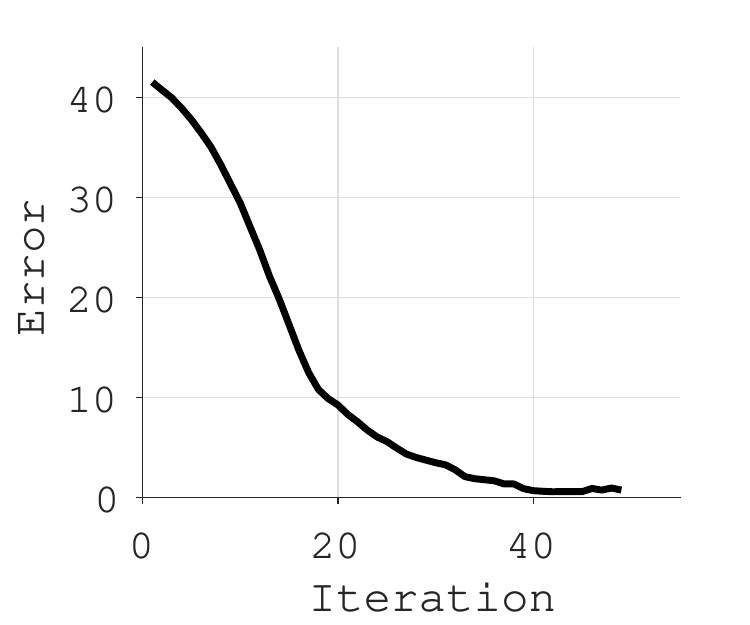}}\\
\vspace{-0.4cm}
    \caption{SMM visual servoing 6-DOF case with occlusion. (a) Initial image. (b) Final image. (c) $I-I^*$ at desired position. (d) Linear velocity. (e) Angular velocity. (f) norm error}
    \label{fig:occlu}
\end{figure}

As the proposed method uses mixture models of photometric information, its sensitivity to partial occlusions is crucial. Hence, robustness to partial occlusion is tested in this subsection. For introducing occlusion, one of the face in image under consideration is partially covered by a synthetic image patch as shown in Fig. \ref{fig:occlu}(b). Here, it must be noted that the desired image is still the one without the patch. For the experiment shown in Fig. \ref{fig:occlu} the starting pose is same as given in Experiment 5 in Table \ref{tb:exp_3D}. Despite the occluded image considered, the control law still converges as evident from velocity and error plot in Fig. \ref{fig:occlu}(d-f). 
\begin{figure}[htp!]
\centering
\subfloat[]{\includegraphics[width=1.5in]{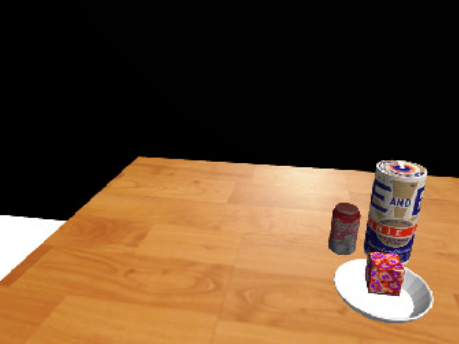}}
\hspace{0.01cm}
\subfloat[]{\includegraphics[width=1.5in]{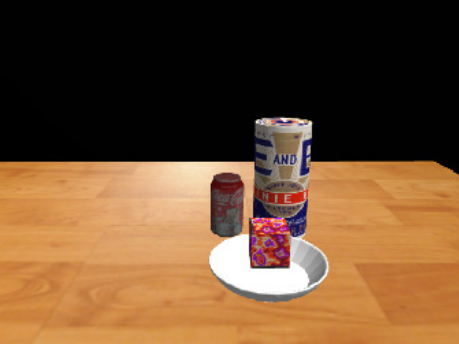}}
\hspace{0.01cm}
\subfloat[]{\includegraphics[width=1.5in]{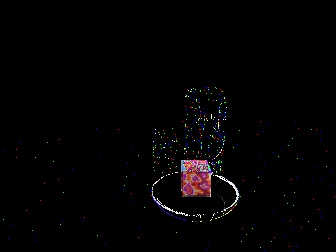}}\\
\vspace{-0.5cm}
\subfloat[]{\includegraphics[width=1.5in]{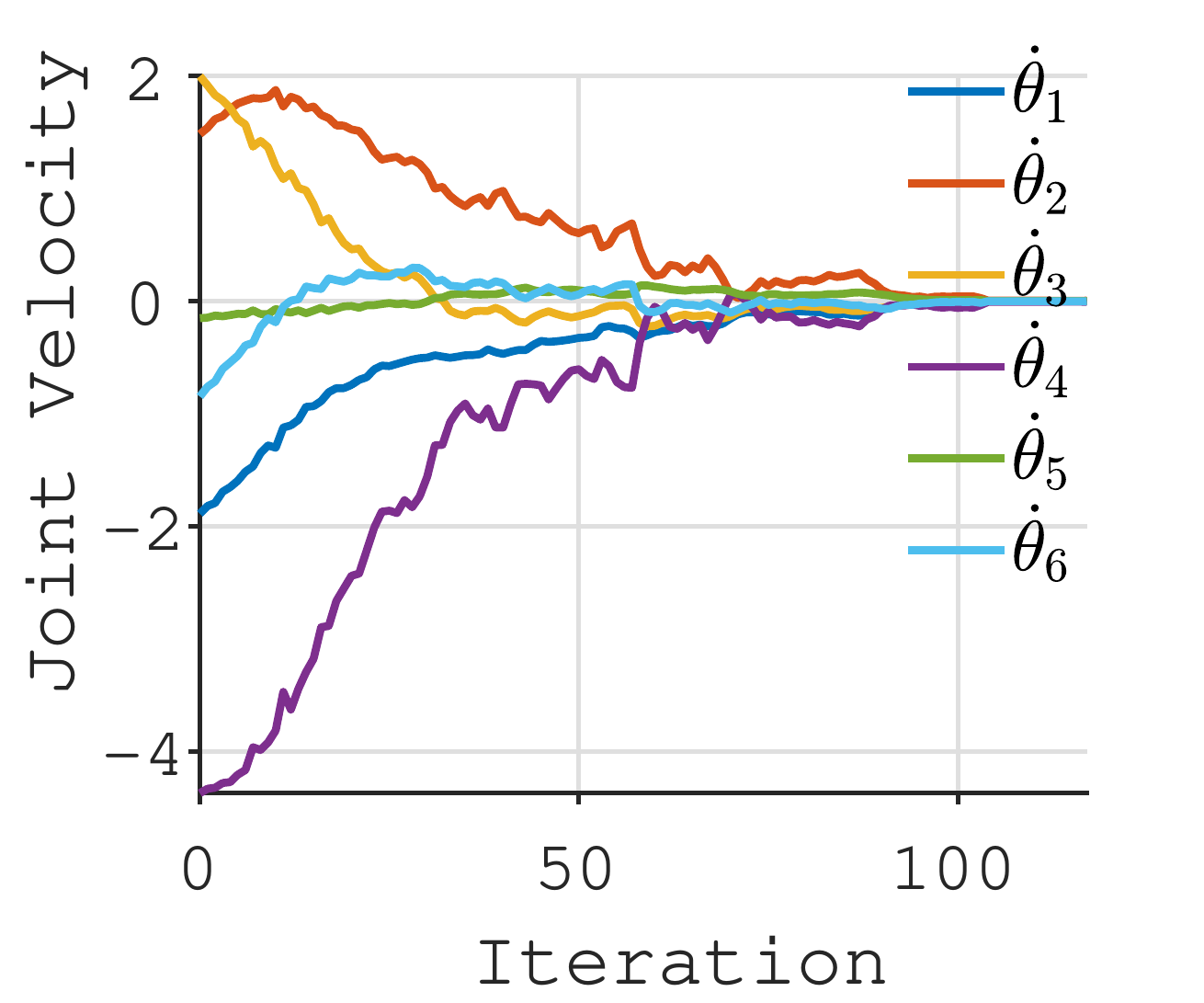}}
\hspace{0.01cm}
\subfloat[]{\includegraphics[width=1.5in]{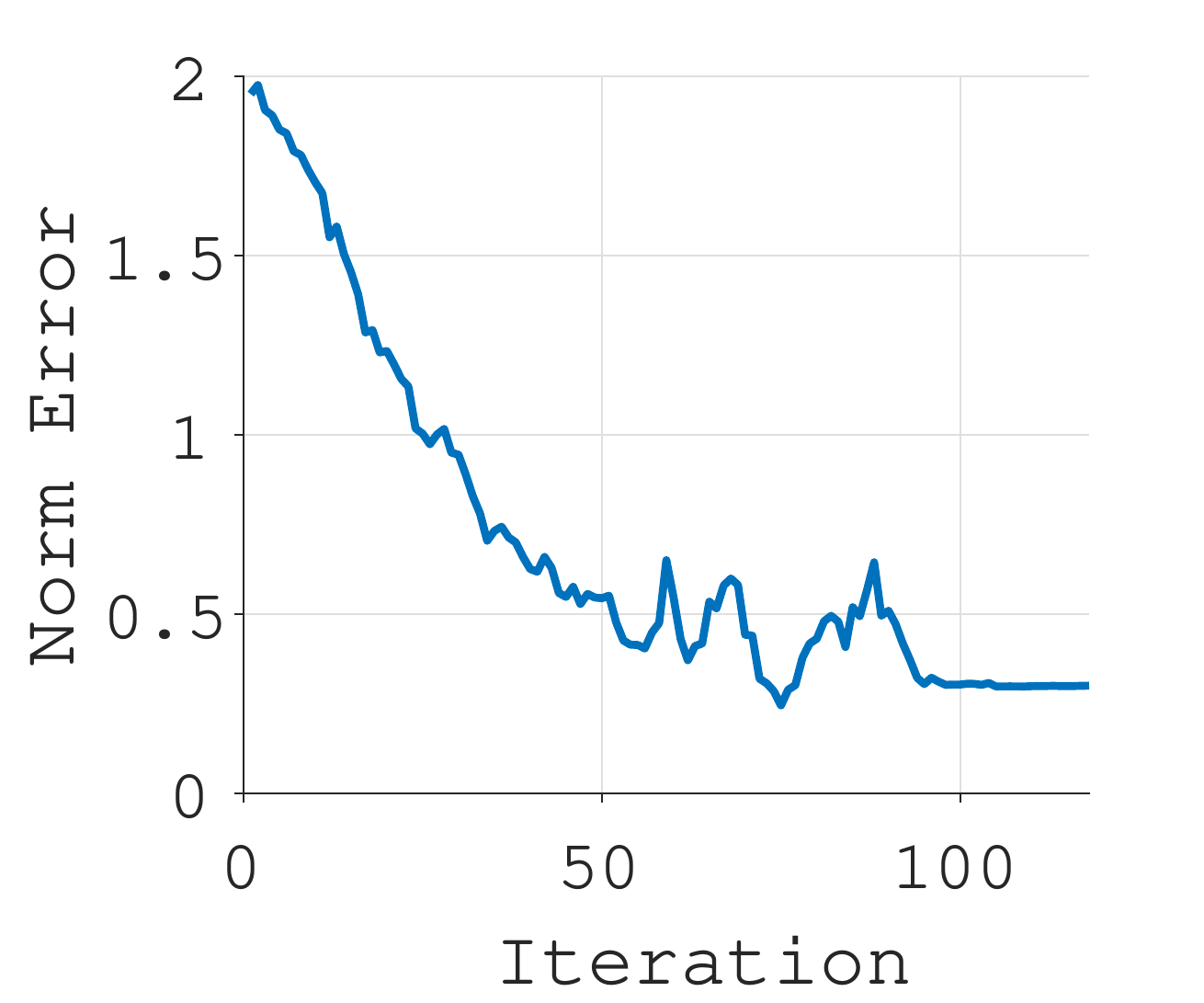}}
\hspace{0.01cm}
\subfloat[]{\includegraphics[width=1.55in]{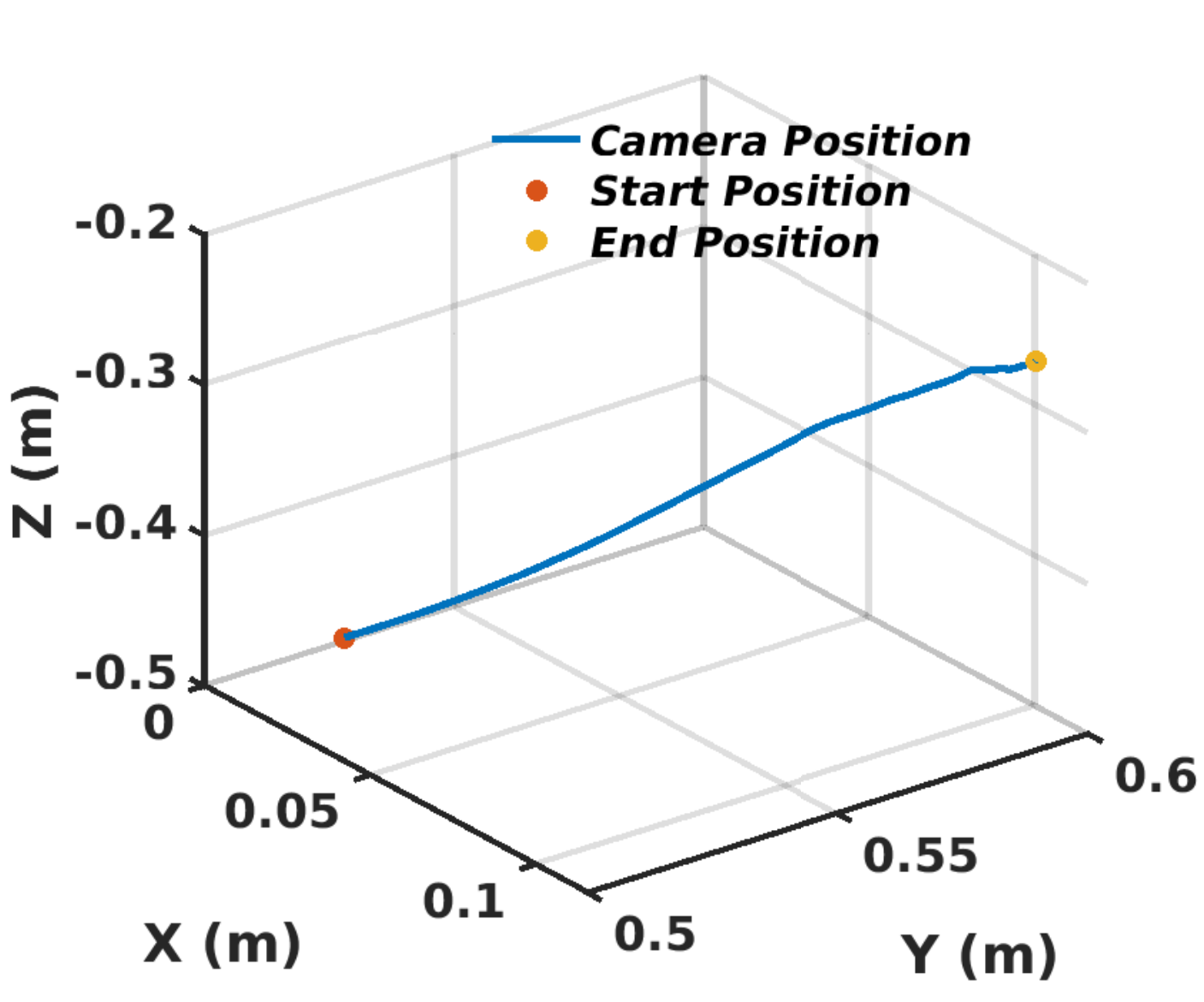}}\\
\vspace{-0.4cm}
\caption{SMM visual servoing in 3D with occlusion. (a) Initial image. (b) Final image. (c) $I-I^*$ at desired position (d) manipulator joint velocities (rad/s). (e) Norm error (pixels). (f) Camera position}
    \label{fig:3D_obstacle}
\end{figure}
A more realistic scenario is considered in Fig. \ref{fig:3D_obstacle} where a small cube was introduced in the scene while performing visual servoing using UR5.  For both cases, there exist error in final and desired images ($\boldsymbol{I}-\boldsymbol{I^*}$) in terms of pixel intensity values  as evident from Fig. \ref{fig:occlu}(c) and Fig. \ref{fig:3D_obstacle}(c). As the final image is not the desired one, the error cannot vanish completely at the end of visual servoing. It is worth noting  that regardless of the non-zero value of error, the convergence is not affected by partial occlusions and final pose is achieved. This shows the robustness of the proposed approach to partial occlusions.

\section{CONCLUSIONS AND FUTURE WORK}\label{sec:future_work}
In this paper, a student $t$-distribution mixture model based framework is proposed for visual servoing. Instead of using image features, the entire image is modeled as a function using SMM. The SMM model enables a method to deal with non-textured/textured objects and also it serves as a replacement of methods where extracting visual features is too complex or too costly. An analytical formulation of interaction matrix using the proposed novel SMM feature model is introduced for visual servoing. The proposed method relies on the minimization of distances between probability density functions defined by SMM of current and desired images. Thus the control law is modeled as an optimization problem for minimizing the cost function.

This control law is able to perform servoing operations in successful manner even with the influence of image lighting changes, texture changes and occlusions. To prove its efficacy, the proposed algorithm was initially implemented on a numerical simulation model for 2D, 3D cases and later a realistic simulation is performed in gazedo using UR5 manipulator. The experimental results showed the effectiveness and accuracy of the proposed method for different images, different poses, different image size and under occlusions. While the proposed system works well on the numerical model, execution on the real-world setup and detailed stability analysis will be taken up as future work. The method will also be extended for visual servoing towards an object which is under motion.
\section*{References}
\bibliography{Manuscript_smm_vs}

\end{document}